\documentclass{article}

\usepackage[bottom]{footmisc}
\usepackage{float}
\usepackage{multirow}
\usepackage{parskip}

\usepackage[small]{caption}
\usepackage{subcaption}
\usepackage[margin=1.00in]{geometry}
\usepackage[utf8]{inputenc}
\usepackage[T1]{fontenc}
\usepackage{url}
\usepackage{booktabs}
\usepackage[dvipsnames,svgnames,table,xcdraw]{xcolor}
\usepackage{amsfonts}
\usepackage{nicefrac}
\usepackage{microtype}
\usepackage{mathpazo}
\usepackage{graphicx}

\usepackage[
  colorlinks=true,
  urlcolor=RoyalBlue,
  citecolor=ForestGreen,
  linkcolor=RoyalBlue
]{hyperref}
\usepackage{algorithm}
\usepackage{algpseudocode}
\usepackage{amsthm}
\usepackage{multicol}
\usepackage{enumitem}
\usepackage{amsmath}
\usepackage{amssymb}
\usepackage{natbib}
\usepackage{tabularx}
\usepackage{appendix}
\usepackage{listings}
\usepackage{inconsolata}
\usepackage{makecell}
\usepackage{xltabular}
\usepackage{tikz}
\usetikzlibrary{positioning, calc}
\usepackage{pgfplots}
\pgfplotsset{compat=newest}
\usepackage{colortbl}

\usepackage{xspace}
\newcommand{\OURS}{\textsc{Squeeze Evolve}\xspace}

\title{
\vspace{-1ex}
\OURS \\ Unified Multi-Model Orchestration for Verifier-Free Evolution}

\author{%
\parbox{\textwidth}{\centering\small
Monishwaran Maheswaran\thanks{\,Equal contribution.~$^\dag$\,Equal second.~$^\ddag$\,Co-advising.~Correspondence: \texttt{monishwaran@berkeley.edu}, \texttt{xuchenfeng@utexas.edu}.}$^{*,1,5}$~~%
Leon Lakhani$^{*,1}$~~%
Zhongzhu Zhou$^{\dag,5}$~~%
Shijia Yang$^{\dag,2}$~~%
Junxiong Wang$^{5}$~~%
\\[1pt]
Coleman Hooper$^{1}$~~%
Yuezhou Hu$^{1}$~~%
Rishabh Tiwari$^{1}$~~%
Jue Wang$^{5}$~~%
Harman Singh$^{1}$~~%
Qingyang Wu$^{5}$~~%
Yuqing Jian$^{5}$~~%
\\[1pt]
Ce Zhang$^{5}$~~%
Kurt Keutzer$^{1}$~~%
Tri Dao$^{4,5}$~~%
Xiaoxia Wu$^{5}$~~%
Ben Athiwaratkun$^{5}$~~%
James Zou$^{\ddag,3,5}$~~%
Chenfeng Xu$^{*,\ddag,2,5}$%
\\[3pt]
{\small
$^{1}$\,UC Berkeley\quad
$^{2}$\,UT Austin\quad
$^{3}$\,Stanford University\quad
$^{4}$\,Princeton University\quad
$^{5}$\,Together AI}%
}%
}

\date{}

\begin{document}
\maketitle
\vspace{-1.2em}
\begin{abstract}
We show that verifier-free evolution is bottlenecked by both diversity and efficiency: without external correction, repeated evolution accelerates collapse toward narrow modes, while the uniform use of a high-cost model wastes compute and quickly becomes economically impractical. We introduce \OURS{}, a unified multi-model orchestration framework for verifier-free evolutionary inference. Our approach is guided by a simple principle: allocate model capability where it has the highest marginal utility. Stronger models are reserved for high-impact stages, while cheaper models handle the other stages at much lower costs. This principle addresses diversity and cost-efficiency jointly while remaining lightweight. \OURS{} naturally supports open-source, closed-source, and mixed-model deployments. Across AIME 2025, HMMT 2025, LiveCodeBench V6, GPQA-Diamond, ARC-AGI-V2, and multimodal vision benchmarks, such as MMMU-Pro and BabyVision, \OURS{} consistently improves the cost–capability frontier over single-model evolution and achieves new state-of-the-art results on several tasks. Empirically, \OURS{} reduces API cost by up to $\sim$3$\times$ and increases fixed-budget serving throughput by up to $\sim$10$\times$. Moreover, on discovery tasks, \OURS{} is the first verifier-free evolutionary method to match, and in some cases exceed, the performance of verifier-based evolutionary methods.
\end{abstract}
\begin{center}
\small
\href{https://github.com/squeeze-evolve/squeeze-evolve}{\textbf{Code}} \quad | \quad \href{https://squeeze-evolve.github.io/}{\textbf{Project Page}}
\end{center}
\vspace{1ex}
\begin{figure}[h]
\centering
\includegraphics[width=\linewidth]{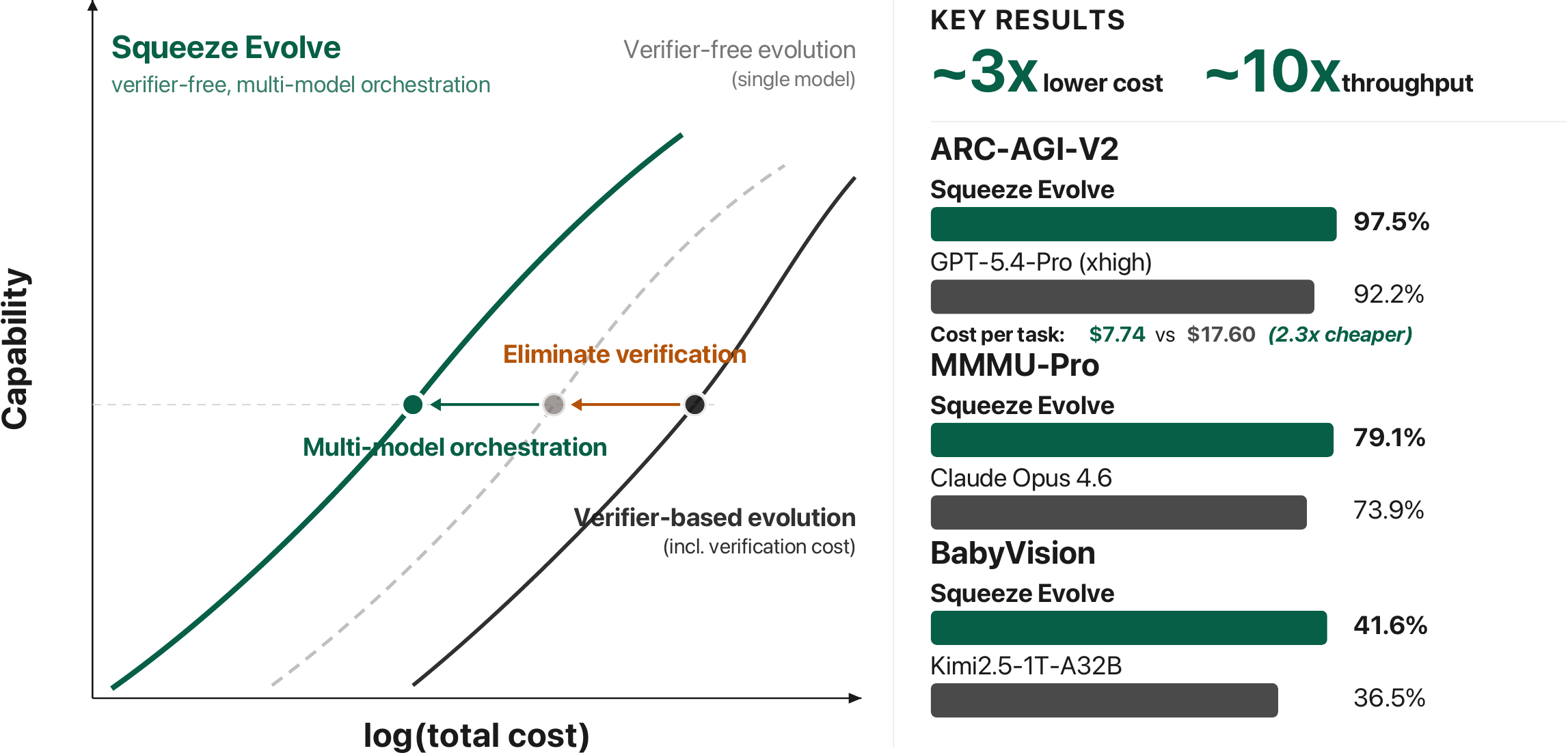}
\caption{\OURS{} shifts the cost--capability frontier left by combining verifier-free evolution with multi-model orchestration. \textbf{Left:} Conceptual scaling curves. \textbf{Right:} Key results across ARC-AGI-V2, MMMU-Pro, and BabyVision.}
\label{fig:teaser}
\end{figure}
\section{Introduction}
\label{sec:introduction}
Test-time scaling has emerged as a practical way to push language models beyond one-shot inference by spending additional compute at test time to search over or refine candidate solutions~\citep{wang2023selfconsistencyimproveschainthought,madaan2023selfrefineiterativerefinementselffeedback,venkatraman2026recursiveselfaggregationunlocksdeep}. A particularly promising direction is self-evolution, where models iteratively improve candidates through selection, mutation, and recombination~\citep{novikov2025alphaevolvecodingagentscientific,openevolve,lange2025shinkaevolveopenendedsampleefficientprogram,liu2026evoxmetaevolutionautomateddiscovery}. When coupled with an external verifier, this paradigm can unlock powerful discovery capabilities. But in many important domains, verification is too expensive and slow, or simply unavailable. For example, in nuclear fusion research, a single tokamak plasma study may require more than 120 million CPU-hours~\citep{howard2016multiscale}. This motivates our focus on \emph{verifier-free evolution}. However, verifier-free evolution is also expensive. In methods such as RSA, the model may generate 500--700$\times$ more tokens than standard single-shot LLM inference, making the cost of additional search increasingly difficult to sustain.

This cost pressure is compounded by a second tension: models differ sharply in both capability and cost. Proprietary frontier models typically lead on broad, high-stakes benchmarks, while open-weight models offer clear advantages in accessibility, controllability, and marginal cost, especially when self-hosted. Based on listed API prices as of March~16,~2026, representative proprietary reasoning models remain substantially more expensive than strong hosted open-weight alternatives, with output-token costs roughly \(4\times\) to \(25\times\) higher across the providers and models considered here~\citep{openai2026o4mini,anthropic2026pricing,google2026pricing,together2026gptoss120b,together2026qwen235b}. Even within the open-weight ecosystem, cost differences can still be substantial across model families and deployment settings. Together, these two pressures suggest that verifier-free evolution must not only scale compute, but allocate it across models of different cost.

As a result, the key question is shifting: rather than only asking \emph{how we can spend more compute and money to unlock new capabilities?}, we must also ask \emph{how we can achieve a given capability target under tight budget constraints?} This is the same principle that has historically driven advances in software and algorithms: progress comes not just from using more resources, but from using them more efficiently and lowering the cost needed to achieve a given capability target.\footnote{\url{https://epochai.substack.com/p/the-least-understood-driver-of-ai}} In this work, we advance this principle, as illustrated in Figure~\ref{fig:teaser}.

To answer the above question, we first take a system perspective: many seemingly disparate test-time methods can be expressed as instances of a single evolutionary framework. Once cast in this unified form, they expose a common design space that can be optimized jointly. 

In Section~\ref{sec:prelim}, we describe how we unify the current test-time scaling method into a single evolutionary framework, where different operator choices recover a wide spectrum of existing test-time strategies. For example, majority voting \citep{wang2023selfconsistencyimproveschainthought} corresponds to a shallow single-step evolution, recursive self-aggregation \citep{venkatraman2026recursiveselfaggregationunlocksdeep} corresponds to a verifier-free multi-step evolutionary process, and verifier-based self-evolve pipelines such as AlphaEvolve \citep{novikov2025alphaevolvecodingagentscientific} correspond to feedback-driven evolutionary search. 

Our unified framework naturally highlights the key problems:

\begin{enumerate}[leftmargin=*]
    \item Given models with different cost--capability trade-offs, which model should be assigned to each operator in the evolutionary pipeline (e.g., initialization, generation, recombination, or fitness estimation)?
    \item How should these models be coordinated across the pipeline to maximize capability per unit cost without incurring excessive orchestration overhead?
\end{enumerate}

We answer these two questions through a comprehensive empirical analysis in Section~\ref{sec:prelim:limitations}. In brief, we find that: 
\begin{enumerate}[leftmargin=*]
\item \textbf{From the verification perspective,} scaling the token budget can partially offset the absence of explicit verification. By spending additional tokens on diverse generation and iterative aggregation, verifier-free evolution can converge reliably toward correct solutions even without external reward signals. This makes verifier-free evolution especially attractive in practice, as it improves capability while avoiding the substantial cost of explicit verification.

\item \textbf{From the performance perspective,} unlike verifier-based method, simple verifier-free evolution causes the upper bound (\textit{e.g., pass@N or best continuous score}) to degrade significantly. Such a drop directly limits the achievable performance of the entire pipeline. We further find that this upper bound is highly correlated with generation diversity, highlighting diversity as a central ingredient for effective verifier-free evolution. This further strongly motivates our use of \emph{multi-model orchestration} to preserve diversity and sustain performance.

\item \textbf{From the cost perspective,} different models are best suited to different roles, and assigning them accordingly can maximize performance per unit cost. In particular, we find that initialization quality largely determines the quality of the final recombination result, while recombination capability varies substantially across models and depends on the candidate set being aggregated. Furthermore, we show that self-model and cross-model's internal signals can serve as reliable fitness signals in the verifier-free setting. These findings provide a foundation for more principled orchestration design.
\end{enumerate}

Motivated by these observations and by the economic mismatch between open and closed model ecosystems, we present \textsc{Squeeze-Evolve}, a multi-model orchestration framework that routes each evolutionary operation to the most cost-effective model based on confidence-derived fitness signals, reserving expensive models for only the highest-marginal-utility steps. 

We evaluate \OURS{} across AIME~2025, HMMT~2025, GPQA-Diamond, LiveCodeBench~V6, MMMU-Pro, BabyVision, ARC-AGI-V2, and circle packing, spanning open-source model pairs, mixed open-source and proprietary model pairs, and multimodal vision settings. In summary, we make the following contributions:
\begin{enumerate}[leftmargin=*]
\item We unify existing test-time scaling methods into a single evolutionary framework and identify the key design axes for multi-model orchestration (Section~\ref{sec:prelim}). A comprehensive motivation analysis reveals that diversity collapse is the central bottleneck of verifier-free evolution, and that model-intrinsic confidence signals can serve as effective fitness proxies for routing (Section~\ref{sec:prelim:limitations}).

\item We introduce confidence-based routing, a lightweight mechanism that assigns each recombination group to the most cost-effective model using only signals already produced during inference (Section~\ref{sec:method}).

\item Across eight benchmarks spanning math (AIME~2025, HMMT~2025, GPQA-Diamond), coding (LiveCodeBench~V6), vision (MMMU-Pro, BabyVision), visual reasoning (ARC-AGI-V2), and scientific discovery (circle packing), \OURS{} reduces API cost by 1.3--3.3$\times$ while preserving or exceeding single-model accuracy. In multiple configurations, \OURS{} \emph{surpasses} the expensive Model~2 used alone (Section~\ref{sec:evaluation}).

\item On multimodal benchmarks, a text-only cheap model that never processes any images matches or exceeds the expensive vision-capable model at 2.3--2.5$\times$ savings, demonstrating that visual understanding is primarily needed at initialization (Section~\ref{sec:vision-task-evaluation}).

\item On ARC-AGI-V2, \OURS{} achieves 97.5\% accuracy at \$7.74/task without code execution, setting a new state-of-the-art cost-capability frontier (Section~\ref{sec:arc-agi-v2}). On circle packing, it is the first verifier-free evolutionary method to match or exceed verifier-based approaches such as AlphaEvolve (Section~\ref{sec:circle-packing}).

\item We co-design the serving system with the routing algorithm: a custom confidence engine reduces scoring latency by 4--10$\times$, latency-matched GPU pools prevent bottlenecks, and the end-to-end routing overhead is only 2.4--4.3\%, while fixed-budget serving throughput increases by up to $\sim$10$\times$ (Section~\ref{sec:system-impl},~\ref{sec:system-results}).
\end{enumerate}
\section{Related Work}
\label{sec:related_work_short}
Our work builds on four lines of research (extended discussion in Appendix~\ref{sec:related_work}).

\textbf{Test-time scaling and self-aggregation.}
Existing methods improve output quality through parallel sampling~\citep{wang2023selfconsistencyimproveschainthought,brown2024largelanguagemonkeysscaling}, sequential refinement~\citep{madaan2023selfrefineiterativerefinementselffeedback}, search~\citep{yao2023treethoughtsdeliberateproblem}, or extended reasoning chains~\citep{openai2024openaio1card,Guo_2025}.
Self-aggregation methods such as RSA~\citep{venkatraman2026recursiveselfaggregationunlocksdeep} and Mixture-of-Agents~\citep{wang2024mixtureofagentsenhanceslargelanguage} combine multiple LLM outputs into refined answers, but use a single model or fixed assignment, leading to diversity collapse~\citep{singh2026v1unifyinggenerationselfverification}.
\OURS{} extends test-time scaling to multi-model orchestration, preserving diverse reasoning lineages across evolutionary loops.

\textbf{Verification and confidence signals.}
External verification relies on outcome or process reward models~\citep{cobbe2021trainingverifierssolvemath,lightman2023letsverifystepstep} or generative verifiers~\citep{zhang2025generativeverifiersrewardmodeling}; DeepConf~\citep{fu2025deepthinkconfidence} uses token-level confidence to filter traces. \OURS{} repurposes the same confidence class as a zero-cost routing signal rather than a filter.

\textbf{LLM-driven evolutionary search.}
FunSearch~\citep{funsearch}, AlphaEvolve~\citep{novikov2025alphaevolvecodingagentscientific}, and EvoX~\citep{liu2026evoxmetaevolutionautomateddiscovery} use LLMs as evolutionary operators but rely on external verifiers and apply one model across all operators. \OURS{} is verifier-free and introduces adaptive per-group model assignment.

\textbf{Model routing.}
Routing frameworks dispatch queries between models~\citep{ong2025routellmlearningroutellms,maheswaran2025arbitrageefficientreasoningadvantageaware}. \OURS{} routes at recombination-group granularity within a multi-step evolutionary pipeline, where per-group decisions compound across loops.

\section{A Unified Formulation of Evolutionary Framework}
\label{sec:prelim}
Although existing methods differ substantially in their implementation details, we show that many can be naturally cast within a common evolutionary framework. This perspective provides a formal foundation for reasoning about test-time evolution while enabling principled optimization of the framework as a whole. It also suggests an efficient multi-model orchestration strategy based on a simple decision rule: invoke the larger model only when the smaller model is likely to exceed its capability limits.

For a query $Q$, we initialize a population $\mathcal{P}^{(0)}$ using an \emph{ancestor} function $p_F$, where each candidate $\tau_i^{(0)} \sim p_\theta(\cdot \mid Q)$ is sampled from a generative model $\theta$. Existing methods differ primarily in how they organize, score, and evolve these candidates. We unify these steps into a single \emph{evolutionary operator} $\Phi_f$, which encapsulates selection followed by recombination:
\begin{equation}
    \Phi_f(\mathcal{P}) = \mathrm{recomb}_f \circ \mathrm{select}_f (\mathcal{P}),  \quad \mathcal{P}^{(t+1)} = \Phi_{f_{t+1}}(\mathcal{P}^{(t)}).
\end{equation}

This induces an iterated evolutionary process where the final population $\mathcal{P}^{(k)}$ is derived via a sequence of operator compositions:
\begin{equation}
    \mathcal{P}^{(k)} = \left( \Phi_{f_k} \circ \Phi_{f_{k-1}} \circ \dots \circ \Phi_{f_1} \right) (\mathcal{P}^{(0)}),
\end{equation}
where each operator $\Phi_{f_i}$ utilizes the fitness signal $f_i$ to transition between generations. In verifier-free evolution, the fitness signal is derived entirely from the models' own outputs (e.g., log-probabilities, consensus frequency) without access to an external verifier or ground-truth reward. Let $f$ denote a \emph{fitness signal}: a function that maps a set of candidate trajectories to quality estimates. $f$ may be implicit (e.g., consensus frequency in majority voting) or explicit (e.g., cross-model log-probabilities in our method).  This unified formulation provides a lens for categorizing existing test-time scaling methods based on how they instantiate the $\mathrm{select}$, $\mathrm{recomb}$ operators and fitness signal $f$, as shown in Tab. \ref{tab:framework-summary}. 

In detail, majority voting (self-consistency) is a degenerate single-step process that generates a population once and selects the largest answer cluster using consensus frequency as an implicit fitness signal. Self-refinement is a multi-step process with a population size of one, where selection reduces to self-evaluation and recombination produces an improved trajectory conditioned on critique. Recursive self-aggregation (RSA) is a multi-step process that repeatedly samples subsets of the current population and applies the model's aggregation operator to synthesize refined candidates, relying entirely on implicit model-internal fitness. AlphaEvolve uses explicit external verifier, where candidate programs are evaluated and the resulting scalar rewards guide future search.  \OURS{} builds on this view but departs from the single-model paradigm in two ways: it uses token log-probabilities already produced during generation as essentially zero-cost self or cross-model confidence signals, and it routes each evolutionary step to either an expensive or cheap model. This enables cost-efficient orchestration without sacrificing accuracy.
\begin{table}[t]
\begin{center}
\small
\setlength{\tabcolsep}{4pt}
\caption{Instantiation of the unified evolutionary framework for test-time scaling methods.}
\label{tab:framework-summary}
\resizebox{\linewidth}{!}{%
\begin{tabular}{l c c l l l l}
\toprule
Method & $k$ & $|\mathcal{P}|$ & $\mathrm{select}$ & $\mathrm{recombination}$ & Fitness $f$ & Model \\
\midrule
Majority Voting & 1 & $N$ & Answer clustering & Identity & Consensus frequency & Single \\
Self-Refinement & $T$ & 1 & Self-critique & Conditioned rewrite & Natural language critique & Single \\
RSA & $T$ & $N$ & $K$-subset & LLM aggregation & Implicit & Single \\
AlphaEvolve & $T$ & Variable & Fitness-guided & LLM aggregation & External $h$ & Multi-model \\
\midrule
\textbf{\OURS} & $T$ & Variable & Fitness-guided & Mix of recombination & Probabilistic fitness & Multi-model \\
\bottomrule
\end{tabular}%
}
\end{center}
\end{table}

\section{Motivation analysis for verifier-free evolutionary framework}
\label{sec:prelim:limitations}
\begin{figure*}[t]
\centering
\begin{minipage}{0.49\linewidth}
\centering
\includegraphics[width=\linewidth]{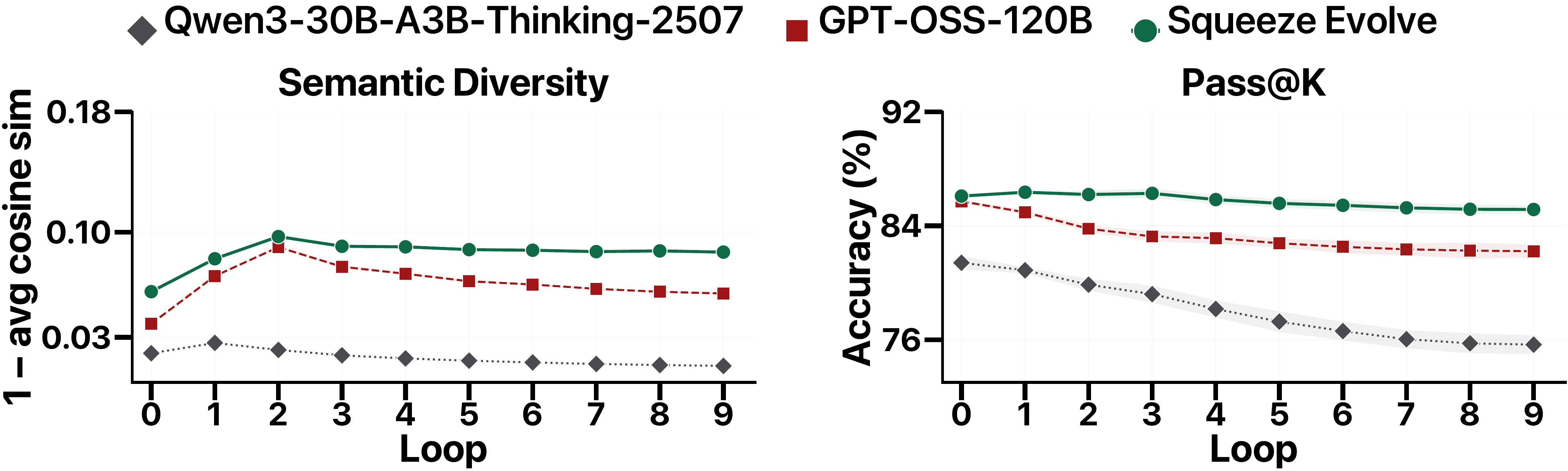}
\end{minipage}
\hfill
\begin{minipage}{0.49\linewidth}
\centering
\includegraphics[width=\linewidth]{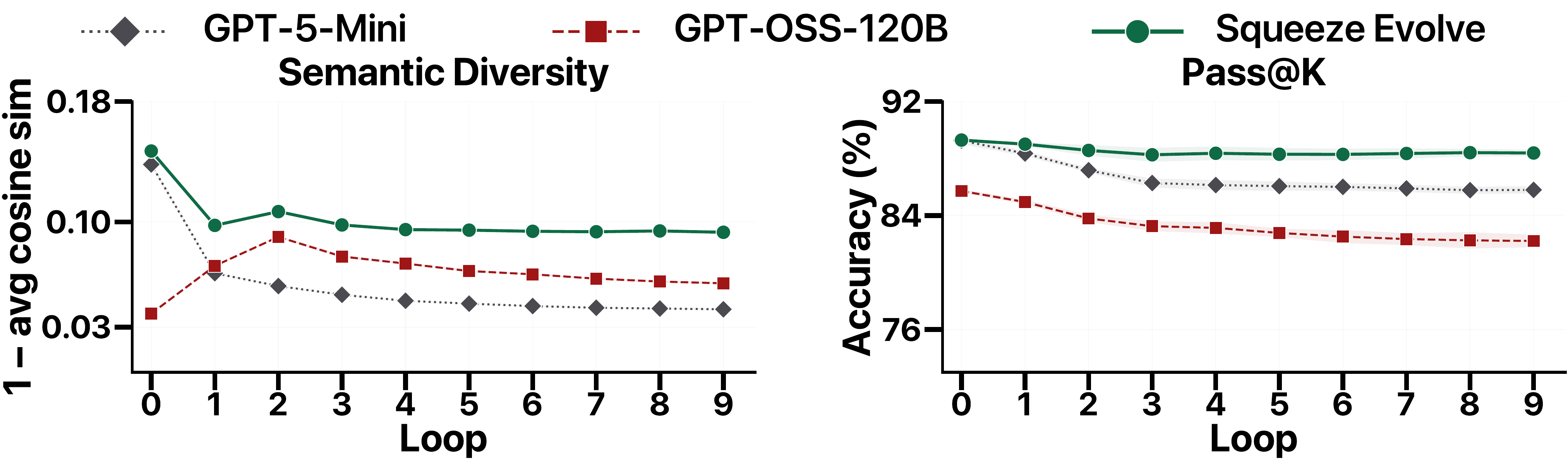}
\end{minipage}
 \caption{\textbf{Single-model open-loop evolution collapses diversity and shrinks the population's pass@$K$ ceiling, while multi-model routing preserves both.}
  GPQA-Diamond results across RSA loops in two comparison settings.
  In both settings, the single-model baselines lose diversity after the early loops and show a corresponding decline in pass@$K$, whereas \OURS{} remains higher and flatter on both metrics. Shaded bands show variation across seeds.}
  \label{fig:diversity-collapse-gpqa}
\end{figure*}
\paragraph{The inherent Pass@K bottleneck of verifier-free evolution.}
In this section, we identify a fundamental bottleneck in verifier-free evolution: without an external verifier, the loop can only amplify trajectories that the current model already knows how to recognize and reproduce. This drives the population toward an increasingly narrow solution mode, causing pass@$K$ to fall along with semantic diversity, as shown in Figure~\ref{fig:diversity-collapse-gpqa} across both GPQA-Diamond~\citep{rein2024gpqa} settings. This failure mode reveals that preserving diversity is necessary for maintaining the population's upper-bound search capacity. This is precisely where multi-model orchestration helps. By introducing models with different priors, failure modes, and reasoning styles, \OURS{} maintains complementary lineages and remains higher and flatter on both diversity and pass@$K$.

\begin{table*}[t]
\begin{center}
\small
\setlength{\tabcolsep}{3pt}
\caption{\textbf{Ancestor function dominates final accuracy.}
Mean final loop (9) accuracy across 4 seeds. Strong-init $\rightarrow$ weak-agg outperforms weak-init $\rightarrow$ strong-agg, indicating initialization quality dominates aggregation quality.}
\label{tab:loop0-summary}
\begin{tabular}{p{0.6\linewidth} l c c c}
\toprule
Model pair & Data & S$\rightarrow$W & W$\rightarrow$S & $\Delta$ \\
\midrule
GPT-OSS-120B / GPT-OSS-20B & HMMT'25 & 0.89 & 0.85 & +4 \\
Qwen3-4B-Thinking-2507 / Qwen3-4B-Instruct-2507 & AIME'25 & 0.88 & 0.65 & +23 \\
\bottomrule
\end{tabular}
\end{center}
\end{table*}

\paragraph{Ancestor function dominates final accuracy.} Results on HMMT~25~\citep{balunović2026matharenaevaluatingllmsuncontaminated} show that using \text{GPT-OSS-120B}~\citep{openai2025gptoss} as the ancestor function and \text{GPT-OSS-20B} for recombination achieves 89\% accuracy, whereas reversing their roles reduces performance to 85\%. The gap becomes much larger on AIME~2025~\citep{balunović2026matharenaevaluatingllmsuncontaminated}: using \text{Qwen3-4B-Thinking}~\citep{qwen3technicalreport} as the ancestor function and \text{Qwen3-4B-Instruct} for recombination reaches 88\%, while the reverse achieves only 65\%, a drop of 23 percentage points (Table~\ref{tab:loop0-summary}). This asymmetry suggests to use the strong model for initialization.

\begin{figure*}[t]
\centering
\begin{subfigure}[t]{0.48\linewidth}
\centering
\includegraphics[width=\linewidth]{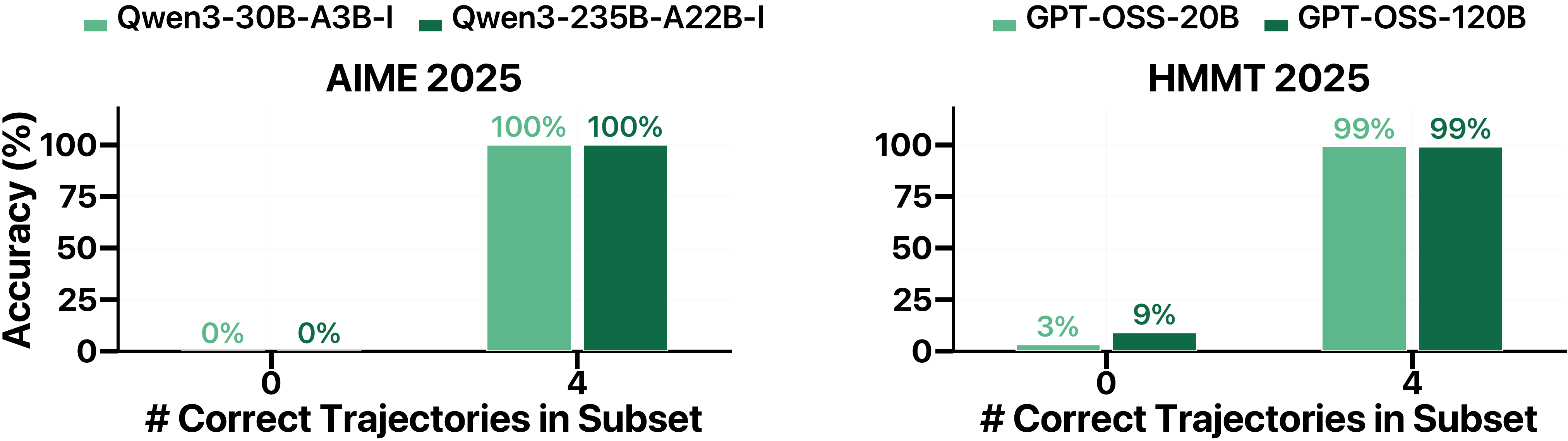}
\caption{Aggregation accuracy at the extremes: 0 vs.\ 4 correct trajectories.}
\label{fig:aggregatable-accuracy}
\end{subfigure}\hfill
\begin{subfigure}[t]{0.48\linewidth}
\centering
\includegraphics[width=\linewidth]{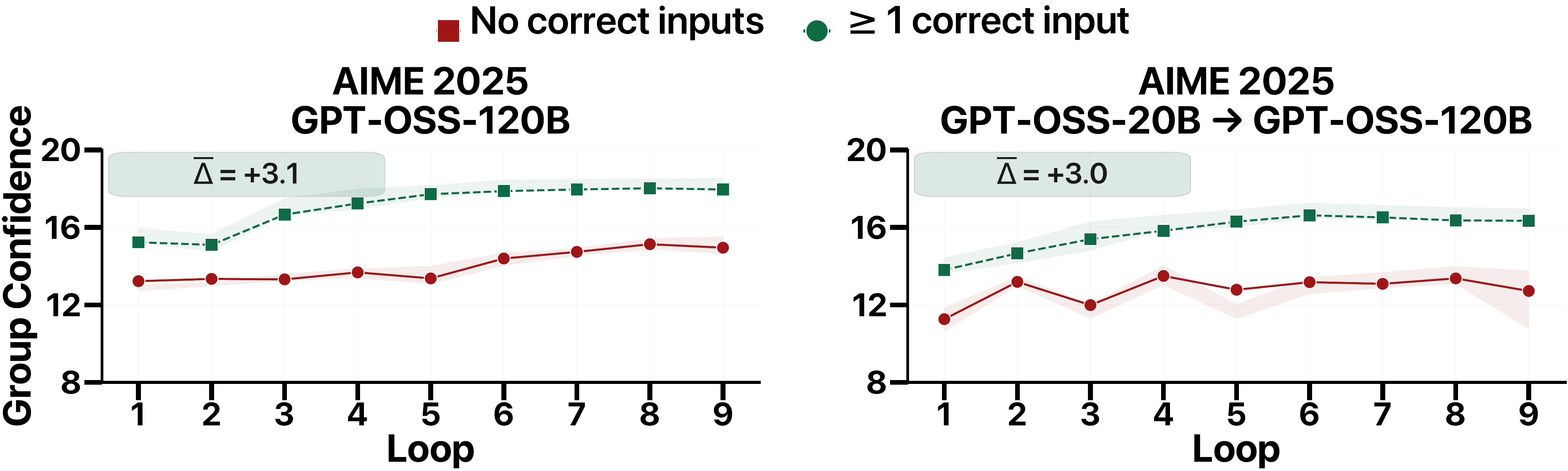}
\caption{Group confidence separates correct from incorrect trajectories.}
\label{fig:gc-confidence}
\label{fig:gc-across-loops}
\label{fig:replay-confidence-gc}
\end{subfigure}
\caption{\textbf{Aggregation success is seed-dependent, and group confidence predicts it.}
\textbf{(a)}~With zero correct seeds, neither model recovers a correct answer; with all seeds correct, both achieve near-perfect accuracy.
Full results across all seed counts (0--4) in Appendix Figure~\ref{fig:agg-accuracy-full}.
\textbf{(b)}~Mean group confidence (GC) across RSA loops~1--9 on AIME~2025, split by whether the subset contains $\geq$1 correct trajectory.
In both self-model and cross-model settings, correct-containing subsets maintain consistently higher GC ($\overline{\Delta} \geq +3.0$).
Full results in Appendix Figures~\ref{fig:gc-baseline-full} and~\ref{fig:gc-routing-full}.}
\end{figure*}

\paragraph{Weak models can also be strong aggregators when the candidate set is strong.} It is not a surprising conclusion, but Figure~\ref{fig:aggregatable-accuracy} makes it explicit: recombination quality depends strongly on the correctness of candidates. On AIME~2025 with Qwen3, aggregation accuracy rises from 0\% when no correct candidate is present and reaches 100\% when all four candidates are correct. The same trend appears on HMMT~2025 with GPT-OSS: accuracy is only 3--9\% when no correct seed is present and reaches 99\% when all four are correct. This observation motivates a key routing strategy: if we can identify subsets with sufficiently strong candidates, we can route them to a cheaper model for aggregation.

\paragraph{Self- and cross-model confidence serve as effective proxies for fitness estimation.} We show that both self- and cross-model confidence closely track the correctness of the population. As shown in Figure~\ref{fig:gc-confidence}, both self-model and cross-model confidence provide a strong proxy for subset quality: high-confidence subsets are substantially more likely to contain correct trajectories and to aggregate successfully. This motivates us to use the confidence for the fitness estimation for the router.
\section{Squeeze Evolve}
\label{sec:method}
\begin{figure}[t]
    \centering
    \includegraphics[width=0.9\linewidth]{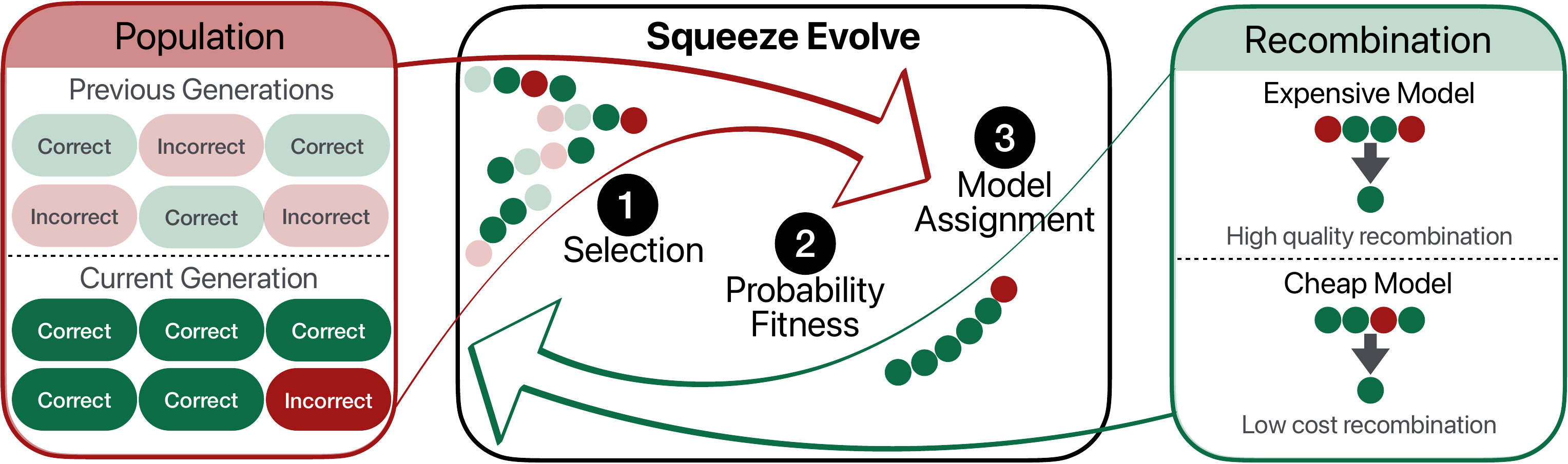}
    \caption{\OURS{} overview. The expensive Model 2 generates the initial population; subsequent loops recombine groups using Model 1 and 2 based on group confidence}
    \label{fig:verifier_free_method}
\end{figure}

Building on the findings of Section~\ref{sec:prelim:limitations}, we instantiate the evolutionary operator $\Phi_f = \mathrm{recomb}_f \circ \mathrm{select}_f$ from Section~\ref{sec:prelim} as a single algorithm (Figure~\ref{fig:verifier_free_method}; Algorithm~\ref{alg:squeeze-evolve}, Appendix~\ref{app:algorithm}). Our key extension is to the recombination operator: a routing function assigns each candidate group to one of $L+1$ tiers based on the fitness signal: $L$ models ordered by increasing cost, plus a lightweight non-LLM aggregation tier. In our experiments we use $L = 2$. The population update rule is also generalized to support accumulation across generations. Operator settings are listed in Table~\ref{tab:instantiations}.

\subsection{Algorithm}
\label{sec:method-formulation}

Each loop scores candidates via the fitness signal $f$, applies $\mathrm{select}_f$ to form groups, routes each group to one of three recombination tiers within $\mathrm{recomb}_f$, and updates the population. We define each component below.

\paragraph{Initialization.}
We initialize the population by sampling all $N$ candidates from the strongest model, which is typically also the most expensive:
\[
\mathcal{P}_q^{(0)} = \{\tau_i \sim p_{M_2}(\cdot \mid Q_q)\}_{i=1}^{N}.
\]
This choice is motivated by our empirical finding that initialization quality is the strongest predictor of final accuracy (Table~\ref{tab:loop0-summary}).

\paragraph{Fitness signal.}
The fitness function $f$ maps each candidate trajectory to a scalar that measures the model's certainty about that trajectory. \OURS uses two model-intrinsic realizations of $f$, both of which serve as proxies for \emph{group difficulty}: they identify groups where candidates are uncertain or conflicting, precisely the regime in which the stronger model (Model~2) provides the greatest marginal value.

\emph{Group confidence} (GC) derives $f$ from the top-$K_\ell$ token log-probabilities already produced during inference. For each token position~$i$ in a trajectory $\tau$, we compute:
\begin{equation}
    c(i) \;=\; -\frac{1}{K_\ell}\sum_{j=1}^{K_\ell} \log p_\theta\!\bigl(v_j^{(i)} \mid t_{<i},\, Q\bigr),
    \label{eq:per-token-confidence}
\end{equation}
where $\{v_1^{(i)}, \ldots, v_{K_\ell}^{(i)}\}$ are the $K_\ell$ most likely tokens under a scoring model~$\theta$. When the predictive distribution is peaked, the top-$K_\ell$ entries are dominated by a few high-probability tokens and $c(i)$ is large; when the distribution is flat, $c(i)$ is small. The candidate-level and group-level confidences are:
\begin{equation}
    C(\tau) = \frac{1}{|\tau|}\sum_{i=1}^{|\tau|} c(i), \qquad
    \mathrm{GC}(g) = \frac{1}{K}\sum_{\tau \in g} C(\tau).
    \label{eq:confidence}
\end{equation}
The per-token confidence $c(i)$ follows the same formulation used by DeepConf~\citep{fu2025deepthinkconfidence} to filter reasoning traces; here we aggregate it to the group level for routing. 
When the scoring model $\theta$ is the generating model itself, this yields \emph{self-confidence} at zero additional cost. When $\theta$ differs from the generator, this is \emph{cross-model confidence} and requires a single prefill-only forward pass per candidate, whose cost we minimize via the custom confidence engine described in Section~\ref{sec:system-impl}.

\emph{Group diversity} provides an equivalent signal when token log-probabilities are unavailable (e.g., APIs that do not expose prefill-only scoring):
\begin{equation}
    D(g) \;=\; \bigl|\{\mathrm{answer}(\tau) : \tau \in g\}\bigr|,
    \label{eq:diversity}
\end{equation}
the number of unique final answers in the group. In principle, diversity can be measured in richer ways (e.g., embedding similarity between trajectories), but we find that this simplest instantiation is already effective. Diversity requires only answer extraction, not token-level scoring. Low GC and high $D$ both indicate that the group's candidates are uncertain or conflicting; in this sense the two signals are complementary views of the same underlying quantity, and the choice between them is determined entirely by API access.

\paragraph{Selection.}
At each loop $t \geq 1$, we form $M$ groups of size $K$ from the current population. Groups can be formed by \emph{uniform sampling} (random $K$-subsets, as in RSA or by \emph{fitness-weighted sampling}, where candidates are drawn with probability $\exp\!\bigl(f(\tau_i)/\zeta\bigr)\big/\sum_j \exp\!\bigl(f(\tau_j)/\zeta\bigr)$ and a temperature $\zeta$ controls the exploitation--exploration balance.

\paragraph{Recombination.}
Based on the group fitness $F(g)$, the routing function assigns each group to one of three recombination strategies of decreasing cost: $\mathcal{B}_2$ (recombined by the more expensive Model 2), $\mathcal{B}_1$ (recombined by Model 1), and $\mathcal{B}_{\mathrm{lite}}$ (aggregated via a lightweight non-LLM method, e.g., majority vote or random sampling from the group). Groups whose fitness indicates sufficient consensus are routed to $\mathcal{B}_{\mathrm{lite}}$, since LLM recombination would add cost with little marginal benefit. Among the remaining groups, we compute a per-problem adaptive threshold at the $p$-th percentile of the fitness distribution:
\begin{equation}
    \theta_q \;=\; \mathrm{Percentile}_{p}\!\bigl(\{F(g) : g \in \mathcal{G}_q \setminus \mathcal{B}_{\mathrm{lite}}\}\bigr).
    \label{eq:threshold}
\end{equation}
Each non-lite group is then assigned to a model:
\begin{equation}
    \mu(g,\, q) \;=\;
    \begin{cases}
        \text{Model~1} & \text{if the group is ``easy'' under } F, \\
        \text{Model~2} & \text{otherwise},
    \end{cases}
    \label{eq:routing}
\end{equation}
where ``easy'' means high confidence ($\mathrm{GC}(g) > \theta_q$) or low diversity ($D(g) < \delta$), depending on which fitness signal is used. Computing $\theta_q$ independently per problem adapts the threshold to each problem's difficulty: hard problems naturally produce lower fitness scores, yet the routing fraction remains approximately $p/100$ regardless. The routing percentile~$p$ is the single hyperparameter practitioners tune at deployment time.
Each model-routed group is recombined via LLM aggregation: the assigned model receives the group's $K$ candidate trajectories as context and generates a single refined trajectory. Because $M_1$ and $M_2$ may use different tokenizers and chat templates, prompts are built per model, and the two batches are executed in parallel. The resulting trajectories from all three tiers are merged back into the population via one of two rules: \emph{replace} discards the previous population entirely, while \emph{accumulate} retains it ($\mathcal{P}_q^{(t)} = \mathcal{P}_q^{(t-1)} \cup \mathcal{R}_{\mathrm{new}}$), preserving high-quality solutions discovered in earlier generations.

\begin{table}[t]
\centering
\small
\setlength{\tabcolsep}{3pt}
\caption{Operator instantiations of \OURS across evaluation settings.}
\label{tab:instantiations}
\resizebox{\linewidth}{!}{%
\begin{tabular}{@{}l l l l l@{}}
\toprule
Setting & Fitness $f$ & $\mathrm{Select}_f$ & $\mathrm{Recomb}_f$ (routing rule) & $\mathrm{Update}$ \\
\midrule
Math / Coding / Vision & GC (Eq.~\ref{eq:confidence}) & Uniform & Percentile on GC & Replace \\
ARC-AGI-V2 (\S\ref{sec:arc-agi-v2}) & Diversity $D$ (Eq.~\ref{eq:diversity}) & Uniform & Threshold on $D$ + lite agg & Replace \\
Circle Packing (\S\ref{sec:circle-packing}) & GC (Eq.~\ref{eq:confidence}) & Fitness-weighted ($\zeta{=}0.5$) & Percentile on GC & Accumulate \\
\bottomrule
\end{tabular}%
}
\end{table}

\subsection{System Implementation}
\label{sec:system-impl}

Routing alone is not enough for practical gains; the deployment must be co-designed with both the scoring mechanism and the serving infrastructure.

\paragraph{Latency-matched serving.}
\OURS{} serves Model~1 and Model~2 in separate GPU pools that are sized so that both pools complete their assigned work in approximately the same wall-clock time per loop. If either pool is substantially faster than the other, the faster pool idles while the slower pool becomes the throughput bottleneck, negating the benefit of routing. Given a routing percentile $p$ and its observed traffic split, we choose integer GPU allocations $G_1 + G_2 = G$ that minimize the gap between the two pools' per-loop service times. We evaluate the resulting throughput gains in Section~\ref{sec:system-results}.

\paragraph{Confidence scoring.}
We use two forms of confidence. \emph{Self-confidence} is essentially free: during generation, the model already produces the token log-probabilities needed to score its own trajectory, so no additional inference is required. \emph{Cross-model confidence} scores a trajectory under a different model from the one that generated it. This requires only a single forward pass per trajectory, with no autoregressive decoding. As a result, cross-model scoring is a prefill-only operation whose cost scales linearly with sequence length.

Importantly, this scoring path fits naturally into our routing pipeline. The scoring model is already resident for the corresponding aggregation branch, so confidence computation does not introduce additional model loading or memory residency overhead. In practice, the $N$ scoring calls in each loop are batched into a single request, so the added latency remains modest relative to the generation stages that dominate end-to-end wall-clock time. We report the resulting routing overhead in the full pipeline in Section~\ref{sec:system-results}.

\paragraph{Confidence engine.}
Standard serving systems~\citep{kwon2023efficient,zheng2024sglangefficientexecutionstructured} are optimized for decode-heavy generation, but cross-model confidence scoring is prefill-only and needs just one scalar per trajectory. To avoid materializing full token-level logprob tensors, we implement a custom prefill path in vLLM that accumulates the confidence statistic directly on GPU and returns only the final scalar, reducing per-request transfer from ${\sim}$13\,MB to ${\sim}$100\,bytes. This achieves $4$--$10{\times}$ lower scoring latency and enables confidence scoring on Qwen3-235B-A22B where the native path runs out of memory (Appendix~\ref{app:prefill-engine}). We quantify end-to-end routing overhead and system throughput in Section~\ref{sec:system-results}.

\section{Evaluation}
\label{sec:evaluation}
All runs use population $N{=}16$, group size $K{=}4$, and $T{=}10$ evolutionary loops, averaged over four seeds, unless stated otherwise.
Costs are measured in actual API dollars per problem using model provider pricing (Table~\ref{tab:pricing}; generation hyperparameters in Table~\ref{tab:hyperparams}, Appendix). The baseline is standard RSA with Model 2 only, which serves as the cost upper bound.

\subsection{Math and Coding}
\label{sec:math-coding-evaluation}
We evaluate \OURS on reasoning benchmarks: AIME 2025, HMMT 2025~\citep{balunović2026matharenaevaluatingllmsuncontaminated}, GPQA-Diamond~\citep{rein2024gpqa} as well as coding benchmark: LiveCodeBench V6~\citep{jain2024livecodebenchholisticcontaminationfree}. Full per-percentile cost breakdowns appear in Tables~\ref{tab:empirical} and~\ref{tab:empirical-hetero} (Appendix).

% -- Highlight results table (main paper) --------------------------------------
{\small
\setlength{\tabcolsep}{3pt}
\renewcommand{\arraystretch}{1.0}
\begin{xltabular}{\textwidth}{@{}l l p{0.20\textwidth} p{0.20\textwidth} c c c@{}}
\caption{Representative results for math and coding benchmarks.
  Each group shows the RSA baseline alongside a
  representative \OURS operating point for that dataset.
  Model name suffixes: \textbf{-I}\,=\,Instruct, \textbf{-T}\,=\,Thinking.
  Full per-percentile breakdowns appear in Tables~\ref{tab:empirical}
  and~\ref{tab:empirical-hetero} (Appendix).}
\label{tab:math-coding-highlights} \\
\toprule
\textbf{Data} & \textbf{Strategy} & \textbf{Model 1} & \textbf{Model 2}
& \textbf{Acc.} & \textbf{\$/Prob} & \textbf{\$ Savings} \\
\midrule
\endfirsthead
\toprule
\textbf{Data} & \textbf{Strategy} & \textbf{Model 1} & \textbf{Model 2}
& \textbf{Acc.} & \textbf{\$/Prob} & \textbf{\$ Savings} \\
\midrule
\endhead
\bottomrule
\endlastfoot

% ========================= Homogeneous =========================
\multicolumn{7}{@{}l}{\textit{Homogeneous (open-source + open-source)}} \\
\midrule
\multirow{2}{*}{AIME25}
  & RSA & --- & Qwen3-30B-A3B-T & 89.2 & \$0.94 & 1.0$\times$ \\
  & \cellcolor{teal!4}\OURS ($p{=}0$)  & \cellcolor{teal!4}Qwen3-30B-A3B-I & \cellcolor{teal!4}Qwen3-30B-A3B-T & \cellcolor{teal!4}90.7 & \cellcolor{teal!4}\$0.66 & \cellcolor{teal!4}1.4$\times$ \\
\midrule
\multirow{2}{*}{HMMT25}
  & RSA & --- & GPT-OSS-120B & 89.7 & \$0.41 & 1.0$\times$ \\
  & \cellcolor{teal!4}\OURS ($p{=}10$) & \cellcolor{teal!4}GPT-OSS-20B & \cellcolor{teal!4}GPT-OSS-120B & \cellcolor{teal!4}92.0 & \cellcolor{teal!4}\$0.25 & \cellcolor{teal!4}1.6$\times$ \\
\midrule
\multirow{2}{*}{GPQA-D}
  & RSA & --- & Qwen3-30B-A3B-T & 74.0 & \$0.57 & 1.0$\times$ \\
  & \cellcolor{teal!4}\OURS ($p{=}0$)  & \cellcolor{teal!4}Qwen3-30B-A3B-I & \cellcolor{teal!4}Qwen3-30B-A3B-T & \cellcolor{teal!4}75.9 & \cellcolor{teal!4}\$0.32 & \cellcolor{teal!4}1.8$\times$ \\
\midrule
\multirow{2}{*}{LCB-V6}
  & RSA & --- & GPT-OSS-120B & 75.9 & \$0.44 & 1.0$\times$ \\
  & \cellcolor{teal!4}\OURS ($p{=}10$) & \cellcolor{teal!4}GPT-OSS-20B & \cellcolor{teal!4}GPT-OSS-120B & \cellcolor{teal!4}75.6 & \cellcolor{teal!4}\$0.22 & \cellcolor{teal!4}2.0$\times$ \\
\midrule

% ========================= Heterogeneous =========================
\multicolumn{7}{@{}l}{\textit{Heterogeneous (open-source + closed-source)}} \\
\midrule
\multirow{2}{*}{AIME25}
  & RSA & --- & GPT-5 mini & 94.2 & \$0.89 & 1.0$\times$ \\
  & \cellcolor{teal!4}\OURS ($p{=}30$) & \cellcolor{teal!4}GPT-OSS-20B & \cellcolor{teal!4}GPT-5 mini & \cellcolor{teal!4}95.4 & \cellcolor{teal!4}\$0.50 & \cellcolor{teal!4}1.8$\times$ \\
\midrule
\multirow{2}{*}{HMMT25}
  & RSA & --- & GPT-5 mini & 93.3 & \$0.94 & 1.0$\times$ \\
  & \cellcolor{teal!4}\OURS ($p{=}30$) & \cellcolor{teal!4}GPT-OSS-20B & \cellcolor{teal!4}GPT-5 mini & \cellcolor{teal!4}93.1 & \cellcolor{teal!4}\$0.56 & \cellcolor{teal!4}1.7$\times$ \\
\midrule
\multirow{2}{*}{GPQA-D}
  & RSA & --- & GPT-5 mini & 85.0 & \$0.52 & 1.0$\times$ \\
  & \cellcolor{teal!4}\OURS ($p{=}20$) & \cellcolor{teal!4}Qwen3-30B-A3B-I & \cellcolor{teal!4}GPT-5 mini & \cellcolor{teal!4}83.6 & \cellcolor{teal!4}\$0.35 & \cellcolor{teal!4}1.5$\times$ \\

\end{xltabular}
}

\begin{figure*}[t]
\centering
\begin{subfigure}[t]{\textwidth}
\centering
\includegraphics[width=\linewidth]{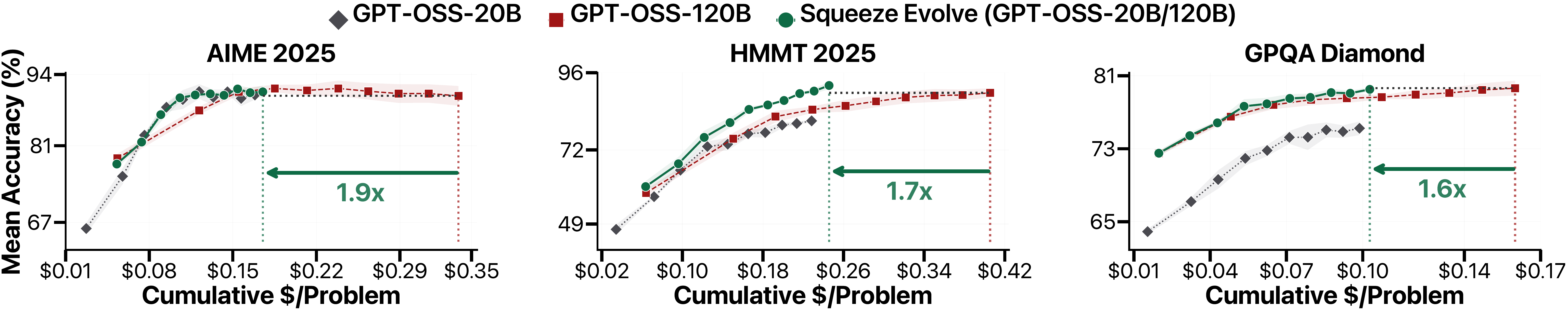}
\caption{}
\label{fig:acc-cost-homo-gptoss}
\end{subfigure}\hfill
\begin{subfigure}[t]{\textwidth}
\centering
\includegraphics[width=\linewidth]{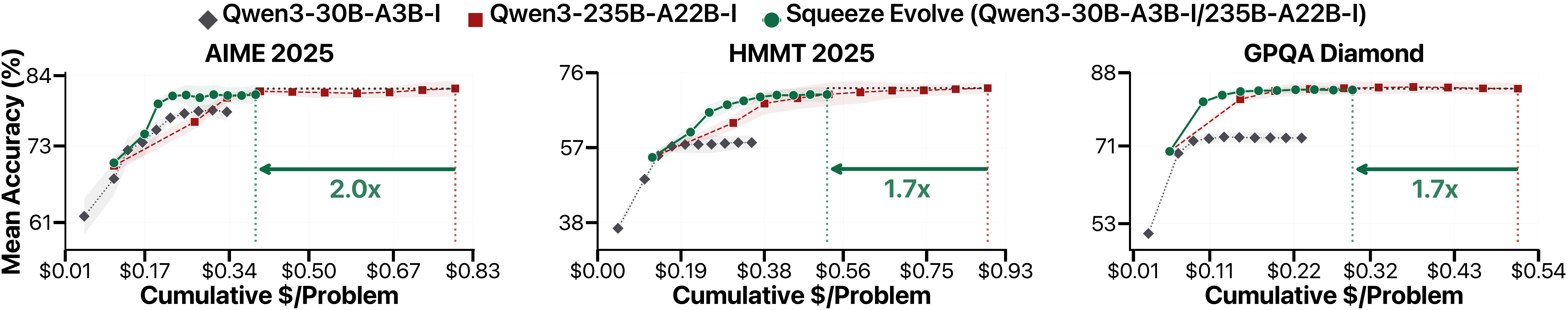}
\caption{}
\label{fig:acc-cost-homo-qwen-235b}
\end{subfigure}\hfill
\begin{subfigure}[t]{\textwidth}
\centering
\includegraphics[width=\linewidth]{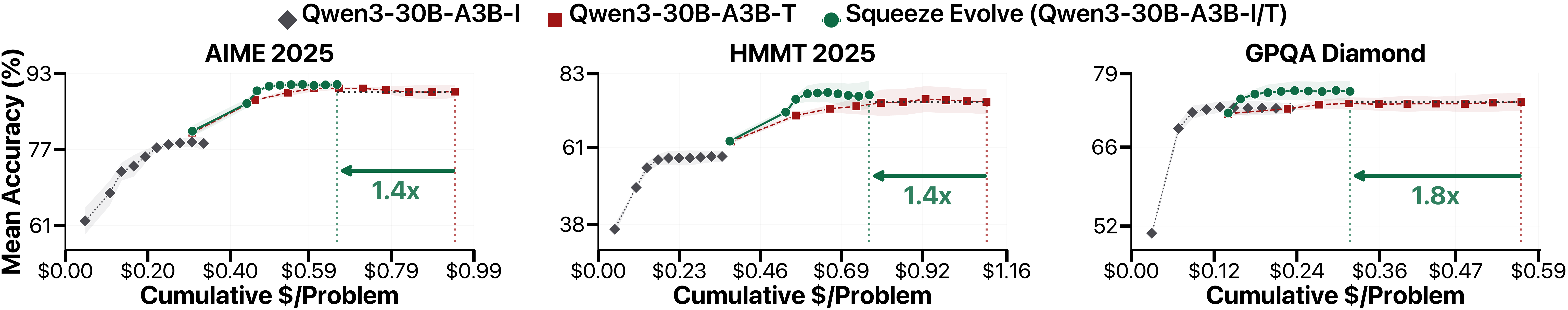}
\caption{}
\label{fig:acc-cost-homo-qwen-thinking}
\end{subfigure}
\caption{\textbf{Accuracy vs.\ cumulative cost for homogeneous model pairs.}
Each point corresponds to one RSA loop (0--9). \OURS{} (green) tracks the RSA accuracy curve while staying significantly further left, achieving comparable quality at 1.4--2.0$\times$ lower cost.}
\label{fig:acc-vs-cost-homogeneous}
\end{figure*}

\begin{figure*}[t]
\centering
\begin{subfigure}[t]{\textwidth}
\centering
\includegraphics[width=\linewidth]{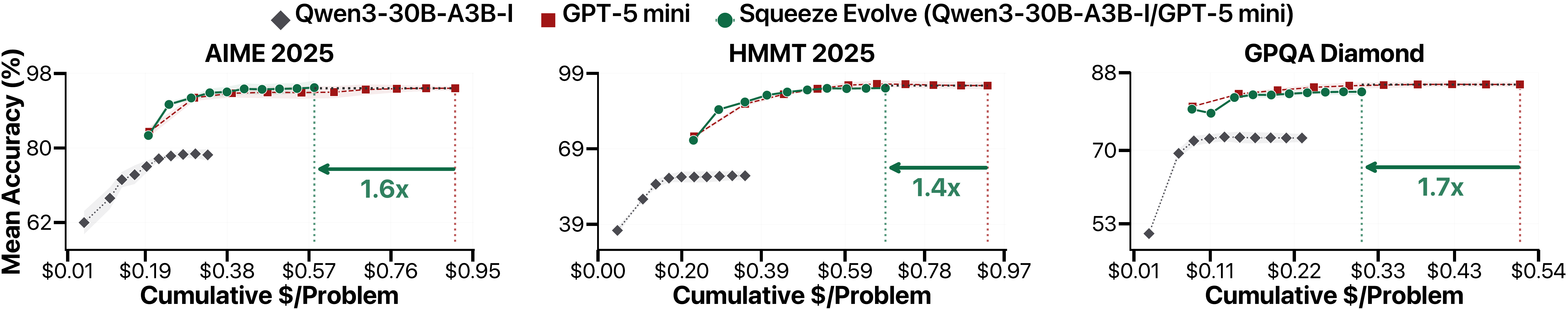}
\caption{}
\label{fig:acc-cost-hetero-qwen-gpt5mini}
\end{subfigure}\hfill
\begin{subfigure}[t]{\textwidth}
\centering
\includegraphics[width=\linewidth]{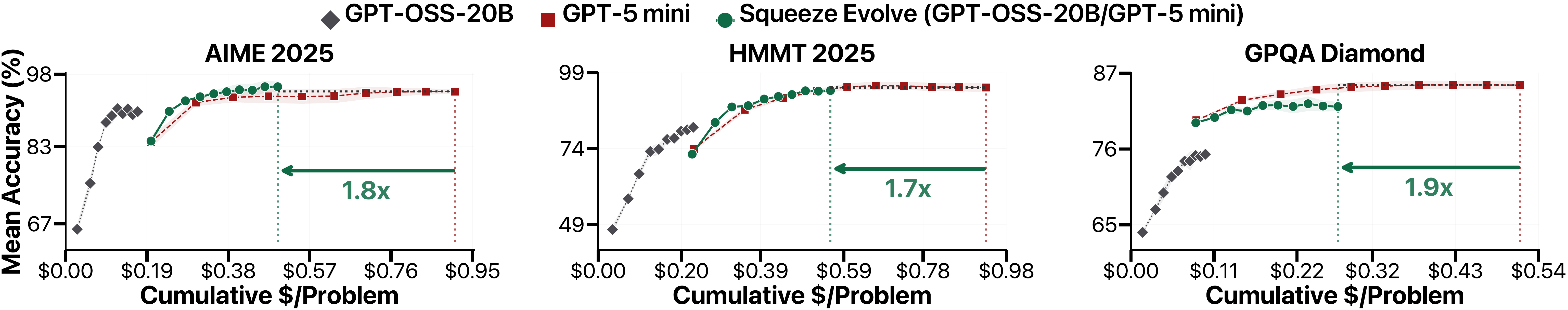}
\caption{}
\label{fig:acc-cost-hetero-gptoss-gpt5mini}
\end{subfigure}
\caption{\textbf{Accuracy vs.\ cumulative cost for heterogeneous model pairs.}
\OURS{} (green) matches the expensive curve at 1.4--1.9$\times$ lower cost, demonstrating that confidence-based routing generalizes across model families and access types.}
\label{fig:acc-vs-cost-heterogeneous}
\end{figure*}

Table~\ref{tab:math-coding-highlights} summarizes representative results across benchmarks; accuracy-vs-cost curves appear in Figures~\ref{fig:acc-vs-cost-homogeneous} and~\ref{fig:acc-vs-cost-heterogeneous}. Notably, no single pair dominates all benchmarks: Qwen3-30B Instruct$\to$Thinking leads on AIME25 and GPQA-Diamond, while GPT-OSS-20B$\to$120B leads on HMMT25 and LiveCodeBench. This demonstrates the generality of \OURS{} across model families and pairing types, and reflects its model-agnostic design: practitioners can select the pair that suits their specific task.

\noindent\textbf{Homogeneous pairs (open-source + open-source).}
We test three open-source pairs that span different axes of the model-pair design space: Qwen3-30B Instruct\,/\,Thinking (same scale, different reasoning mode), Qwen3-30B\,/\,235B Instruct (different scale, same mode), and GPT-OSS-20B\,/\,120B (different scale, both thinking).

Across all three, \OURS matches or exceeds the accuracy of Model~2 alone while costing \textbf{1.4--2.1$\times$ less}. In two of the three pairs, \OURS actually \emph{surpasses} Model~2: by 1.5 points on AIME25 for Instruct$\to$Thinking and by 2.3 points on HMMT25 for GPT-OSS. Even when Model~1 is much smaller (Qwen3-30B vs.\ 235B), accuracy stays within 1 point while cost is nearly halved. The pattern extends to code generation, where the GPT-OSS pair matches Model~2 on LiveCodeBench~V6 at 2.0$\times$ savings.

\noindent\textbf{Heterogeneous pairs (open-source + closed-source).}
We pair two open-source Model~1s (Qwen3-30B Instruct and GPT-OSS-20B) with GPT-5 mini~\citep{openai2025gpt5} as Model~2, sweeping $p \in \{0, 10, 20, 30\}$ (Model~1 scores candidates via prefill since GPT-5 mini does not expose output logprobs; this cost is included in all figures).

\OURS achieves \textbf{1.4--3.3$\times$ savings} depending on routing aggressiveness. At conservative settings ($p{=}30$), GPT-OSS-20B paired with GPT-5 mini \emph{exceeds} Model~2 alone on AIME25 (95.4\% vs.\ 94.2\%) at 1.8$\times$ savings. At the most aggressive setting ($p{=}0$), savings reach 3$\times$ with accuracy drops of only 1.5--6 points (Table~\ref{tab:empirical-hetero}).

\noindent Across all five model-pair configurations, \OURS reduces cost by 1.3--3.3$\times$ while preserving accuracy. The routing percentile $p$ acts as a single deployment knob that smoothly trades accuracy for cost.

% -------------------------------------------------

\subsection{Multimodal Vision Task}
\label{sec:vision-task-evaluation}
We evaluate \OURS on two multimodal benchmarks: MMMU-Pro~\citep{yue2025mmmuprorobustmultidisciplinemultimodal} and BabyVision~\citep{chen2026babyvisionvisualreasoninglanguage}, using $T{=}5$ loops (other settings match Section~\ref{sec:math-coding-evaluation}). We test a homogeneous pair (Kimi-2.5 Instant\,/\,Thinking~\citep{kimiteam2026kimik25visualagentic}, both vision-capable) and a heterogeneous pair (Qwen3.5-35B~\citep{qwen3.5}\,$\to$\,Kimi-2.5 Thinking, where Model~1 operates in \emph{text-only mode} after loop~0).

\begin{figure*}[t]
\centering
\includegraphics[width=0.66\linewidth]{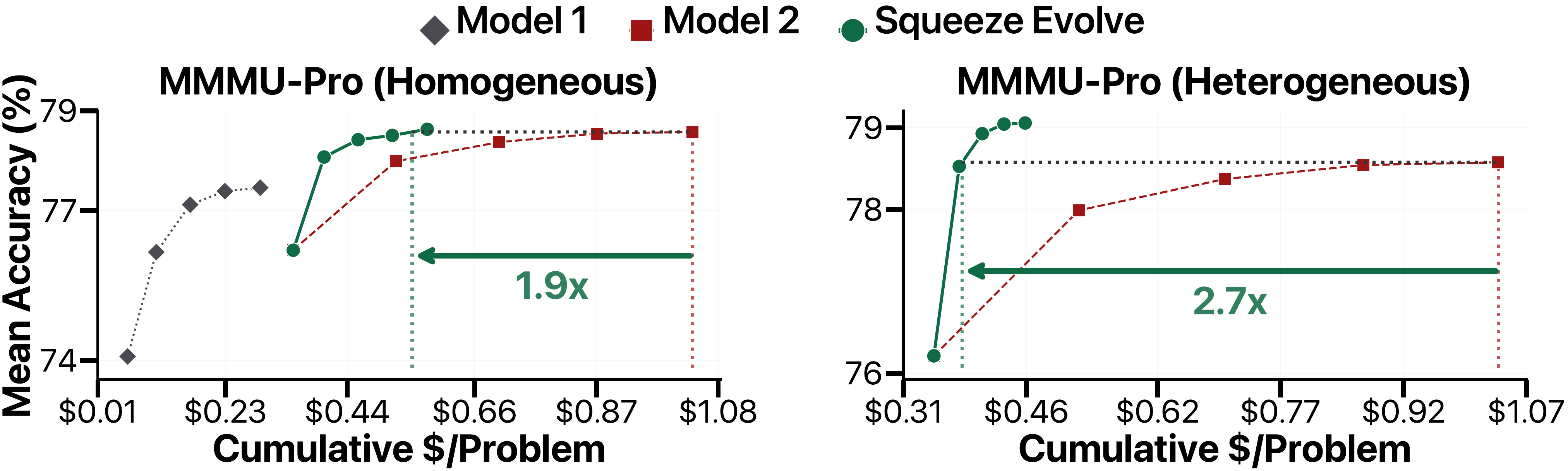}
\caption{\textbf{Accuracy vs.\ cumulative cost on MMMU-Pro for homogeneous and heterogeneous vision pairs.}
Left: Kimi-2.5 Instant (Model~1)\,/\,Thinking (Model~2). Right: Qwen3.5-35B (Model~1, text-only)\,$\to$\,Kimi-2.5 Thinking (Model~2). Savings are measured at matched accuracy. The heterogeneous pair achieves 2.7$\times$ savings despite Model~1 never seeing any images.}
\label{fig:acc-vs-cost-vision}
\end{figure*}

Table~\ref{tab:vision-highlights} summarizes representative results; accuracy-vs-cost curves for MMMU-Pro appear in Figure~\ref{fig:acc-vs-cost-vision}. On MMMU-Pro, the homogeneous pair matches Model~2 at \textbf{1.9$\times$} savings, while the heterogeneous pair \emph{surpasses} Model~2 (79.1\% vs.\ 78.6\%) at \textbf{2.7$\times$} savings, even though Model~1 never sees any images. On BabyVision, the homogeneous pair preserves accuracy at \textbf{2.5$\times$} savings. The heterogeneous result further reinforces the finding from Section~\ref{sec:prelim:limitations} that initialization quality is the dominant factor: once loop~0 grounds the population in image content, subsequent aggregation can be delegated to a cheaper text-only model. Full breakdowns appear in Tables~\ref{tab:empirical-vision-tasks} and~\ref{tab:empirical-hetero-tasks} (Appendix).
% -- Highlight results: Vision tasks (main paper) ------------------------------
{\small
\setlength{\tabcolsep}{3pt}
\renewcommand{\arraystretch}{1.0}
\begin{xltabular}{\textwidth}{@{}l l p{0.20\textwidth} p{0.20\textwidth} c c c@{}}
\caption{Representative results for multimodal vision benchmarks.
  $^\dagger$Model~1 operates in text-only mode (no image input after loop~0).
  Full breakdowns appear in Tables~\ref{tab:empirical-vision-tasks}
  and~\ref{tab:empirical-hetero-tasks} (Appendix).}
\label{tab:vision-highlights} \\
\toprule
\textbf{Data} & \textbf{Strategy} & \textbf{Model 1} & \textbf{Model 2}
& \textbf{Acc.} & \textbf{\$/Prob} & \textbf{\$ Savings} \\
\midrule
\endfirsthead
\toprule
\textbf{Data} & \textbf{Strategy} & \textbf{Model 1} & \textbf{Model 2}
& \textbf{Acc.} & \textbf{\$/Prob} & \textbf{\$ Savings} \\
\midrule
\endhead
\bottomrule
\endlastfoot

\multirow{2}{*}{MMMU-Pro}
  & RSA & --- & Kimi-2.5-Thinking & 78.58 & \$1.04 & 1.0$\times$ \\
  & \cellcolor{teal!4}\OURS ($p{=}0$) & \cellcolor{teal!4}Qwen3.5-35B-A3B$^\dagger$ & \cellcolor{teal!4}Kimi-2.5-Thinking & \cellcolor{teal!4}79.06 & \cellcolor{teal!4}\$0.46 & \cellcolor{teal!4}2.3$\times$ \\
\midrule
\multirow{2}{*}{BabyVision}
  & RSA & --- & Kimi-2.5-Thinking & 43.23 & \$2.05 & 1.0$\times$ \\
  & \cellcolor{teal!4}\OURS ($p{=}0$) & \cellcolor{teal!4}Kimi-2.5-Instant & \cellcolor{teal!4}Kimi-2.5-Thinking & \cellcolor{teal!4}41.56 & \cellcolor{teal!4}\$0.81 & \cellcolor{teal!4}2.5$\times$ \\

\end{xltabular}
}

% -------------------------------------------------

\subsection{ARC-AGI-V2}
\label{sec:arc-agi-v2}

We evaluate \textsc{Squeeze Evolve} on ARC-AGI-V2~\citep{chollet2025arcagi2} public evaluation set. Since the Gemini API does not expose logprobs, we use answer diversity (Eq.~\ref{eq:diversity}) as the fitness signal (Table~\ref{tab:instantiations}). Groups with non-zero diversity are recombined by Gemini~3.1~Pro~\citep{google2025gemini3-1}; consensus groups fall back to majority vote. With this routing, \OURS achieves 97.5\% at \$7.74/task.

Using this result as a baseline, we further reduce cost by adding Gemini~3.0~Flash~\citep{google2025gemini3} as Model 1 to the recombination function, yielding a three-way routing rule: high-diversity groups with 3 or more unique answers invoke the expensive Gemini~3.1~Pro Model 2, lower diversity groups are handled by Gemini~3.0~Flash, and groups that have already reached consensus are aggregated via lightweight non-LLM methods (e.g., majority vote). 

With this recombination function, we observe immediate convergence to the pass@k score after one aggregation step, achieving the same 97.5\% accuracy for only \$5.93/task, a $1.2\times$ savings.

This is a new SoTA cost-capability frontier result on ARC-AGI-V2 public evaluation set to date. Even compared to code-execution-based approaches, \textsc{Squeeze Evolve} reaches comparable accuracy at a lower cost to Confluence Lab~\citep{confluencelabs2026arcagi2} (97.9\%, \$11.77/task) and Imbue~\citep{imbue2026arcagi2} (95.1\%, \$8.71/task).
% -- ARC-AGI-V2 results (main paper) -------------------------------------------
\begin{table}[t]
\centering
{\small
\setlength{\tabcolsep}{3pt}
\renewcommand{\arraystretch}{1.0}
\caption{ARC-AGI-V2 public evaluation results.
  $N{=}4$, $K{=}2$.
  $^\dagger$Uses code execution and program synthesis.
  Extended results in Table~\ref{tab:arc-agi-v2-full} (Appendix).}
\label{tab:arc-agi-v2}
\begin{tabular}{@{}l l l c r r@{}}
\toprule
\textbf{Strategy} & \textbf{Model 1} & \textbf{Model 2}
& \textbf{Acc.} & \textbf{\$/Task} & \textbf{Savings} \\
\midrule
\multicolumn{6}{@{}l}{\textit{Code-execution methods}} \\
\midrule
Imbue$^\dagger$ & --- & Gemini~3.1~Pro & 95.1 & \$8.71 & --- \\
Confluence Lab$^\dagger$ & --- & --- & 97.9 & \$11.77 & --- \\
\midrule
\multicolumn{6}{@{}l}{\textit{Full pipeline ($T{=}10$)}} \\
\midrule
RSA & --- & Gemini~3.1~Pro & 93.3 & \$28.85 & 1.0$\times$ \\
\cellcolor{teal!4}\OURS & \cellcolor{teal!4}--- & \cellcolor{teal!4}Gemini~3.1~Pro & \cellcolor{teal!4}\textbf{97.5} & \cellcolor{teal!4}\$7.74 & \cellcolor{teal!4}3.7$\times$ \\
\midrule
\multicolumn{6}{@{}l}{\textit{Single recombination ($T{=}2$)}} \\
\midrule
\OURS & --- & Gemini~3.1~Pro & 94.2 & \$5.62 & 5.1$\times$ \\
\cellcolor{teal!4}\OURS & \cellcolor{teal!4}Gemini~3.0~Flash & \cellcolor{teal!4}Gemini~3.1~Pro & \cellcolor{teal!4}\textbf{97.5} & \cellcolor{teal!4}\textbf{\$5.93} & \cellcolor{teal!4}\textbf{4.9$\times$} \\
\bottomrule
\end{tabular}
}
\end{table}

% -------------------------------------------------

\subsection{Circle Packing: Scientific Discovery}
\label{sec:circle-packing}

We apply \textsc{Squeeze Evolve} to the circle packing problem studied in AlphaEvolve~\citep{novikov2025alphaevolvecodingagentscientific} and subsequent evolutionary frameworks: pack $n=26$ non-overlapping circles in a unit square to maximize the sum of their radii. Unlike the reasoning and visual tasks above, this is an open-ended optimization problem with a continuous objective, showcasing \OURS's capability for evolutionary discovery. We use GPT-OSS-20B as Model 1 and GPT-OSS-120B as Model 2 with $N{=}128$, $K{=}4$, and $T{=}50$ loops. The fitness signal is group confidence with fitness-weighted selection ($\zeta{=}0.5$), a fixed confidence threshold at the 50th percentile for routing, and the accumulate update rule (Table~\ref{tab:instantiations}). At termination, we draw $N$ candidates from the cumulative pool via confidence-weighted sampling and report the highest score.
\begin{table}[H]
\centering
\small
\setlength{\tabcolsep}{4pt}
\caption{Comparison of methods on Circle Packing ($n{=}26$, $\uparrow$). ShinkaEvolve uses an ensemble of Claude Sonnet-4, GPT-4.1, GPT-4.1-mini, GPT-4.1-nano, and o4-mini.}
\label{tab:circle-packing}
\begin{tabular}{@{}llc@{}}
\toprule
\textbf{Method} & \textbf{Model} & \textbf{Score ($\uparrow$)} \\
\midrule
ShinkaEvolve~\citep{lange2025shinkaevolveopenendedsampleefficientprogram} & Ensemble (see caption) & 2.635982 \\
\rowcolor{teal!8} \OURS & GPT-OSS-120B + 20B & 2.635896 \\
AlphaEvolve~\citep{novikov2025alphaevolvecodingagentscientific} & Gemini-2.0 Pro + Flash & 2.635862 \\
OpenEvolve~\citep{openevolve} & Gemini-2.0 Flash + Claude 3.7 Sonnet & 2.634292 \\
\bottomrule
\end{tabular}
\end{table}
\OURS achieves a comparable score to the best evolutionary frameworks (Table~\ref{tab:circle-packing}), notably without executing generated programs in-flight or using closed-weight models. While other frameworks rely on running candidates and feeding execution results back to inform subsequent generations, \OURS uses no ground-truth feedback or evaluator output. Instead, model-intrinsic confidence exhibits a non-zero correlation with the objective score, and this weak signal suffices to improve both the average and best programs over iterations, suggesting that confidence can serve as a practical surrogate for verification in discovery settings. Analysis of the algorithm and source code as well as hyperparameters appear in Appendix~\ref{sec:circle_packing}.

\section{System Results}
\label{sec:system-results}
\paragraph{Routing overhead.}
A natural systems question is whether confidence scoring and model dispatch introduce enough additional latency to undermine multi-model routing.
To isolate this cost, we compare two conditions under identical inference configurations $(N{=}16, K{=}4, T{=}10)$: \textbf{RSA-$M_2$}, standard RSA with all calls executed by Model~2 and no routing logic, and \textbf{\OURS{}-$M_2$}, which enables confidence scoring and threshold computation but forces every aggregation call to Model~2. The difference isolates the routing overhead itself, and is a conservative worst-case measurement since \OURS{} normally reduces latency by routing a subset of aggregations to Model~1.
Across all three models, routing adds only \textbf{2.4\text{--}4.3\%} to end-to-end latency on average, confirming that confidence scoring is a batched prefill-only operation whose cost is negligible relative to generation. Per-benchmark breakdowns, including the measurement protocol and overhead definitions, appear in Appendix~\ref{app:routing-overhead}.

\begin{figure}[h]
\centering
\includegraphics[width=\columnwidth]{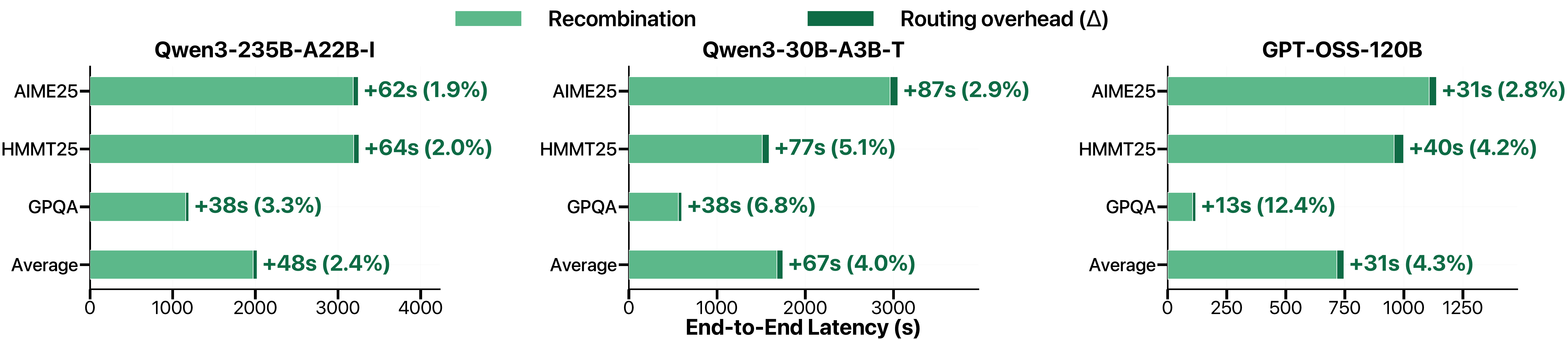}
\caption{\textbf{Routing overhead is minimal.}
Routing adds 1.9--6.8\% to end-to-end latency for the Qwen models and 2.8--12.4\% for GPT-OSS-120B, with the higher relative overhead on GPQA reflecting its short absolute generation time (106s).
Full results in Table~\ref{tab:routing-overhead} (Appendix).}
\label{fig:routing-overhead}
\end{figure}

\begin{figure*}[h]
\centering
\includegraphics[width=\linewidth]{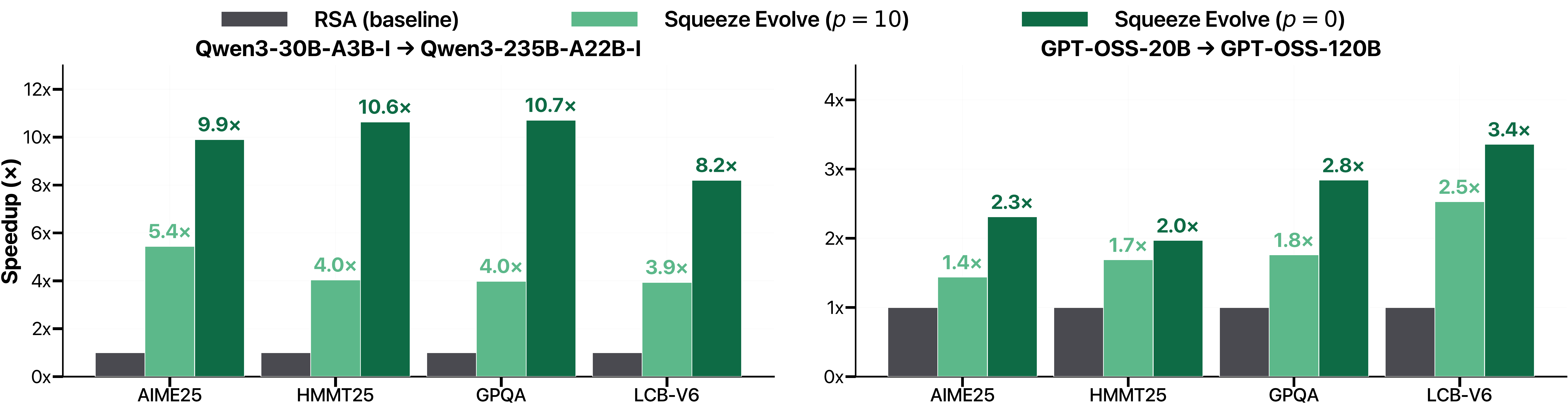}
\caption{\textbf{Fixed-budget throughput speedup over RSA.}
Under the same total GPU budget, the Qwen pair achieves 4--10$\times$ speedup and the GPT-OSS pair 1.4--3.4$\times$.
Full results in Table~\ref{tab:throughput} (Appendix).}
\label{fig:throughput}
\end{figure*}

\paragraph{System throughput.}
We next ask whether routing improves steady-state serving throughput under a fixed GPU budget $G$. Unlike routing overhead, throughput is a property of the full deployment: if either model pool is underprovisioned, the slower pool becomes the bottleneck and erases the benefit of cheaper aggregation. We compare RSA and \OURS{} under the same total budget: RSA allocates all $G$ GPUs to Model~2, while \OURS{} partitions them into a large-model pool $G_L$ and a small-model pool $G_S$ ($G_L + G_S = G$), sized so that their loop service times are approximately matched. We report requests per second rather than tokens per second because Model~1 and Model~2 produce different numbers of output tokens for the same query, making a token-based metric an unfair comparison.

Figure~\ref{fig:throughput} shows that the Qwen3-30B/235B pair achieves \textbf{4--10$\times$} throughput speedup owing to the large Model~1 to Model~2 size ratio, while the GPT-OSS pair yields \textbf{1.4--3.4$\times$} speedup. The larger gains for the Qwen pair likely reflect a larger effective inference asymmetry between its small and large models under our serving setup. For MoE models, this asymmetry is both due to headline total parameter count (which affects the number of GPUs required for serving the large/small model) as well as by active parameters per token  (which affects the time for decoding each token). By these measures, the GPT-OSS 20B/120B pair is substantially less separated than the Qwen3 30B/235B pair, which helps explain its smaller throughput gains. Full per-benchmark breakdowns, observed routing shares, GPU splits, and measurement protocol appear in Appendix~\ref{app:throughput}.

\section{Future Work}
Several directions naturally extend \OURS{}. Our routing relies on model-intrinsic confidence and answer diversity, which are lightweight but inherently noisy proxies; incorporating sparse or approximate verification (e.g., executing a small fraction of candidate programs or training a lightweight correctness classifier) could sharpen fitness estimation at modest additional cost, particularly for scientific discovery tasks where the gap between verifier-free and verifier-based methods is narrowest. Population size, group size, loop count, and routing threshold are currently fixed per task, and learning to adjust these dynamically, such as stopping early upon convergence or expanding when diversity collapses, would improve both efficiency and robustness. \OURS{} currently operates on complete trajectories; decomposing reasoning into intermediate steps and selectively regenerating only uncertain segments could reduce redundant computation while preserving the strongest partial solutions. Finally, the empirical success of confidence-based routing raises open theoretical questions about when model-intrinsic confidence reliably separates correct from incorrect populations and what convergence guarantees can be established for verifier-free multi-model evolution.

\bibliographystyle{plain}
{\small
\bibliography{references}

@misc{wang2023selfconsistencyimproveschainthought,
      title={Self-Consistency Improves Chain of Thought Reasoning in Language Models}, 
      author={Xuezhi Wang and Jason Wei and Dale Schuurmans and Quoc Le and Ed Chi and Sharan Narang and Aakanksha Chowdhery and Denny Zhou},
      year={2023},
      eprint={2203.11171},
      archivePrefix={arXiv},
      primaryClass={cs.CL},
      url={https://arxiv.org/abs/2203.11171}, 
}

@misc{madaan2023selfrefineiterativerefinementselffeedback,
      title={Self-Refine: Iterative Refinement with Self-Feedback}, 
      author={Aman Madaan and Niket Tandon and Prakhar Gupta and Skyler Hallinan and Luyu Gao and Sarah Wiegreffe and Uri Alon and Nouha Dziri and Shrimai Prabhumoye and Yiming Yang and Shashank Gupta and Bodhisattwa Prasad Majumder and Katherine Hermann and Sean Welleck and Amir Yazdanbakhsh and Peter Clark},
      year={2023},
      eprint={2303.17651},
      archivePrefix={arXiv},
      primaryClass={cs.CL},
      url={https://arxiv.org/abs/2303.17651}, 
}

@misc{venkatraman2026recursiveselfaggregationunlocksdeep,
      title={Recursive Self-Aggregation Unlocks Deep Thinking in Large Language Models}, 
      author={Siddarth Venkatraman and Vineet Jain and Sarthak Mittal and Vedant Shah and Johan Obando-Ceron and Yoshua Bengio and Brian R. Bartoldson and Bhavya Kailkhura and Guillaume Lajoie and Glen Berseth and Nikolay Malkin and Moksh Jain},
      year={2026},
      eprint={2509.26626},
      archivePrefix={arXiv},
      primaryClass={cs.LG},
      url={https://arxiv.org/abs/2509.26626}, 
}

@misc{novikov2025alphaevolvecodingagentscientific,
      title={AlphaEvolve: A coding agent for scientific and algorithmic discovery}, 
      author={Alexander Novikov and Ngân Vũ and Marvin Eisenberger and Emilien Dupont and Po-Sen Huang and Adam Zsolt Wagner and Sergey Shirobokov and Borislav Kozlovskii and Francisco J. R. Ruiz and Abbas Mehrabian and M. Pawan Kumar and Abigail See and Swarat Chaudhuri and George Holland and Alex Davies and Sebastian Nowozin and Pushmeet Kohli and Matej Balog},
      year={2025},
      eprint={2506.13131},
      archivePrefix={arXiv},
      primaryClass={cs.AI},
      url={https://arxiv.org/abs/2506.13131}, 
}

@misc{lange2025shinkaevolveopenendedsampleefficientprogram,
      title={ShinkaEvolve: Towards Open-Ended And Sample-Efficient Program Evolution}, 
      author={Robert Tjarko Lange and Yuki Imajuku and Edoardo Cetin},
      year={2025},
      eprint={2509.19349},
      archivePrefix={arXiv},
      primaryClass={cs.CL},
      url={https://arxiv.org/abs/2509.19349}, 
}

@misc{liu2026evoxmetaevolutionautomateddiscovery,
      title={EvoX: Meta-Evolution for Automated Discovery}, 
      author={Shu Liu and Shubham Agarwal and Monishwaran Maheswaran and Mert Cemri and Zhifei Li and Qiuyang Mang and Ashwin Naren and Ethan Boneh and Audrey Cheng and Melissa Z. Pan and Alexander Du and Kurt Keutzer and Alvin Cheung and Alexandros G. Dimakis and Koushik Sen and Matei Zaharia and Ion Stoica},
      year={2026},
      eprint={2602.23413},
      archivePrefix={arXiv},
      primaryClass={cs.LG},
      url={https://arxiv.org/abs/2602.23413}, 
}

@software{openevolve,
  title = {OpenEvolve: an open-source evolutionary coding agent},
  author = {Asankhaya Sharma},
  year = {2025},
  publisher = {GitHub},
  url = {https://github.com/algorithmicsuperintelligence/openevolve}
}

@article{chollet2025arcagi2,
  title={{ARC-AGI-2}: A New Challenge for Frontier {AI} Reasoning Systems},
  author={Chollet, Fran{\c{c}}ois and Knoop, Mike and Kamradt, Greg and Landers, Chris},
  journal={arXiv preprint arXiv:2505.11831},
  year={2025}
}

@misc{imbue2026arcagi2,
  author       = {{Imbue}},
  title        = {Beating {ARC-AGI-2} with Code Evolution},
  year         = {2026},
  url          = {https://imbue.com/research/2026-02-27-arc-agi-2-evolution/},
  note         = {Blog post, accessed March 2026},
}

@misc{confluencelabs2026arcagi2,
  author       = {{Confluence Labs}},
  title        = {State-of-the-art {ARC-AGI-2} solver},
  year         = {2026},
  url          = {https://github.com/confluence-labs/arc-agi-2},
  note         = {GitHub repository, accessed March 2026},
}

@article{howard2016multiscale,
  author    = {Howard, N. T. and Holland, C. and White, A. E. and Greenwald, M. and Candy, J.},
  title     = {Multi-scale gyrokinetic simulation of tokamak plasmas: enhanced heat loss due to cross-scale coupling of plasma turbulence},
  journal   = {Nuclear Fusion},
  volume    = {56},
  year      = {2016}}

@misc{openai2025gptoss,
      title={gpt-oss-120b \& gpt-oss-20b Model Card}, 
      author={Sandhini Agarwal and Lama Ahmad and Jason Ai and Sam Altman and Andy Applebaum and Edwin Arbus and Rahul K. Arora and Yu Bai and Bowen Baker and Haiming Bao et al.},
      year={2025},
      eprint={2508.10925},
      archivePrefix={arXiv},
      primaryClass={cs.CL},
      url={https://arxiv.org/abs/2508.10925}, 
}

@misc{openai2026o4mini,
  title        = {o4-mini Model},
  author       = {{OpenAI}},
  year         = {2026},
  howpublished = {\url{https://developers.openai.com/api/docs/models/o4-mini}},
  note         = {OpenAI API model page, accessed March 16, 2026}
}

@misc{anthropic2026pricing,
  title        = {Pricing},
  author       = {{Anthropic}},
  year         = {2026},
  howpublished = {\url{https://platform.claude.com/docs/en/about-claude/pricing}},
  note         = {Claude API pricing page, accessed March 16, 2026}
}

@misc{google2026pricing,
  title        = {Gemini Developer API Pricing},
  author       = {{Google}},
  year         = {2026},
  howpublished = {\url{https://ai.google.dev/gemini-api/docs/pricing}},
  note         = {Google AI for Developers pricing page, accessed March 16, 2026}
}

@misc{together2026gptoss120b,
  title        = {gpt-oss-120B API},
  author       = {{Together AI}},
  year         = {2026},
  howpublished = {\url{https://www.together.ai/models/gpt-oss-120b}},
  note         = {Together AI model page, accessed March 16, 2026}
}

@misc{together2026qwen235b,
  title        = {Qwen3 235B A22B Instruct 2507 FP8 API},
  author       = {{Together AI}},
  year         = {2026},
  howpublished = {\url{https://www.together.ai/models/qwen3-235b-a22b-instruct-2507-fp8}},
  note         = {Together AI model page, accessed March 16, 2026}
}

@misc{snell2024scalingllmtesttimecompute,
      title={Scaling LLM Test-Time Compute Optimally can be More Effective than Scaling Model Parameters}, 
      author={Charlie Snell and Jaehoon Lee and Kelvin Xu and Aviral Kumar},
      year={2024},
      eprint={2408.03314},
      archivePrefix={arXiv},
      primaryClass={cs.LG},
      url={https://arxiv.org/abs/2408.03314}, 
}

@misc{wu2025inferencescalinglawsempirical,
      title={Inference Scaling Laws: An Empirical Analysis of Compute-Optimal Inference for Problem-Solving with Language Models}, 
      author={Yangzhen Wu and Zhiqing Sun and Shanda Li and Sean Welleck and Yiming Yang},
      year={2025},
      eprint={2408.00724},
      archivePrefix={arXiv},
      primaryClass={cs.AI},
      url={https://arxiv.org/abs/2408.00724}, 
}

@misc{muennighoff2025s1simpletesttimescaling,
      title={s1: Simple test-time scaling}, 
      author={Niklas Muennighoff and Zitong Yang and Weijia Shi and Xiang Lisa Li and Li Fei-Fei and Hannaneh Hajishirzi and Luke Zettlemoyer and Percy Liang and Emmanuel Candès and Tatsunori Hashimoto},
      year={2025},
      eprint={2501.19393},
      archivePrefix={arXiv},
      primaryClass={cs.CL},
      url={https://arxiv.org/abs/2501.19393}, 
}

@misc{weng2023largelanguagemodelsbetter,
      title={Large Language Models are Better Reasoners with Self-Verification}, 
      author={Yixuan Weng and Minjun Zhu and Fei Xia and Bin Li and Shizhu He and Shengping Liu and Bin Sun and Kang Liu and Jun Zhao},
      year={2023},
      eprint={2212.09561},
      archivePrefix={arXiv},
      primaryClass={cs.AI},
      url={https://arxiv.org/abs/2212.09561}, 
}

@misc{yao2023treethoughtsdeliberateproblem,
      title={Tree of Thoughts: Deliberate Problem Solving with Large Language Models}, 
      author={Shunyu Yao and Dian Yu and Jeffrey Zhao and Izhak Shafran and Thomas L. Griffiths and Yuan Cao and Karthik Narasimhan},
      year={2023},
      eprint={2305.10601},
      archivePrefix={arXiv},
      primaryClass={cs.CL},
      url={https://arxiv.org/abs/2305.10601}, 
}

@misc{zhang2024accessinggpt4levelmathematical,
      title={Accessing GPT-4 level Mathematical Olympiad Solutions via Monte Carlo Tree Self-refine with LLaMa-3 8B}, 
      author={Di Zhang and Xiaoshui Huang and Dongzhan Zhou and Yuqiang Li and Wanli Ouyang},
      year={2024},
      eprint={2406.07394},
      archivePrefix={arXiv},
      primaryClass={cs.AI},
      url={https://arxiv.org/abs/2406.07394}, 
}

@misc{zhou2024languageagenttreesearch,
      title={Language Agent Tree Search Unifies Reasoning Acting and Planning in Language Models}, 
      author={Andy Zhou and Kai Yan and Michal Shlapentokh-Rothman and Haohan Wang and Yu-Xiong Wang},
      year={2024},
      eprint={2310.04406},
      archivePrefix={arXiv},
      primaryClass={cs.AI},
      url={https://arxiv.org/abs/2310.04406}, 
}

@misc{agrawal2026gepareflectivepromptevolution,
      title={GEPA: Reflective Prompt Evolution Can Outperform Reinforcement Learning}, 
      author={Lakshya A Agrawal and Shangyin Tan and Dilara Soylu and Noah Ziems and Rishi Khare and Krista Opsahl-Ong and Arnav Singhvi and Herumb Shandilya and Michael J Ryan and Meng Jiang and Christopher Potts and Koushik Sen and Alexandros G. Dimakis and Ion Stoica and Dan Klein and Matei Zaharia and Omar Khattab},
      year={2026},
      eprint={2507.19457},
      archivePrefix={arXiv},
      primaryClass={cs.CL},
      url={https://arxiv.org/abs/2507.19457}, 
}

@misc{assumpo2026codeevolveopensourceevolutionary,
      title={CodeEvolve: an open source evolutionary coding agent for algorithmic discovery and optimization}, 
      author={Henrique Assumpção and Diego Ferreira and Leandro Campos and Fabricio Murai},
      year={2026},
      eprint={2510.14150},
      archivePrefix={arXiv},
      primaryClass={cs.AI},
      url={https://arxiv.org/abs/2510.14150}, 
}

@misc{openai2024openaio1card,
      title={OpenAI o1 System Card}, 
      author={Aaron Jaech and Adam Kalai and Adam Lerer and Adam Richardson and Ahmed El-Kishky and Aiden Low and Alec Helyar and Aleksander Madry and Alex Beutel and Alex Carney et al.},
      year={2024},
      eprint={2412.16720},
      archivePrefix={arXiv},
      primaryClass={cs.AI},
      url={https://arxiv.org/abs/2412.16720}, 
}

@article{Guo_2025,
   title={DeepSeek-R1 incentivizes reasoning in LLMs through reinforcement learning},
   volume={645},
   ISSN={1476-4687},
   url={http://dx.doi.org/10.1038/s41586-025-09422-z},
   DOI={10.1038/s41586-025-09422-z},
   number={8081},
   journal={Nature},
   publisher={Springer Science and Business Media LLC},
   author={Guo, Daya and Yang, Dejian and Zhang, Haowei and Song, Junxiao and Wang, Peiyi and Zhu, Qihao and Xu, Runxin and Zhang, Ruoyu and Ma, Shirong and Bi, Xiao et al.},
   year={2025},
   month=sep, pages={633–638} }

@inproceedings{kwon2023efficient,
  title={Efficient Memory Management for Large Language Model Serving with PagedAttention},
  author={Woosuk Kwon and Zhuohan Li and Siyuan Zhuang and Ying Sheng and Lianmin Zheng and Cody Hao Yu and Joseph E. Gonzalez and Hao Zhang and Ion Stoica},
  booktitle={Proceedings of the ACM SIGOPS 29th Symposium on Operating Systems Principles},
  year={2023}
}

@misc{zheng2024sglangefficientexecutionstructured,
      title={SGLang: Efficient Execution of Structured Language Model Programs}, 
      author={Lianmin Zheng and Liangsheng Yin and Zhiqiang Xie and Chuyue Sun and Jeff Huang and Cody Hao Yu and Shiyi Cao and Christos Kozyrakis and Ion Stoica and Joseph E. Gonzalez and Clark Barrett and Ying Sheng},
      year={2024},
      eprint={2312.07104},
      archivePrefix={arXiv},
      primaryClass={cs.AI},
      url={https://arxiv.org/abs/2312.07104}, 
}

@misc{yue2025mmmuprorobustmultidisciplinemultimodal,
      title={MMMU-Pro: A More Robust Multi-discipline Multimodal Understanding Benchmark}, 
      author={Xiang Yue and Tianyu Zheng and Yuansheng Ni and Yubo Wang and Kai Zhang and Shengbang Tong and Yuxuan Sun and Botao Yu and Ge Zhang and Huan Sun and Yu Su and Wenhu Chen and Graham Neubig},
      year={2025},
      eprint={2409.02813},
      archivePrefix={arXiv},
      primaryClass={cs.CL},
      url={https://arxiv.org/abs/2409.02813}, 
}

@misc{chen2026babyvisionvisualreasoninglanguage,
      title={BabyVision: Visual Reasoning Beyond Language}, 
      author={Liang Chen and Weichu Xie and Yiyan Liang and Hongfeng He and Hans Zhao and Zhibo Yang and Zhiqi Huang and Haoning Wu and Haoyu Lu and Y. charles and Yiping Bao and Yuantao Fan and Guopeng Li and Haiyang Shen and Xuanzhong Chen and Wendong Xu and Shuzheng Si and Zefan Cai and Wenhao Chai and Ziqi Huang and Fangfu Liu and Tianyu Liu and Baobao Chang and Xiaobo Hu and Kaiyuan Chen and Yixin Ren and Yang Liu and Yuan Gong and Kuan Li},
      year={2026},
      eprint={2601.06521},
      archivePrefix={arXiv},
      primaryClass={cs.CV},
      url={https://arxiv.org/abs/2601.06521}, 
}

@misc{qwen3technicalreport,
      title={Qwen3 Technical Report}, 
      author={Qwen Team},
      year={2025},
      eprint={2505.09388},
      archivePrefix={arXiv},
      primaryClass={cs.CL},
      url={https://arxiv.org/abs/2505.09388}, 
}

@misc{kimiteam2026kimik25visualagentic,
      title={Kimi K2.5: Visual Agentic Intelligence}, 
      author={Kimi Team and Tongtong Bai and Yifan Bai and Yiping Bao and S. H. Cai and Yuan Cao and Y. Charles and H. S. Che and Cheng Chen and Guanduo Chen and Huarong Chen et al.},
      year={2026},
      eprint={2602.02276},
      archivePrefix={arXiv},
      primaryClass={cs.CL},
      url={https://arxiv.org/abs/2602.02276}, 
}

@misc{qwen3.5,
    title  = {{Qwen3.5}: Towards Native Multimodal Agents},
    author = {{Qwen Team}},
    month  = {February},
    year   = {2026},
    url    = {https://qwen.ai/blog?id=qwen3.5}
}

@misc{balunović2026matharenaevaluatingllmsuncontaminated,
      title={MathArena: Evaluating LLMs on Uncontaminated Math Competitions}, 
      author={Mislav Balunović and Jasper Dekoninck and Ivo Petrov and Nikola Jovanović and Martin Vechev},
      year={2026},
      eprint={2505.23281},
      archivePrefix={arXiv},
      primaryClass={cs.AI},
      url={https://arxiv.org/abs/2505.23281}, 
}

@inproceedings{rein2024gpqa,
  title={Gpqa: A graduate-level google-proof q\&a benchmark},
  author={Rein, David and Hou, Betty Li and Stickland, Asa Cooper and Petty, Jackson and Pang, Richard Yuanzhe and Dirani, Julien and Michael, Julian and Bowman, Samuel R},
  booktitle={First Conference on Language Modeling},
  year={2024}
}

@techreport{openai2025gpt5,
  title={{GPT-5} System Card},
  author={{OpenAI}},
  year={2025},
  month={August},
  institution={OpenAI},
  url={https://cdn.openai.com/gpt-5-system-card.pdf}
}

@techreport{google2025gemini3,
  title={{Gemini 3 Flash} Model Card},
  author={{Google DeepMind}},
  year={2025},
  month={December},
  institution={Google DeepMind},
  url={https://storage.googleapis.com/deepmind-media/Model-Cards/Gemini-3-Flash-Model-Card.pdf}
}

@techreport{google2025gemini3-1,
  title={{Gemini 3.1 Pro} Model Card},
  author={{Google DeepMind}},
  year={2026},
  month={February},
  institution={Google DeepMind},
  url={https://storage.googleapis.com/deepmind-media/Model-Cards/Gemini-3-1-Pro-Model-Card.pdf}
}

@misc{jain2024livecodebenchholisticcontaminationfree,
      title={LiveCodeBench: Holistic and Contamination Free Evaluation of Large Language Models for Code}, 
      author={Naman Jain and King Han and Alex Gu and Wen-Ding Li and Fanjia Yan and Tianjun Zhang and Sida Wang and Armando Solar-Lezama and Koushik Sen and Ion Stoica},
      year={2024},
      eprint={2403.07974},
      archivePrefix={arXiv},
      primaryClass={cs.SE},
      url={https://arxiv.org/abs/2403.07974}, 
}

@misc{bansal2024smallerweakerbettertraining,
      title={Smaller, Weaker, Yet Better: Training LLM Reasoners via Compute-Optimal Sampling}, 
      author={Hritik Bansal and Arian Hosseini and Rishabh Agarwal and Vinh Q. Tran and Mehran Kazemi},
      year={2024},
      eprint={2408.16737},
      archivePrefix={arXiv},
      primaryClass={cs.CL},
      url={https://arxiv.org/abs/2408.16737}, 
}

@misc{brown2024largelanguagemonkeysscaling,
      title={Large Language Monkeys: Scaling Inference Compute with Repeated Sampling}, 
      author={Bradley Brown and Jordan Juravsky and Ryan Ehrlich and Ronald Clark and Quoc V. Le and Christopher Ré and Azalia Mirhoseini},
      year={2024},
      eprint={2407.21787},
      archivePrefix={arXiv},
      primaryClass={cs.LG},
      url={https://arxiv.org/abs/2407.21787}, 
}

@misc{setlur2025scalingtesttimecomputeverification,
      title={Scaling Test-Time Compute Without Verification or RL is Suboptimal}, 
      author={Amrith Setlur and Nived Rajaraman and Sergey Levine and Aviral Kumar},
      year={2025},
      eprint={2502.12118},
      archivePrefix={arXiv},
      primaryClass={cs.LG},
      url={https://arxiv.org/abs/2502.12118}, 
}

@misc{li2025llmsgeneratebetteranswer,
      title={LLMs Can Generate a Better Answer by Aggregating Their Own Responses}, 
      author={Zichong Li and Xinyu Feng and Yuheng Cai and Zixuan Zhang and Tianyi Liu and Chen Liang and Weizhu Chen and Haoyu Wang and Tuo Zhao},
      year={2025},
      eprint={2503.04104},
      archivePrefix={arXiv},
      primaryClass={cs.CL},
      url={https://arxiv.org/abs/2503.04104}, 
}

@misc{madaan2025rethinkingthinkingtokensllms,
      title={Rethinking Thinking Tokens: LLMs as Improvement Operators}, 
      author={Lovish Madaan and Aniket Didolkar and Suchin Gururangan and John Quan and Ruan Silva and Ruslan Salakhutdinov and Manzil Zaheer and Sanjeev Arora and Anirudh Goyal},
      year={2025},
      eprint={2510.01123},
      archivePrefix={arXiv},
      primaryClass={cs.LG},
      url={https://arxiv.org/abs/2510.01123}, 
}

@misc{khairi2025makingtakingbestn,
      title={Making, not Taking, the Best of N}, 
      author={Ammar Khairi and Daniel D'souza and Marzieh Fadaee and Julia Kreutzer},
      year={2025},
      eprint={2510.00931},
      archivePrefix={arXiv},
      primaryClass={cs.CL},
      url={https://arxiv.org/abs/2510.00931}, 
}

@misc{wang2024mixtureofagentsenhanceslargelanguage,
      title={Mixture-of-Agents Enhances Large Language Model Capabilities}, 
      author={Junlin Wang and Jue Wang and Ben Athiwaratkun and Ce Zhang and James Zou},
      year={2024},
      eprint={2406.04692},
      archivePrefix={arXiv},
      primaryClass={cs.CL},
      url={https://arxiv.org/abs/2406.04692}, 
}

@misc{singh2026v1unifyinggenerationselfverification,
      title={$V_1$: Unifying Generation and Self-Verification for Parallel Reasoners}, 
      author={Harman Singh and Xiuyu Li and Kusha Sareen and Monishwaran Maheswaran and Sijun Tan and Xiaoxia Wu and Junxiong Wang and Alpay Ariyak and Qingyang Wu and Samir Khaki and Rishabh Tiwari and Long Lian and Yucheng Lu and Boyi Li and Alane Suhr and Ben Athiwaratkun and Kurt Keutzer},
      year={2026},
      eprint={2603.04304},
      archivePrefix={arXiv},
      primaryClass={cs.CL},
      url={https://arxiv.org/abs/2603.04304}, 
}

@misc{lightman2023letsverifystepstep,
      title={Let's Verify Step by Step}, 
      author={Hunter Lightman and Vineet Kosaraju and Yura Burda and Harri Edwards and Bowen Baker and Teddy Lee and Jan Leike and John Schulman and Ilya Sutskever and Karl Cobbe},
      year={2023},
      eprint={2305.20050},
      archivePrefix={arXiv},
      primaryClass={cs.LG},
      url={https://arxiv.org/abs/2305.20050}, 
}

@misc{cobbe2021trainingverifierssolvemath,
      title={Training Verifiers to Solve Math Word Problems}, 
      author={Karl Cobbe and Vineet Kosaraju and Mohammad Bavarian and Mark Chen and Heewoo Jun and Lukasz Kaiser and Matthias Plappert and Jerry Tworek and Jacob Hilton and Reiichiro Nakano and Christopher Hesse and John Schulman},
      year={2021},
      eprint={2110.14168},
      archivePrefix={arXiv},
      primaryClass={cs.LG},
      url={https://arxiv.org/abs/2110.14168}, 
}

@misc{zhang2025generativeverifiersrewardmodeling,
      title={Generative Verifiers: Reward Modeling as Next-Token Prediction}, 
      author={Lunjun Zhang and Arian Hosseini and Hritik Bansal and Mehran Kazemi and Aviral Kumar and Rishabh Agarwal},
      year={2025},
      eprint={2408.15240},
      archivePrefix={arXiv},
      primaryClass={cs.LG},
      url={https://arxiv.org/abs/2408.15240}, 
}

@misc{lu2025doesverificationpayoff,
      title={When Does Verification Pay Off? A Closer Look at LLMs as Solution Verifiers}, 
      author={Jack Lu and Ryan Teehan and Jinran Jin and Mengye Ren},
      year={2025},
      eprint={2512.02304},
      archivePrefix={arXiv},
      primaryClass={cs.CL},
      url={https://arxiv.org/abs/2512.02304}, 
}

@misc{fu2025deepthinkconfidence,
      title={Deep Think with Confidence}, 
      author={Yichao Fu and Xuewei Wang and Yuandong Tian and Jiawei Zhao},
      year={2025},
      eprint={2508.15260},
      archivePrefix={arXiv},
      primaryClass={cs.LG},
      url={https://arxiv.org/abs/2508.15260}, 
}

@article{funsearch,
	author = {Romera-Paredes, Bernardino and Barekatain, Mohammadamin and Novikov, Alexander and Balog, Matej and Kumar, M. Pawan and Dupont, Emilien and Ruiz, Francisco J. R. and Ellenberg, Jordan S. and Wang, Pengming and Fawzi, Omar and Kohli, Pushmeet and Fawzi, Alhussein},
	date = {2024/01/01},
	doi = {10.1038/s41586-023-06924-6},
	id = {Romera-Paredes2024},
	isbn = {1476-4687},
	journal = {Nature},
	number = {7995},
	pages = {468--475},
	title = {Mathematical discoveries from program search with large language models},
	url = {https://doi.org/10.1038/s41586-023-06924-6},
	volume = {625},
	year = {2024}}

@misc{lehman2022evolutionlargemodels,
      title={Evolution through Large Models}, 
      author={Joel Lehman and Jonathan Gordon and Shawn Jain and Kamal Ndousse and Cathy Yeh and Kenneth O. Stanley},
      year={2022},
      eprint={2206.08896},
      archivePrefix={arXiv},
      primaryClass={cs.NE},
      url={https://arxiv.org/abs/2206.08896}, 
}

@misc{ong2025routellmlearningroutellms,
      title={RouteLLM: Learning to Route LLMs with Preference Data}, 
      author={Isaac Ong and Amjad Almahairi and Vincent Wu and Wei-Lin Chiang and Tianhao Wu and Joseph E. Gonzalez and M Waleed Kadous and Ion Stoica},
      year={2025},
      eprint={2406.18665},
      archivePrefix={arXiv},
      primaryClass={cs.LG},
      url={https://arxiv.org/abs/2406.18665}, 
}

@misc{valkanas2025c3pooptimizedlargelanguage,
      title={C3PO: Optimized Large Language Model Cascades with Probabilistic Cost Constraints for Reasoning}, 
      author={Antonios Valkanas and Soumyasundar Pal and Pavel Rumiantsev and Yingxue Zhang and Mark Coates},
      year={2025},
      eprint={2511.07396},
      archivePrefix={arXiv},
      primaryClass={cs.LG},
      url={https://arxiv.org/abs/2511.07396}, 
}

@misc{maheswaran2025arbitrageefficientreasoningadvantageaware,
      title={Arbitrage: Efficient Reasoning via Advantage-Aware Speculation}, 
      author={Monishwaran Maheswaran and Rishabh Tiwari and Yuezhou Hu and Kerem Dilmen and Coleman Hooper and Haocheng Xi and Nicholas Lee and Mehrdad Farajtabar and Michael W. Mahoney and Kurt Keutzer and Amir Gholami},
      year={2025},
      eprint={2512.05033},
      archivePrefix={arXiv},
      primaryClass={cs.CL},
      url={https://arxiv.org/abs/2512.05033}, 
}

@misc{yuksekgonul2026learningdiscovertesttime,
      title={Learning to Discover at Test Time}, 
      author={Mert Yuksekgonul and Daniel Koceja and Xinhao Li and Federico Bianchi and Jed McCaleb and Xiaolong Wang and Jan Kautz and Yejin Choi and James Zou and Carlos Guestrin and Yu Sun},
      year={2026},
      eprint={2601.16175},
      archivePrefix={arXiv},
      primaryClass={cs.LG},
      url={https://arxiv.org/abs/2601.16175}, 
}

@misc{cemri2026adaevolveadaptivellmdriven,
      title={AdaEvolve: Adaptive LLM Driven Zeroth-Order Optimization}, 
      author={Mert Cemri and Shubham Agrawal and Akshat Gupta and Shu Liu and Audrey Cheng and Qiuyang Mang and Ashwin Naren and Lutfi Eren Erdogan and Koushik Sen and Matei Zaharia and Alex Dimakis and Ion Stoica},
      year={2026},
      eprint={2602.20133},
      archivePrefix={arXiv},
      primaryClass={cs.NE},
      url={https://arxiv.org/abs/2602.20133}, 
}
}

\clearpage
\appendix
\clearpage
\section*{Appendix}

% A. Background & Setup

\section{Related Work}
\label{sec:related_work}

\textbf{Test-time scaling.}
Test-time scaling invests additional inference compute to improve output quality~\citep{snell2024scalingllmtesttimecompute,wu2025inferencescalinglawsempirical}, through parallel sampling~\citep{wang2023selfconsistencyimproveschainthought,brown2024largelanguagemonkeysscaling}, sequential refinement~\citep{madaan2023selfrefineiterativerefinementselffeedback,muennighoff2025s1simpletesttimescaling}, search~\citep{yao2023treethoughtsdeliberateproblem,zhang2024accessinggpt4levelmathematical,zhou2024languageagenttreesearch}, or extended reasoning chains~\citep{openai2024openaio1card,Guo_2025}.
Compute-optimal sampling with weaker models can outperform a single strong model~\citep{bansal2024smallerweakerbettertraining}, though scaling without verification remains suboptimal~\citep{setlur2025scalingtesttimecomputeverification}.
All of these operate within a single-model regime; \OURS{} extends test-time scaling to multi-model orchestration by routing evolutionary operations across models of different cost.

\textbf{Self-aggregation and recursive refinement.}
Several methods combine multiple LLM outputs into a refined answer, including RSA~\citep{venkatraman2026recursiveselfaggregationunlocksdeep}, generative self-aggregation~\citep{li2025llmsgeneratebetteranswer}, Parallel-Distill-Refine~\citep{madaan2025rethinkingthinkingtokensllms}, and Best-of-N refinement~\citep{khairi2025makingtakingbestn}.
Mixture-of-Agents~\citep{wang2024mixtureofagentsenhanceslargelanguage} layers multiple LLMs but uses a fixed model assignment rather than adaptive routing.
$\textbf{V}_{1}$~\citep{singh2026v1unifyinggenerationselfverification} demonstrates that RSA suffers from diversity collapse (monotonically declining pass@$N$) and proposes pairwise self-verification as an orthogonal remedy.
\OURS{} addresses the same bottleneck from a complementary angle: multi-model orchestration preserves diverse reasoning lineages, while confidence-based routing delegates easy aggregation groups to cheaper models.

\textbf{Verification and confidence signals.}
External verification spans outcome reward models~\citep{cobbe2021trainingverifierssolvemath}, process reward models~\citep{lightman2023letsverifystepstep}, and generative verifiers~\citep{zhang2025generativeverifiersrewardmodeling}, while self-verification can improve reasoning~\citep{weng2023largelanguagemodelsbetter}, though its benefits are situation-dependent~\citep{lu2025doesverificationpayoff}.
DeepConf~\citep{fu2025deepthinkconfidence} uses token-level confidence to filter low-quality reasoning traces, achieving large token savings.
\OURS{} uses the same class of model-intrinsic confidence signals not to filter or verify candidates, but as a routing signal that assigns each recombination group to a model, requiring no trained reward model or external evaluator.

\textbf{LLM-driven evolutionary search.}
LLMs serve as evolutionary operators for discovering programs, prompts, and algorithms~\citep{lehman2022evolutionlargemodels,funsearch,novikov2025alphaevolvecodingagentscientific, yuksekgonul2026learningdiscovertesttime}, with subsequent systems varying primarily in selection and variation strategies~\citep{openevolve,lange2025shinkaevolveopenendedsampleefficientprogram,agrawal2026gepareflectivepromptevolution,assumpo2026codeevolveopensourceevolutionary,cemri2026adaevolveadaptivellmdriven}.
EvoX~\citep{liu2026evoxmetaevolutionautomateddiscovery} meta-evolves the search strategy itself rather than fixing it.
These systems rely on external verifiers and apply a single model uniformly across all operators.
\OURS{} operates in the verifier-free regime and introduces adaptive model assignment: the evolutionary template remains unchanged, but each recombination group is routed to a model commensurate with its difficulty.

\textbf{Model routing and cost-efficient inference.}
Cascading and routing frameworks route entire queries between a strong and a weak model~\citep{ong2025routellmlearningroutellms,valkanas2025c3pooptimizedlargelanguage}.
Arbitrage~\citep{maheswaran2025arbitrageefficientreasoningadvantageaware} moves to finer granularity by routing individual reasoning steps between draft and target models, achieving 2$\times$ latency reduction.
\OURS{} routes at a similarly fine granularity but within a multi-step evolutionary pipeline: individual recombination groups are assigned to models based on per-group confidence, and because these decisions compound across loops, savings accumulate multiplicatively.
\clearpage
\section{Datasets and Benchmarks}
\label{app:datasets}

Table~\ref{tab:datasets} summarizes the benchmarks used in this work. We describe each below.

\begin{table}[h]
\centering
\small
\caption{Summary of evaluation benchmarks.}
\label{tab:datasets}
\begin{tabular}{llll}
\toprule
\textbf{Benchmark} & \textbf{Size} & \textbf{Answer Format} & \textbf{Metric} \\
\midrule
AIME 2025 & 30 & Integer (000--999) & Accuracy \\
HMMT Feb.\ 2025 & 30 & Short answer & Accuracy \\
GPQA-Diamond & 198 & 4-way MC & Accuracy \\
LiveCodeBench V6 & 175 & Code & Pass@1 \\
MMMU-Pro & 1{,}730 & Up to 10-way MC & Accuracy \\
BabyVision & 388 & Short answer & Accuracy \\
ARC-AGI-V2 & 120 & Output grid & Pass@2 \\
Circle Packing & 1 & Program & Objective \\
\bottomrule
\end{tabular}
\end{table}

\paragraph{AIME 2025~\citep{balunović2026matharenaevaluatingllmsuncontaminated}.}
The American Invitational Mathematics Examination consists of 30 problems (15 from AIME~I, 15 from AIME~II) covering algebra, geometry, number theory, and combinatorics. Each answer is an integer in $[0, 999]$, scored by exact match. We source problems via MathArena.

\paragraph{HMMT February 2025~\citep{balunović2026matharenaevaluatingllmsuncontaminated}.}
The Harvard--MIT Mathematics Tournament February competition comprises 30 individual-round problems (10 Algebra, 10 Geometry, 10 Combinatorics). Answers are short numerical or symbolic expressions, scored by exact match. We source problems via MathArena.

\paragraph{GPQA-Diamond~\citep{rein2024gpqa}.}
A 198-question subset of GPQA filtered for maximum difficulty: both domain experts answered correctly while most non-experts failed even with unrestricted web access. Questions span graduate-level biology, physics, and chemistry in 4-way multiple-choice format.

\paragraph{LiveCodeBench V6~\citep{jain2024livecodebenchholisticcontaminationfree}.}
A competitive programming benchmark sourcing problems from LeetCode, AtCoder, and Codeforces. Models generate code solutions evaluated against hidden test cases; we report pass@1. Continuous collection mitigates data contamination.

\paragraph{MMMU-Pro~\citep{yue2025mmmuprorobustmultidisciplinemultimodal}.}
A harder variant of MMMU spanning 1{,}730 multimodal questions across 30 subjects in six disciplines (Art \& Design, Business, Science, Health \& Medicine, Humanities \& Social Science, Tech \& Engineering). Answer choices are augmented from 4 to up to 10 options, and text-only solvable questions are filtered out.

\paragraph{BabyVision~\citep{chen2026babyvisionvisualreasoninglanguage}.}
A visual reasoning benchmark of 388 items across 22 subclasses in four categories: fine-grained discrimination, visual tracking, spatial perception, and visual pattern recognition. It tests core visual abilities independent of linguistic knowledge; human adults achieve 94.1\% while leading MLLMs score below 50\%. BabyVision uses an LLM-as-Judge (GPT-4o) for evaluation.

\paragraph{ARC-AGI-V2~\citep{chollet2025arcagi2}.}
A benchmark of 120 public evaluation tasks testing abstract reasoning and compositional generalization. Each task provides demonstration input--output grid pairs; the model must infer the transformation rule and produce the correct output grid. Scored by pass@2 across test pairs (exact grid match with two attempts).

\paragraph{Circle Packing ($n{=}26$)~\citep{novikov2025alphaevolvecodingagentscientific}.}
An open-ended optimization problem: pack 26 non-overlapping circles in a unit square to maximize the sum of their radii. This is a single continuous-objective instance used to evaluate evolutionary discovery capabilities. The metric is the objective value (sum of radii).

\clearpage

\section{Generation Hyperparameters}
\label{app:hyperparams}
\autoref{tab:hyperparams} lists the generation hyperparameters used for each model.
These are model-provided hyperparameters and differ from RSA except for GPT-OSS.
% ── Generation hyperparameters table ──────────────────────────────────────────
\begin{table}[h]
\centering
\caption{Generation hyperparameters for each model.}
\label{tab:hyperparams}
\begin{tabular}{@{}lcccccc@{}}
\toprule
\textbf{Model} & \textbf{Effort} & \textbf{Temp.} & \textbf{Top-K} & \textbf{Top-P} & \textbf{Min-P} & \textbf{Gen.\ Len.} \\
\midrule
Qwen3-4B-Instruct-2507        & --- & 0.7 & 20  & 0.8  & 0 & 8K  \\
Qwen3-30B-A3B-Instruct-2507   & --- & 0.7 & 20  & 0.8  & 0 & 8K  \\
Qwen3-235B-A22B-Instruct-2507 & --- & 0.7 & 20  & 0.8  & 0 & 16K \\
Qwen3-235B-A22B-Thinking-2507 & --- & 0.6 & 20  & 0.95 & 0 & 32K \\
Qwen3-30B-A3B-Thinking-2507   & --- & 0.6 & 20  & 0.95 & 0 & 32K \\
Qwen3-4B-Thinking-2507        & --- & 0.6 & 20  & 0.95 & 0 & 32K \\
Qwen3.5-35B-A3B                & --- & 1.0 & 20  & 0.95 & 0 & 64K \\
\midrule
GPT-OSS-20B                    & medium & 1   & $-1$ & 1    & 0 & 16K \\
GPT-OSS-120B                   & medium & 1   & $-1$ & 1    & 0 & 16K \\
\midrule
GPT-5 Mini                     & medium & \multicolumn{4}{c}{default} & 32K \\
\midrule
Gemini-3-Flash-Preview         & high & \multicolumn{4}{c}{default} & 64K \\
Gemini-3.1-Pro-Preview         & high & \multicolumn{4}{c}{default} & 64K \\
\midrule
Kimi-2.5-Thinking              & --- & 1.0 & 20  & 0.95 & 0 & 64K \\
Kimi-2.5-Instant               & --- & 1.0 & 20  & 0.95 & 0 & 64K \\
\bottomrule
\end{tabular}
\end{table}

\clearpage

\section{Empirical Cost Model}
\label{sec:empirical-cost}
We report per-token API pricing from commercial inference providers to
ground the routing savings of \OURS{} in real-world dollar costs.
Table~\ref{tab:pricing} lists the models used in our experiments together with
their input and output prices from Alibaba Cloud, Together~AI, Google, and OpenAI.
% -- API pricing table --------------------------------------------------------
{\small
\setlength{\tabcolsep}{3pt}
\renewcommand{\arraystretch}{1.2}
\begin{xltabular}{\textwidth}{@{}l X c c@{}}
\caption{Per-token API pricing (\$/1M tokens) for each model used in our experiments.}
\label{tab:pricing} \\
\toprule
\textbf{Provider} & \textbf{Model} & \textbf{Input price} & \textbf{Output price} \\
\midrule
\endfirsthead
\toprule
\textbf{Provider} & \textbf{Model} & \textbf{Input price} & \textbf{Output price} \\
\midrule
\endhead
\midrule
\multicolumn{4}{r}{\textit{Continued on next page}} \\
\endfoot
\bottomrule
\endlastfoot

\multirow{5}{*}{Alibaba Cloud}
  & Qwen3-30B-A3B-Instruct-2507   & \$0.108 & \$0.431 \\
  & Qwen3-30B-A3B-Thinking-2507   & \$0.108 & \$1.076 \\
  & Qwen3-235B-A22B-Instruct-2507 & \$0.287 & \$0.920 \\
  & Qwen3.5-35B-A3B               & \$0.057 & \$0.459 \\
\midrule
\multirow{2}{*}{Google Gemini API}
  & Gemini-3-Flash-Preview        & \$0.50  & \$3.00 \\
  & Gemini-3.1-Pro-Preview        & \$2.00  & \$12.00 \\
\midrule
OpenAI
  & GPT-5 Mini                    & \$0.25  & \$2.00 \\
\midrule
\multirow{3}{*}{Together AI}
  & GPT-OSS-20B                   & \$0.05  & \$0.20 \\
  & GPT-OSS-120B                  & \$0.15  & \$0.60 \\
  & Kimi-K2.5                     & \$0.50  & \$2.80 \\
\end{xltabular}
}
\clearpage

\section{Algorithm}
\label{app:algorithm}
\begin{algorithm}[h]
\caption{\OURS}
\label{alg:squeeze-evolve}
\begin{algorithmic}[1]
\Require Query set $\{Q_q\}$, Model 2: $M_2$, Model 1 $M_1$, fitness $f$, operators $\mathrm{Select}$, $\mathrm{Route}$, $\mathrm{LiteAgg}$, $\mathrm{Update}$, population size $N$, group size $K$, groups per problem $M$, loops $T$
\Ensure Final populations $\{\mathcal{P}_q^{(T)}\}$
\Statex \textbf{Loop 0 --- Initialization (Model 2 only):}
\For{each problem $q$}
    \State $\mathcal{P}_q^{(0)} \gets \{\tau_i \sim p_{M_2}(\cdot \mid Q_q)\}_{i=1}^{N}$
\EndFor
\Statex \textbf{Loops 1\ldots$T$ --- Fitness-routed evolution:}
\For{$t = 1, \ldots, T$}
    \For{each problem $q$}
        \State Score every $\tau \in \mathcal{P}_q^{(t-1)}$: compute $f(\tau)$
        \State $\mathcal{G}_q \gets \mathrm{Select}(\mathcal{P}_q^{(t-1)},\, K,\, M,\, f)$
        \State $F(g) \gets \mathrm{GroupFitness}(g, f)$ for each $g \in \mathcal{G}_q$
        \State $(\mathcal{B}_1,\, \mathcal{B}_2,\, \mathcal{B}_{\mathrm{lite}}) \gets \mathrm{Route}(\mathcal{G}_q,\, F)$
    \EndFor
    \State $\mathcal{R}_1 \gets \mathrm{Agg}(M_1, \mathcal{B}_1)$ \textbf{$\|$} $\mathcal{R}_2 \gets \mathrm{Agg}(M_2, \mathcal{B}_2)$ \textbf{$\|$} $\mathcal{R}_{\mathrm{lite}} \gets \mathrm{LiteAgg}(\mathcal{B}_{\mathrm{lite}})$
    \State $\mathcal{P}_q^{(t)} \gets \mathrm{Update}\!\bigl(\mathcal{P}_q^{(t-1)},\; \mathcal{R}_1 \cup \mathcal{R}_2 \cup \mathcal{R}_{\mathrm{lite}}\bigr)$
\EndFor
\end{algorithmic}
\end{algorithm}

\clearpage

% B. Analysis Details

\section{Full Aggregation Accuracy Results}
\label{app:agg-accuracy-full}
Accuracy rises monotonically with the number of correct seeds. The large model maintains a consistent advantage at intermediate seed counts (1--3), while both models converge at the extremes (0 and 4).
\begin{figure*}[h]
  \centering
  \begin{subfigure}[t]{0.48\linewidth}
    \centering
    \includegraphics[width=\linewidth]{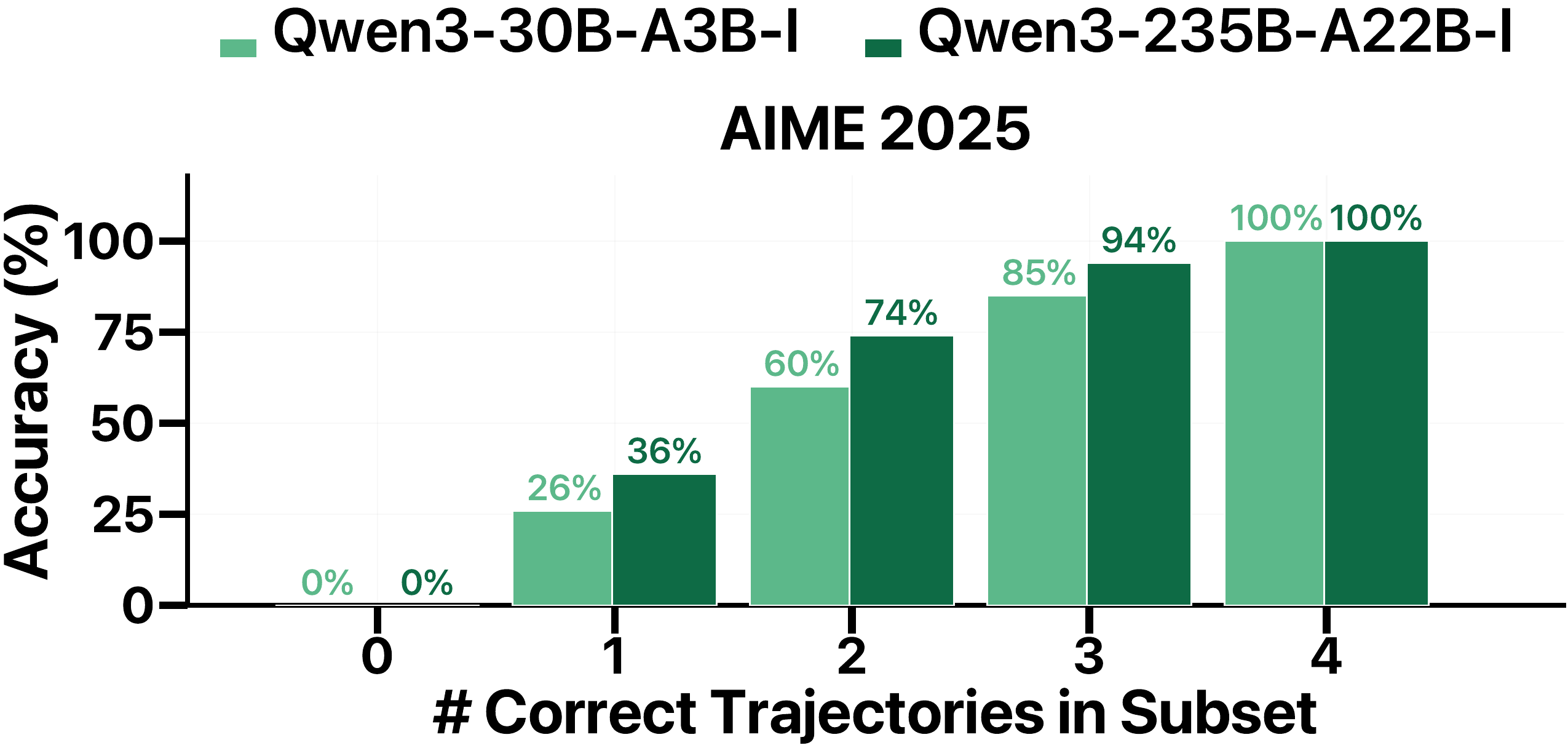}
  \end{subfigure}\hfill
  \begin{subfigure}[t]{0.48\linewidth}
    \centering
    \includegraphics[width=\linewidth]{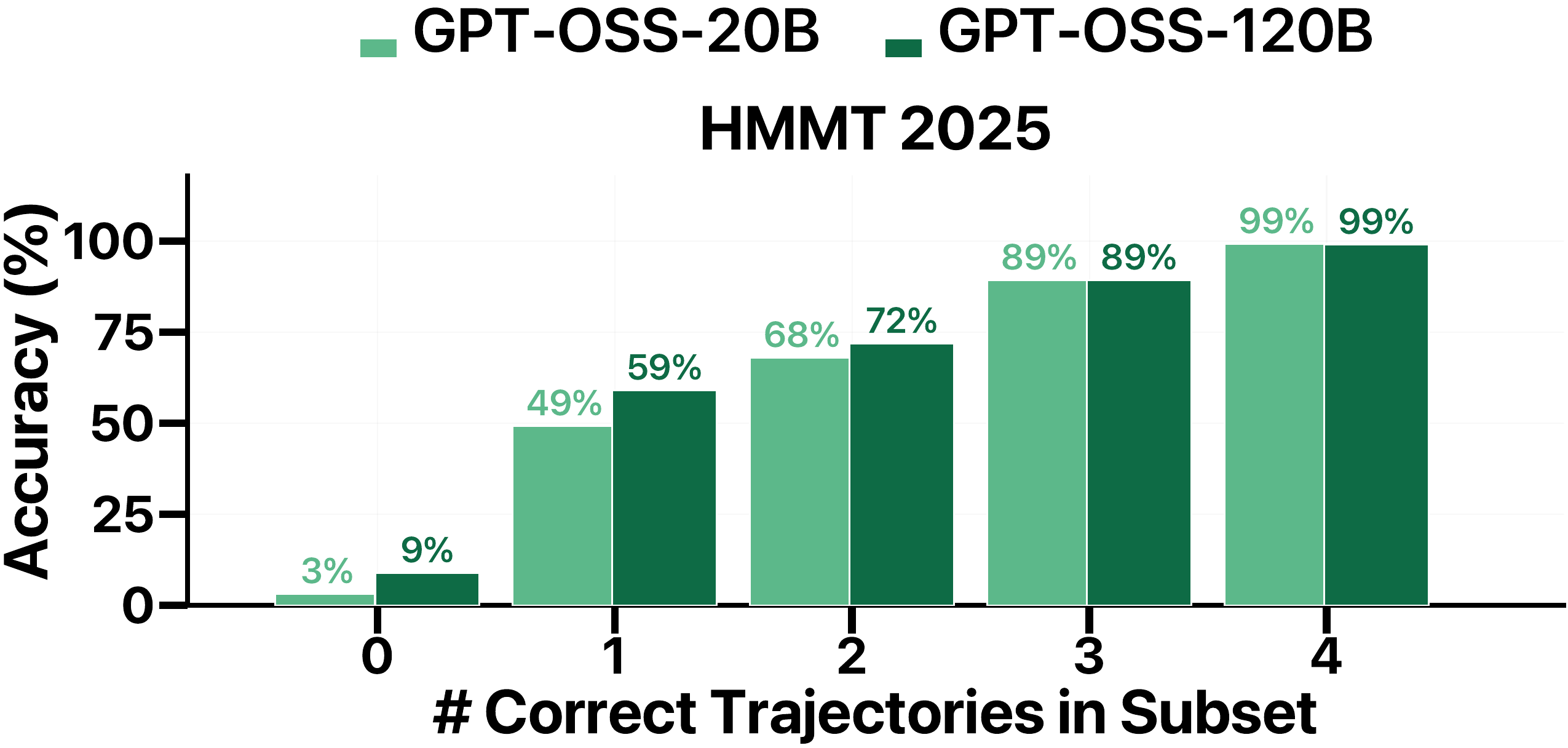}
  \end{subfigure}
  \caption{\textbf{Full aggregation accuracy vs.\ number of correct trajectories in subset.}
  \emph{Left:} AIME~2025 (Qwen3-30B-A3B-Instruct vs.\ Qwen3-235B-A22B-Instruct).
  \emph{Right:} HMMT~2025 (GPT-OSS-20B vs.\ GPT-OSS-120B).}
  \label{fig:agg-accuracy-full}
\end{figure*}

\section{Full Group Confidence Results}
\label{app:gc-full}
\begin{figure*}[h]
  \centering
  \includegraphics[width=\linewidth]{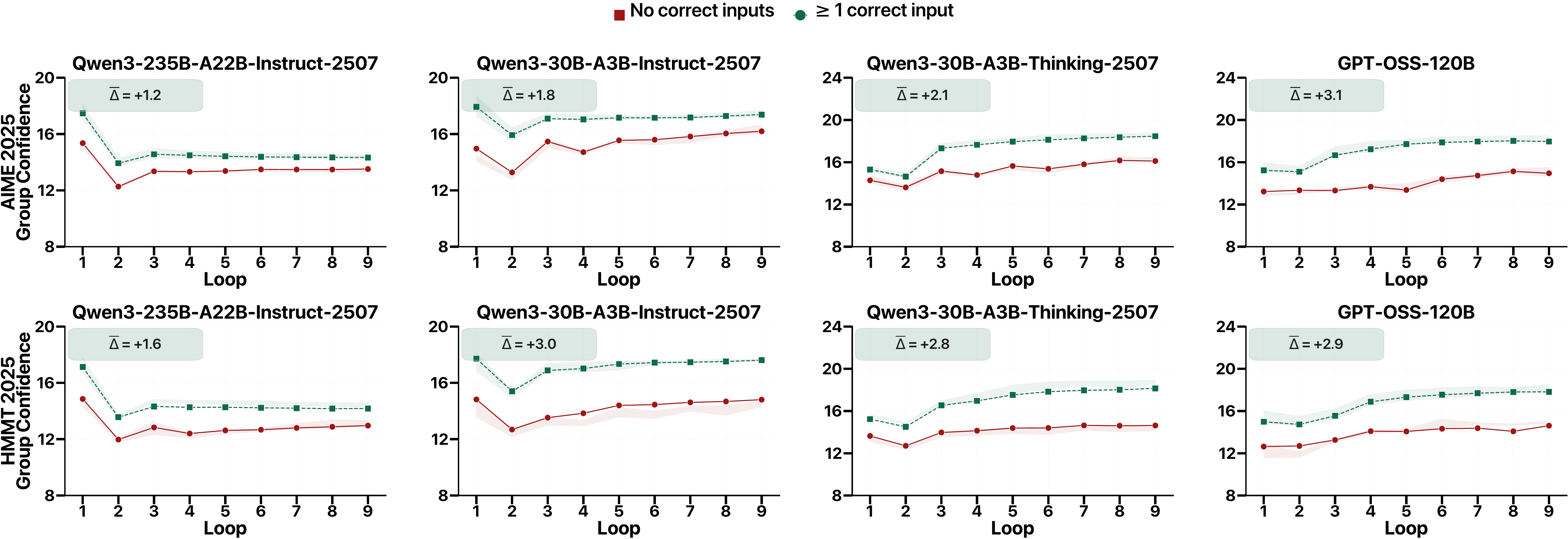}
  \caption{\textbf{Self-model group confidence by correctness across all baseline models.}
  Mean GC (with 40th--60th percentile band) across RSA loops~1--9 for four scorer models: Qwen3-235B-A22B-Instruct, Qwen3-30B-A3B-Instruct, Qwen3-30B-A3B-Thinking, and GPT-OSS-120B.
  Top row: AIME~2025; bottom row: HMMT~2025.
  Across all models and benchmarks, subsets containing correct trajectories maintain consistently higher GC than all-incorrect subsets.}
  \label{fig:gc-baseline-full}
\end{figure*}

\begin{figure*}[ht]
  \centering
  \includegraphics[width=\linewidth]{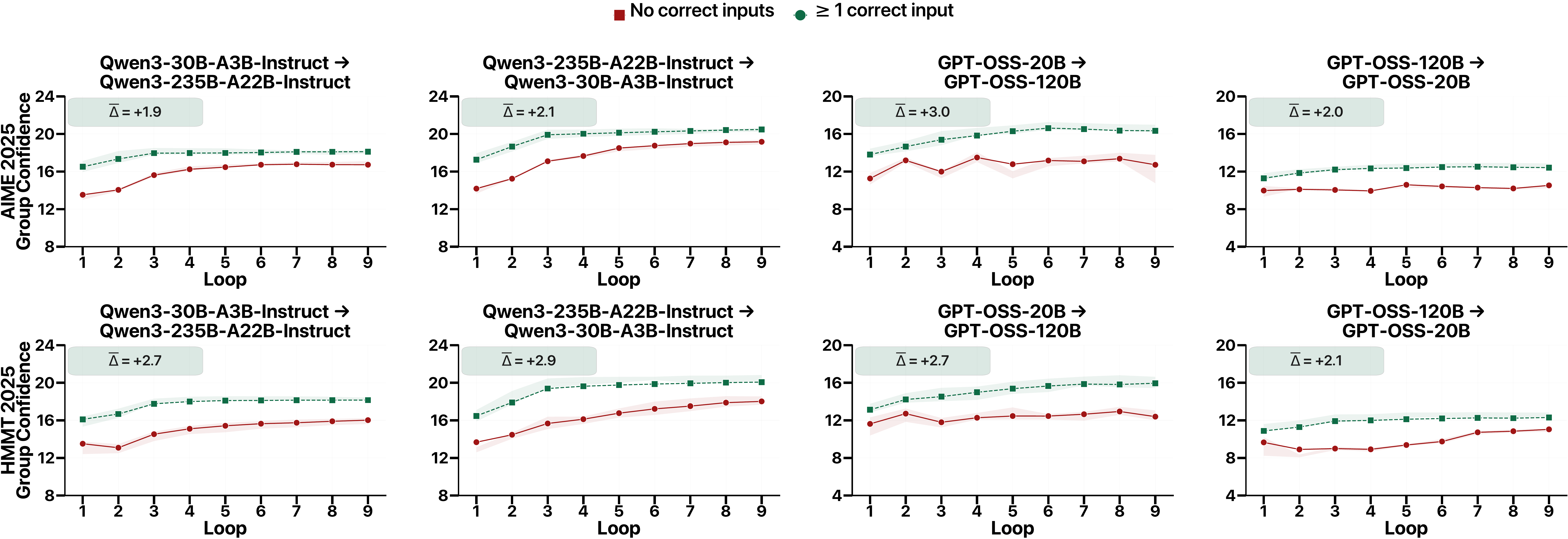}
  \caption{\textbf{Cross-model group confidence by correctness across all routing configurations.}
  Mean GC (with 40th--60th percentile band) for groups containing no correct inputs
  vs.\ groups with $\geq$1 correct input, pooled across seeds.
  Columns show four routing pairs: forward routing
  (Qwen3-30B-A3B-Instruct $\to$ Qwen3-235B-A22B-Instruct and
   GPT-OSS-20B $\to$ GPT-OSS-120B)
  and reverse routing
  (Qwen3-235B-A22B-Instruct $\to$ Qwen3-30B-A3B-Instruct and
   GPT-OSS-120B $\to$ GPT-OSS-20B).
  Top row: AIME~2025; bottom row: HMMT~2025.
  GC reliably separates correct from incorrect groups across all routing
  directions, model families, and benchmarks.}
  \label{fig:gc-routing-full}
\end{figure*}

\clearpage

% C. Empirical Cost Results

\section{Empirical Cost Results: Homogeneous Model Pairs for Reasoning Tasks}
\label{app:empirical-cost-homogeneous}
\begin{figure*}[t]
  \centering
  \begin{subfigure}[t]{0.8\textwidth}
    \centering
    \includegraphics[width=\linewidth]{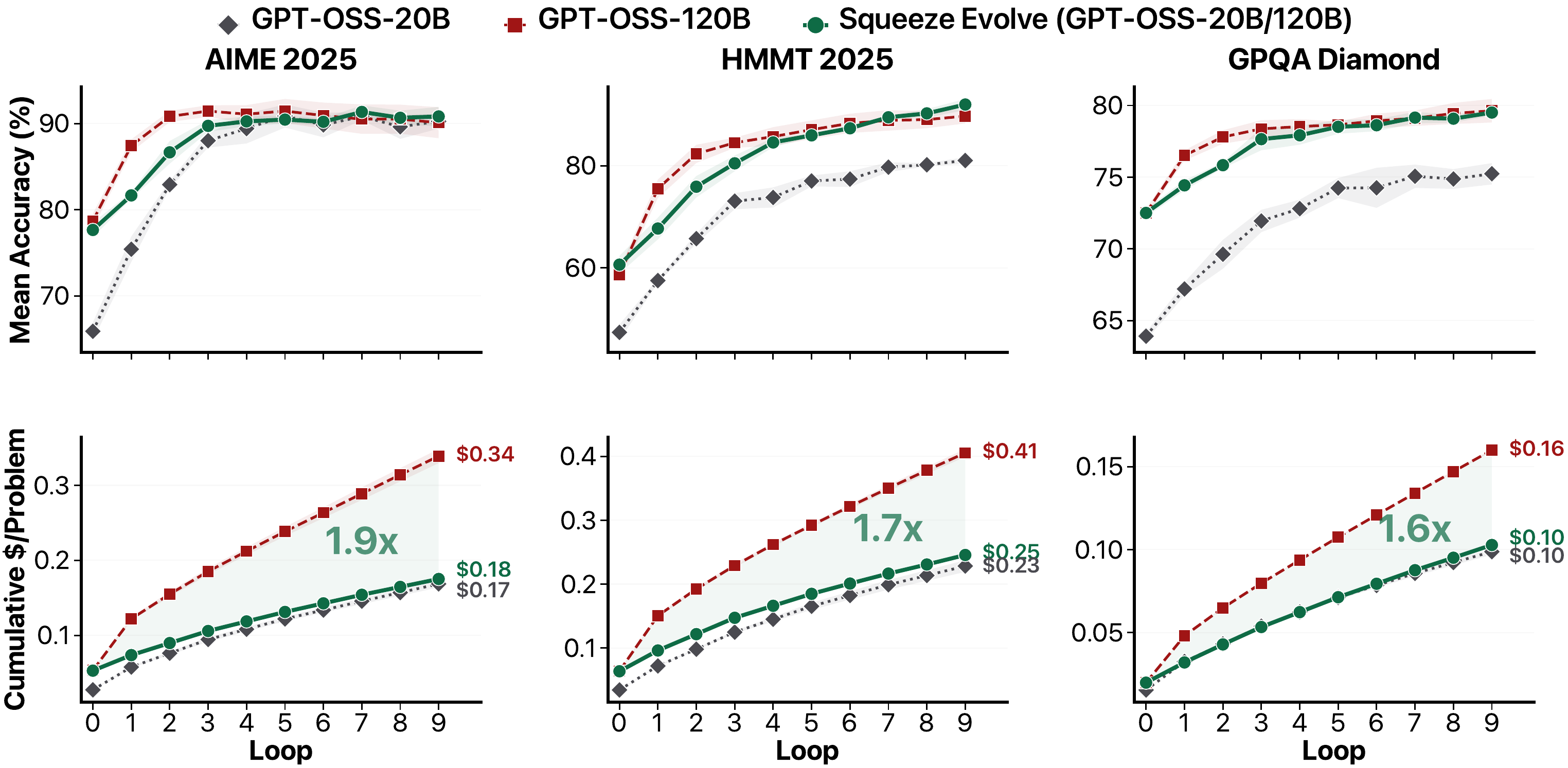}
    \caption{}
    \label{fig:emp-cost-homo-gptoss}
  \end{subfigure}

  \vspace{0.5em}

  \begin{subfigure}[t]{0.8\textwidth}
    \centering
    \includegraphics[width=\linewidth]{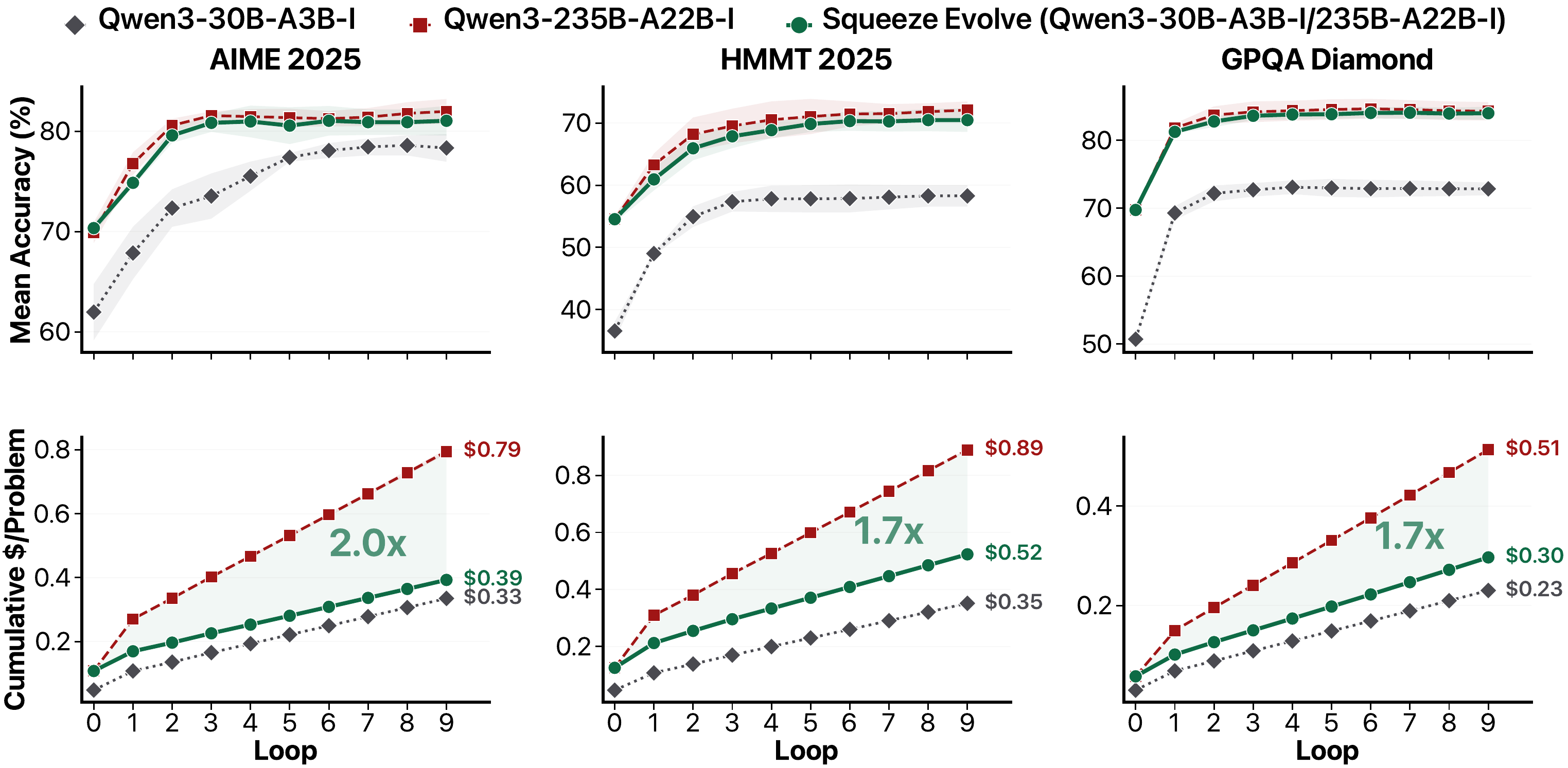}
    \caption{}
    \label{fig:emp-cost-homo-qwen-235b}
  \end{subfigure}

  \vspace{0.5em}

  \begin{subfigure}[t]{0.8\textwidth}
    \centering
    \includegraphics[width=\linewidth]{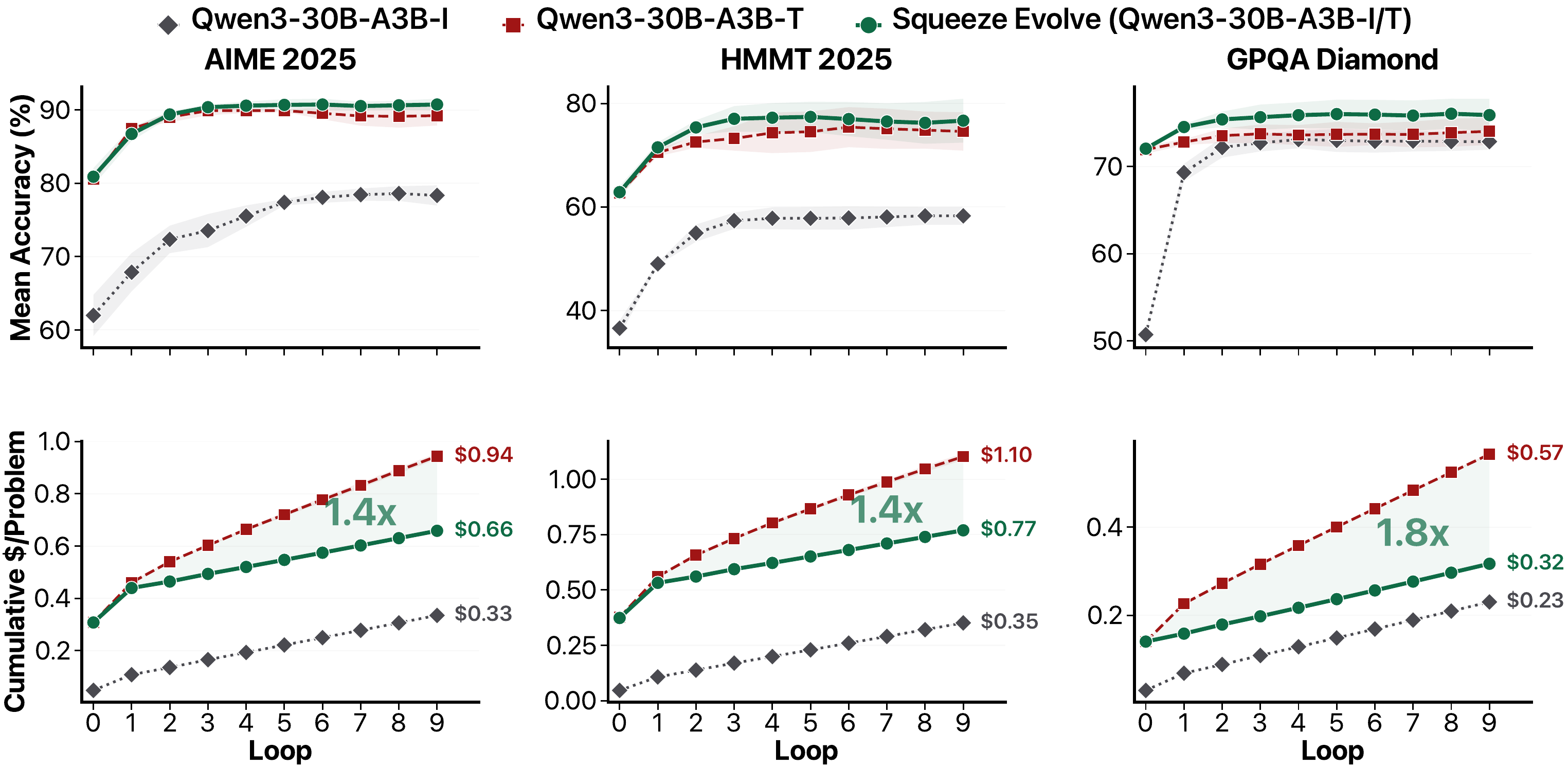}
    \caption{}
    \label{fig:emp-cost-homo-qwen-thinking}
  \end{subfigure}

  \caption{\textbf{Empirical cost results for homogeneous model pairs.}
  \emph{Top row} of each panel: mean accuracy (\%) across RSA loops.
  \emph{Bottom row}: cumulative API cost per problem (\$).
  \OURS{} (green) matches or exceeds the Model 2-only baseline (red) in accuracy while substantially reducing cost.
  The shaded region highlights the cost savings, which grow with each loop as more candidates are routed to the cheaper Model 1.}
  \label{fig:empirical-cost-homogeneous}
\end{figure*}

% -- Empirical cost results table ----------------------------------------------
{\small
\setlength{\tabcolsep}{3pt}
\renewcommand{\arraystretch}{1.0}
\begin{xltabular}{\textwidth}{@{}c l p{0.22\textwidth} p{0.22\textwidth} c c c@{}}
\caption{Empirical (dollar) cost results across datasets and model
configurations. \$/Prob is the average API cost per problem.}
\label{tab:empirical} \\
\toprule
\textbf{Data} & \textbf{Strategy} & \textbf{Model 1} & \textbf{Model 2}
& \textbf{Acc.} & \textbf{\$/Prob} & \textbf{\$ Savings} \\
\midrule
\endfirsthead
\toprule
\textbf{Data} & \textbf{Strategy} & \textbf{Model 1} & \textbf{Model 2}
& \textbf{Acc.} & \textbf{\$/Prob} & \textbf{\$ Savings} \\
\midrule
\endhead
\midrule
\multicolumn{7}{r}{\textit{Continued on next page}} \\
\endfoot
\bottomrule
\endlastfoot

% ========================= AIME25 =========================
\multirow{11}{*}{\rotatebox{90}{AIME25}}
  & RSA & Qwen3-30B-A3B-I & --- & 77.8 & \$0.33 & --- \\
  & RSA & --- & Qwen3-30B-A3B-T & 89.2 & \$0.94 & 1.0$\times$ \\
  & \cellcolor{teal!4}\OURS ($p{=}0$)  & \cellcolor{teal!4}Qwen3-30B-A3B-I & \cellcolor{teal!4}Qwen3-30B-A3B-T & \cellcolor{teal!4}90.7 & \cellcolor{teal!4}\$0.66 & \cellcolor{teal!4}1.4$\times$ \\
\cmidrule{2-7}
  & RSA & Qwen3-30B-A3B-I & --- & 77.8 & \$0.33 & --- \\
  & RSA & --- & Qwen3-235B-A22B-I & 82.0 & \$0.79 & 1.0$\times$ \\
  & \cellcolor{teal!4}\OURS ($p{=}10$) & \cellcolor{teal!4}Qwen3-30B-A3B-I & \cellcolor{teal!4}Qwen3-235B-A22B-I & \cellcolor{teal!4}80.1 & \cellcolor{teal!4}\$0.47 & \cellcolor{teal!4}1.7$\times$ \\
  & \cellcolor{teal!4}\OURS ($p{=}0$)  & \cellcolor{teal!4}Qwen3-30B-A3B-I & \cellcolor{teal!4}Qwen3-235B-A22B-I & \cellcolor{teal!4}81.0 & \cellcolor{teal!4}\$0.39 & \cellcolor{teal!4}2.0$\times$ \\
\cmidrule{2-7}
  & RSA & GPT-OSS-20B & --- & 90.0 & \$0.17 & --- \\
  & RSA & --- & GPT-OSS-120B & 90.1 & \$0.34 & 1.0$\times$ \\
  & \cellcolor{teal!4}\OURS ($p{=}10$) & \cellcolor{teal!4}GPT-OSS-20B & \cellcolor{teal!4}GPT-OSS-120B & \cellcolor{teal!4}90.5 & \cellcolor{teal!4}\$0.21 & \cellcolor{teal!4}1.6$\times$ \\
  & \cellcolor{teal!4}\OURS ($p{=}0$)  & \cellcolor{teal!4}GPT-OSS-20B & \cellcolor{teal!4}GPT-OSS-120B & \cellcolor{teal!4}90.8 & \cellcolor{teal!4}\$0.18 & \cellcolor{teal!4}1.9$\times$ \\
\midrule

% ========================= HMMT25 =========================
\multirow{11}{*}{\rotatebox{90}{HMMT25}}
  & RSA & Qwen3-30B-A3B-I & --- & 57.7 & \$0.35 & --- \\
  & RSA & --- & Qwen3-30B-A3B-T & 74.6 & \$1.10 & 1.0$\times$ \\
  & \cellcolor{teal!4}\OURS ($p{=}0$)  & \cellcolor{teal!4}Qwen3-30B-A3B-I & \cellcolor{teal!4}Qwen3-30B-A3B-T & \cellcolor{teal!4}76.7 & \cellcolor{teal!4}\$0.77 & \cellcolor{teal!4}1.4$\times$ \\
\cmidrule{2-7}
  & RSA  & Qwen3-30B-A3B-I & --- & 57.7 & \$0.35 & --- \\
  & RSA  & --- & Qwen3-235B-A22B-I & 72.1 & \$0.89 & 1.0$\times$ \\
  & \cellcolor{teal!4}\OURS ($p{=}10$) & \cellcolor{teal!4}Qwen3-30B-A3B-I & \cellcolor{teal!4}Qwen3-235B-A22B-I & \cellcolor{teal!4}71.4 & \cellcolor{teal!4}\$0.52 & \cellcolor{teal!4}1.7$\times$ \\
  & \cellcolor{teal!4}\OURS ($p{=}0$)  & \cellcolor{teal!4}Qwen3-30B-A3B-I & \cellcolor{teal!4}Qwen3-235B-A22B-I & \cellcolor{teal!4}67.4 & \cellcolor{teal!4}\$0.44 & \cellcolor{teal!4}2.0$\times$ \\
\cmidrule{2-7}
  & RSA  & GPT-OSS-20B & --- & 80.8 & \$0.23 & --- \\
  & RSA  & --- & GPT-OSS-120B & 89.7 & \$0.41 & 1.0$\times$ \\
  & \cellcolor{teal!4}\OURS ($p{=}10$) & \cellcolor{teal!4}GPT-OSS-20B & \cellcolor{teal!4}GPT-OSS-120B & \cellcolor{teal!4}92.0 & \cellcolor{teal!4}\$0.25 & \cellcolor{teal!4}1.6$\times$ \\
  & \cellcolor{teal!4}\OURS ($p{=}0$)  & \cellcolor{teal!4}GPT-OSS-20B & \cellcolor{teal!4}GPT-OSS-120B & \cellcolor{teal!4}87.9 & \cellcolor{teal!4}\$0.22 & \cellcolor{teal!4}1.8$\times$ \\
\midrule

% ====================== GPQA-Diamond ======================
\multirow{11}{*}{\rotatebox{90}{GPQA-Diamond}}
  & RSA  & Qwen3-30B-A3B-I & --- & 72.5 & \$0.23 & --- \\
  & RSA  & --- & Qwen3-30B-A3B-T & 74.0 & \$0.57 & 1.0$\times$ \\
  & \cellcolor{teal!4}\OURS ($p{=}0$)  & \cellcolor{teal!4}Qwen3-30B-A3B-I & \cellcolor{teal!4}Qwen3-30B-A3B-T & \cellcolor{teal!4}75.9 & \cellcolor{teal!4}\$0.32 & \cellcolor{teal!4}1.8$\times$ \\
\cmidrule{2-7}
  & RSA  & Qwen3-30B-A3B-I & --- & 72.5 & \$0.23 & --- \\
  & RSA  & --- & Qwen3-235B-A22B-I & 84.3 & \$0.51 & 1.0$\times$ \\
  & \cellcolor{teal!4}\OURS ($p{=}10$) & \cellcolor{teal!4}Qwen3-30B-A3B-I & \cellcolor{teal!4}Qwen3-235B-A22B-I & \cellcolor{teal!4}84.0 & \cellcolor{teal!4}\$0.30 & \cellcolor{teal!4}1.7$\times$ \\
  & \cellcolor{teal!4}\OURS ($p{=}0$)  & \cellcolor{teal!4}Qwen3-30B-A3B-I & \cellcolor{teal!4}Qwen3-235B-A22B-I & \cellcolor{teal!4}83.8 & \cellcolor{teal!4}\$0.25 & \cellcolor{teal!4}2.1$\times$ \\
\cmidrule{2-7}
  & RSA  & GPT-OSS-20B & --- & 75.0 & \$0.10 & --- \\
  & RSA  & --- & GPT-OSS-120B & 79.6 & \$0.16 & 1.0$\times$ \\
  & \cellcolor{teal!4}\OURS ($p{=}10$) & \cellcolor{teal!4}GPT-OSS-20B & \cellcolor{teal!4}GPT-OSS-120B & \cellcolor{teal!4}79.5 & \cellcolor{teal!4}\$0.10 & \cellcolor{teal!4}1.6$\times$ \\
  & \cellcolor{teal!4}\OURS ($p{=}0$)  & \cellcolor{teal!4}GPT-OSS-20B & \cellcolor{teal!4}GPT-OSS-120B & \cellcolor{teal!4}79.0 & \cellcolor{teal!4}\$0.09 & \cellcolor{teal!4}1.9$\times$ \\
\midrule

% ========================= LCB-V6 =========================
\multirow{12}{*}{\rotatebox{90}{LCB-V6}}
  & RSA  & Qwen3-30B-A3B-I & --- & 46.1 & \$0.19 & --- \\
  & RSA  & --- & Qwen3-30B-A3B-T & 64.2 & \$0.82 & 1.0$\times$ \\
  & \cellcolor{teal!4}\OURS ($p{=}10$) & \cellcolor{teal!4}Qwen3-30B-A3B-I & \cellcolor{teal!4}Qwen3-30B-A3B-T & \cellcolor{teal!4}62.7 & \cellcolor{teal!4}\$0.63 & \cellcolor{teal!4}1.3$\times$ \\
  & \cellcolor{teal!4}\OURS ($p{=}0$)  & \cellcolor{teal!4}Qwen3-30B-A3B-I & \cellcolor{teal!4}Qwen3-30B-A3B-T & \cellcolor{teal!4}59.1 & \cellcolor{teal!4}\$0.51 & \cellcolor{teal!4}1.6$\times$ \\
\cmidrule{2-7}
  & RSA  & Qwen3-30B-A3B-I & --- & 46.1 & \$0.19 & --- \\
  & RSA  & --- & Qwen3-235B-A22B-I & 59.1 & \$0.33 & 1.0$\times$ \\
  & \cellcolor{teal!4}\OURS ($p{=}10$) & \cellcolor{teal!4}Qwen3-30B-A3B-I & \cellcolor{teal!4}Qwen3-235B-A22B-I & \cellcolor{teal!4}55.9 & \cellcolor{teal!4}\$0.22 & \cellcolor{teal!4}1.5$\times$ \\
  & \cellcolor{teal!4}\OURS ($p{=}0$)  & \cellcolor{teal!4}Qwen3-30B-A3B-I & \cellcolor{teal!4}Qwen3-235B-A22B-I & \cellcolor{teal!4}55.3 & \cellcolor{teal!4}\$0.19 & \cellcolor{teal!4}1.7$\times$ \\
\cmidrule{2-7}
  & RSA  & GPT-OSS-20B & --- & 68.9 & \$0.14 & --- \\
  & RSA  & --- & GPT-OSS-120B & 75.9 & \$0.44 & 1.0$\times$ \\
  & \cellcolor{teal!4}\OURS ($p{=}10$) & \cellcolor{teal!4}GPT-OSS-20B & \cellcolor{teal!4}GPT-OSS-120B & \cellcolor{teal!4}75.6 & \cellcolor{teal!4}\$0.22 & \cellcolor{teal!4}2.0$\times$ \\
  & \cellcolor{teal!4}\OURS ($p{=}0$)  & \cellcolor{teal!4}GPT-OSS-20B & \cellcolor{teal!4}GPT-OSS-120B & \cellcolor{teal!4}73.3 & \cellcolor{teal!4}\$0.18 & \cellcolor{teal!4}2.4$\times$ \\
\end{xltabular}
}

\clearpage

\section{Empirical Cost Results: Heterogeneous Model Pairs for Reasoning Tasks}
\label{app:empirical-cost-heterogeneous}
\begin{figure*}[t]
  \centering
  \begin{subfigure}[t]{0.8\textwidth}
    \centering
    \includegraphics[width=\linewidth]{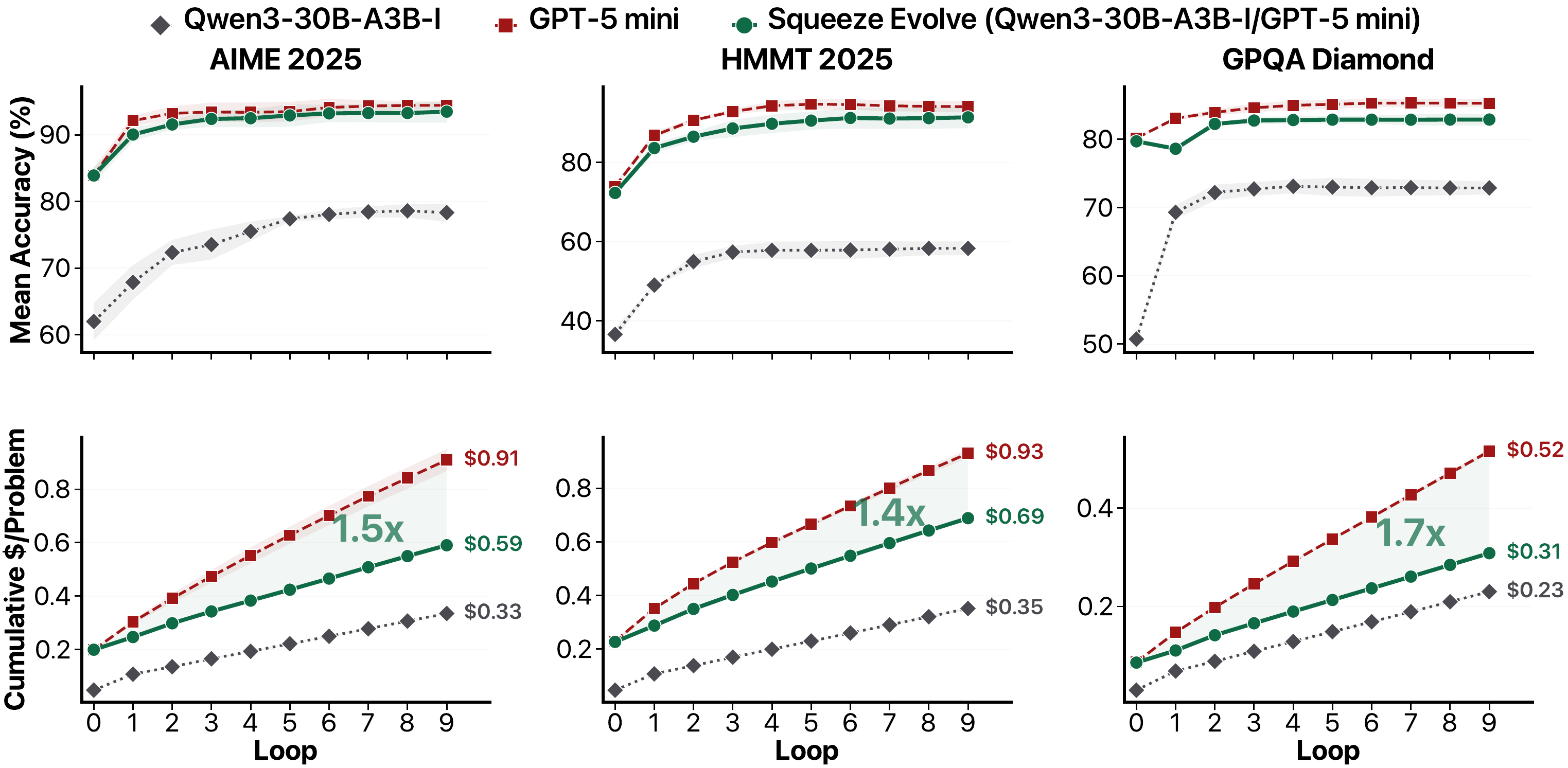}
    \caption{}
    \label{fig:emp-cost-hetero-qwen-gpt5mini}
  \end{subfigure}
  \vspace{0.5em}
  \begin{subfigure}[t]{0.8\textwidth}
    \centering
    \includegraphics[width=\linewidth]{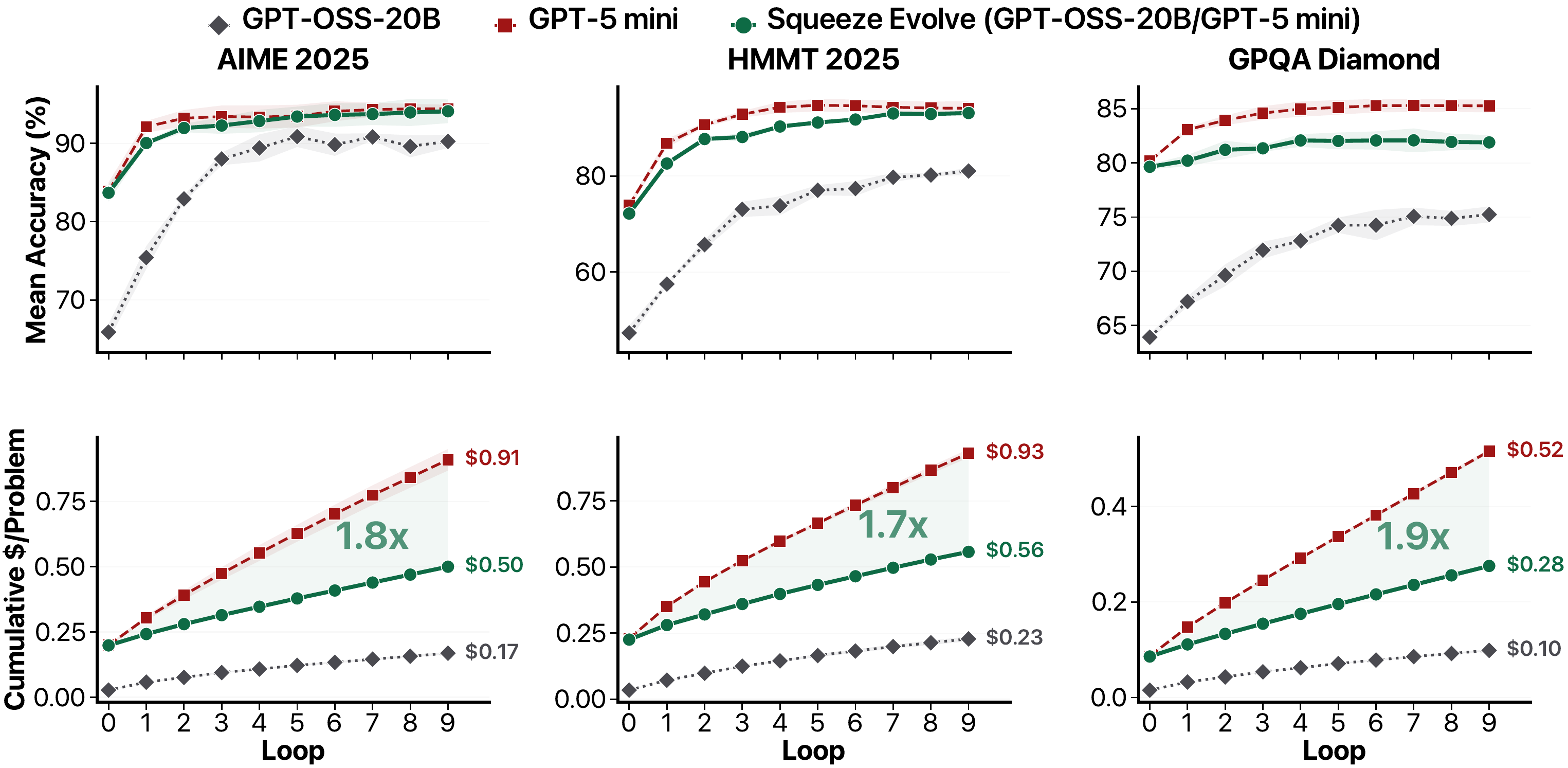}
    \caption{}
    \label{fig:emp-cost-hetero-gptoss-gpt5mini}
  \end{subfigure}

  \caption{\textbf{Empirical cost results for heterogeneous model pairs.}
  \emph{Top row} of each panel: mean accuracy (\%) across RSA loops.
  \emph{Bottom row}: cumulative API cost per problem (\$).
  \OURS{} (green) matches or exceeds the Model 2 baseline (red) in accuracy while substantially reducing cost.
  The shaded region highlights the cost savings, which grow with each loop as more candidates are routed to the cheaper Model 1.}
  \label{fig:empirical-cost-heterogeneous}
\end{figure*}

% -- Empirical cost results table ----------------------------------------------
{\small
\setlength{\tabcolsep}{3pt}
\renewcommand{\arraystretch}{1.0}
\begin{xltabular}{\textwidth}{@{}c l p{0.22\textwidth} p{0.20\textwidth} c c c@{}}
\caption{Empirical (dollar) cost results across datasets and model
configurations. \$/Prob is the average API cost per problem.}
\label{tab:empirical-hetero} \\
\toprule
\textbf{Data} & \textbf{Strategy} & \textbf{Model 1} & \textbf{Model 2}
& \textbf{Acc.} & \textbf{\$/Prob} & \textbf{\$ Savings} \\
\midrule
\endfirsthead
\toprule
\textbf{Data} & \textbf{Strategy} & \textbf{Model 1} & \textbf{Model 2}
& \textbf{Acc.} & \textbf{\$/Prob} & \textbf{\$ Savings} \\
\midrule
\endhead
\midrule
\multicolumn{7}{r}{\textit{Continued on next page}} \\
\endfoot
\bottomrule
\endlastfoot

% ========================= AIME25 =========================
\multirow{12}{*}{\rotatebox{90}{AIME25}}
  & RSA & Qwen3-30B-A3B-I & --- & 78.8 & \$0.34 & --- \\
  & RSA & --- & GPT-5 mini & 94.2 & \$0.89 & 1.0$\times$ \\
  & \cellcolor{teal!4}\OURS ($p{=}30$) & \cellcolor{teal!4}Qwen3-30B-A3B-I & \cellcolor{teal!4}GPT-5 mini & \cellcolor{teal!4}93.5 & \cellcolor{teal!4}\$0.64 & \cellcolor{teal!4}1.4$\times$ \\
  & \cellcolor{teal!4}\OURS ($p{=}20$) & \cellcolor{teal!4}Qwen3-30B-A3B-I & \cellcolor{teal!4}GPT-5 mini & \cellcolor{teal!4}93.5 & \cellcolor{teal!4}\$0.59 & \cellcolor{teal!4}1.5$\times$ \\
  & \cellcolor{teal!4}\OURS ($p{=}10$) & \cellcolor{teal!4}Qwen3-30B-A3B-I & \cellcolor{teal!4}GPT-5 mini & \cellcolor{teal!4}93.1 & \cellcolor{teal!4}\$0.53 & \cellcolor{teal!4}1.7$\times$ \\
  & \cellcolor{teal!4}\OURS ($p{=}0$) & \cellcolor{teal!4}Qwen3-30B-A3B-I & \cellcolor{teal!4}GPT-5 mini & \cellcolor{teal!4}91.9 & \cellcolor{teal!4}\$0.46 & \cellcolor{teal!4}2.0$\times$ \\
\cmidrule{2-7}
  & RSA & GPT-OSS-20B & --- & 90.6 & \$0.17 & --- \\
  & RSA & --- & GPT-5 mini & 94.2 & \$0.89 & 1.0$\times$ \\
  & \cellcolor{teal!4}\OURS ($p{=}30$) & \cellcolor{teal!4}GPT-OSS-20B & \cellcolor{teal!4}GPT-5 mini & \cellcolor{teal!4}95.4 & \cellcolor{teal!4}\$0.50 & \cellcolor{teal!4}1.8$\times$ \\
  & \cellcolor{teal!4}\OURS ($p{=}20$) & \cellcolor{teal!4}GPT-OSS-20B & \cellcolor{teal!4}GPT-5 mini & \cellcolor{teal!4}94.6 & \cellcolor{teal!4}\$0.46 & \cellcolor{teal!4}1.9$\times$ \\
  & \cellcolor{teal!4}\OURS ($p{=}10$) & \cellcolor{teal!4}GPT-OSS-20B & \cellcolor{teal!4}GPT-5 mini & \cellcolor{teal!4}92.8 & \cellcolor{teal!4}\$0.39 & \cellcolor{teal!4}2.3$\times$ \\
  & \cellcolor{teal!4}\OURS ($p{=}0$) & \cellcolor{teal!4}GPT-OSS-20B & \cellcolor{teal!4}GPT-5 mini & \cellcolor{teal!4}92.7 & \cellcolor{teal!4}\$0.30 & \cellcolor{teal!4}3.0$\times$ \\
\midrule

% ========================= HMMT25 =========================
\multirow{12}{*}{\rotatebox{90}{HMMT25}}
  & RSA & Qwen3-30B-A3B-I & --- & 58.9 & \$0.35 & --- \\
  & RSA & --- & GPT-5 mini & 93.3 & \$0.94 & 1.0$\times$ \\
  & \cellcolor{teal!4}\OURS ($p{=}30$) & \cellcolor{teal!4}Qwen3-30B-A3B-I & \cellcolor{teal!4}GPT-5 mini & \cellcolor{teal!4}93.1 & \cellcolor{teal!4}\$0.69 & \cellcolor{teal!4}1.4$\times$ \\
  & \cellcolor{teal!4}\OURS ($p{=}20$) & \cellcolor{teal!4}Qwen3-30B-A3B-I & \cellcolor{teal!4}GPT-5 mini & \cellcolor{teal!4}90.2 & \cellcolor{teal!4}\$0.66 & \cellcolor{teal!4}1.4$\times$ \\
  & \cellcolor{teal!4}\OURS ($p{=}10$) & \cellcolor{teal!4}Qwen3-30B-A3B-I & \cellcolor{teal!4}GPT-5 mini & \cellcolor{teal!4}88.1 & \cellcolor{teal!4}\$0.59 & \cellcolor{teal!4}1.6$\times$ \\
  & \cellcolor{teal!4}\OURS ($p{=}0$) & \cellcolor{teal!4}Qwen3-30B-A3B-I & \cellcolor{teal!4}GPT-5 mini & \cellcolor{teal!4}87.6 & \cellcolor{teal!4}\$0.51 & \cellcolor{teal!4}1.9$\times$ \\
\cmidrule{2-7}
  & RSA & GPT-OSS-20B & --- & 81.2 & \$0.22 & --- \\
  & RSA & --- & GPT-5 mini & 93.3 & \$0.94 & 1.0$\times$ \\
  & \cellcolor{teal!4}\OURS ($p{=}30$) & \cellcolor{teal!4}GPT-OSS-20B & \cellcolor{teal!4}GPT-5 mini & \cellcolor{teal!4}93.1 & \cellcolor{teal!4}\$0.56 & \cellcolor{teal!4}1.7$\times$ \\
  & \cellcolor{teal!4}\OURS ($p{=}20$) & \cellcolor{teal!4}GPT-OSS-20B & \cellcolor{teal!4}GPT-5 mini & \cellcolor{teal!4}91.8 & \cellcolor{teal!4}\$0.51 & \cellcolor{teal!4}1.8$\times$ \\
  & \cellcolor{teal!4}\OURS ($p{=}10$) & \cellcolor{teal!4}GPT-OSS-20B & \cellcolor{teal!4}GPT-5 mini & \cellcolor{teal!4}89.8 & \cellcolor{teal!4}\$0.43 & \cellcolor{teal!4}2.2$\times$ \\
  & \cellcolor{teal!4}\OURS ($p{=}0$) & \cellcolor{teal!4}GPT-OSS-20B & \cellcolor{teal!4}GPT-5 mini & \cellcolor{teal!4}89.3 & \cellcolor{teal!4}\$0.35 & \cellcolor{teal!4}2.7$\times$ \\
\midrule

% ========================= GPQA-DIAMOND =========================
\multirow{12}{*}{\rotatebox{90}{GPQA-Diamond}}
  & RSA & Qwen3-30B-A3B-I & --- & 73.3 & \$0.23 & --- \\
  & RSA & --- & GPT-5 mini & 85.0 & \$0.52 & 1.0$\times$ \\
  & \cellcolor{teal!4}\OURS ($p{=}30$) & \cellcolor{teal!4}Qwen3-30B-A3B-I & \cellcolor{teal!4}GPT-5 mini & \cellcolor{teal!4}82.6 & \cellcolor{teal!4}\$0.37 & \cellcolor{teal!4}1.4$\times$ \\
  & \cellcolor{teal!4}\OURS ($p{=}20$) & \cellcolor{teal!4}Qwen3-30B-A3B-I & \cellcolor{teal!4}GPT-5 mini & \cellcolor{teal!4}83.6 & \cellcolor{teal!4}\$0.35 & \cellcolor{teal!4}1.5$\times$ \\
  & \cellcolor{teal!4}\OURS ($p{=}10$) & \cellcolor{teal!4}Qwen3-30B-A3B-I & \cellcolor{teal!4}GPT-5 mini & \cellcolor{teal!4}83.2 & \cellcolor{teal!4}\$0.31 & \cellcolor{teal!4}1.7$\times$ \\
  & \cellcolor{teal!4}\OURS ($p{=}0$) & \cellcolor{teal!4}Qwen3-30B-A3B-I & \cellcolor{teal!4}GPT-5 mini & \cellcolor{teal!4}82.2 & \cellcolor{teal!4}\$0.26 & \cellcolor{teal!4}2.0$\times$ \\
\cmidrule{2-7}
  & RSA & GPT-OSS-20B & --- & 75.5 & \$0.10 & --- \\
  & RSA & --- & GPT-5 mini & 85.0 & \$0.52 & 1.0$\times$ \\
  & \cellcolor{teal!4}\OURS ($p{=}30$) & \cellcolor{teal!4}GPT-OSS-20B & \cellcolor{teal!4}GPT-5 mini & \cellcolor{teal!4}82.1 & \cellcolor{teal!4}\$0.27 & \cellcolor{teal!4}1.9$\times$ \\
  & \cellcolor{teal!4}\OURS ($p{=}20$) & \cellcolor{teal!4}GPT-OSS-20B & \cellcolor{teal!4}GPT-5 mini & \cellcolor{teal!4}81.8 & \cellcolor{teal!4}\$0.25 & \cellcolor{teal!4}2.1$\times$ \\
  & \cellcolor{teal!4}\OURS ($p{=}10$) & \cellcolor{teal!4}GPT-OSS-20B & \cellcolor{teal!4}GPT-5 mini & \cellcolor{teal!4}80.5 & \cellcolor{teal!4}\$0.20 & \cellcolor{teal!4}2.5$\times$ \\
  & \cellcolor{teal!4}\OURS ($p{=}0$) & \cellcolor{teal!4}GPT-OSS-20B & \cellcolor{teal!4}GPT-5 mini & \cellcolor{teal!4}78.8 & \cellcolor{teal!4}\$0.16 & \cellcolor{teal!4}3.3$\times$ \\

\end{xltabular}
}

\clearpage

\section{Empirical Cost Results: Homogeneous Model Pairs for Vision Tasks}
\label{app:empirical-vision-homogeneous}
% -- Empirical cost results: Vision tasks (homogeneous) -----------------------
{\small
\setlength{\tabcolsep}{3pt}
\renewcommand{\arraystretch}{1.3}
\begin{xltabular}{\textwidth}{@{}c l p{0.22\textwidth} p{0.22\textwidth} c c c@{}}
\caption{Empirical (dollar) cost results for vision tasks with homogeneous model pairs.
\$/Prob is the average API cost per problem.}
\label{tab:empirical-vision-tasks} \\
\toprule
\textbf{Data} & \textbf{Strategy} & \textbf{Model 1} & \textbf{Model 2}
& \textbf{Acc.} & \textbf{\$/Prob} & \textbf{\$ Savings} \\
\midrule
\endfirsthead
\toprule
\textbf{Data} & \textbf{Strategy} & \textbf{Model 1} & \textbf{Model 2}
& \textbf{Acc.} & \textbf{\$/Prob} & \textbf{\$ Savings} \\
\midrule
\endhead
\midrule
\multicolumn{7}{r}{\textit{Continued on next page}} \\
\endfoot
\bottomrule
\endlastfoot

% ========================= MMMU-PRO =========================
\multirow{3}{*}{\rotatebox{90}{\scriptsize MMMU-Pro}}
  & RSA & Kimi-2.5-Instant & --- & 77.46 & \$0.29 & --- \\
  & RSA & --- & Kimi-2.5-Thinking & 78.58 & \$1.04 & 1.0$\times$ \\
  & \cellcolor{teal!4}\OURS ($p{=}0$) & \cellcolor{teal!4}Kimi-2.5-Instant & \cellcolor{teal!4}Kimi-2.5-Thinking & \cellcolor{teal!4}78.63 & \cellcolor{teal!4}\$0.58 & \cellcolor{teal!4}1.8$\times$ \\
\midrule

% ========================= BABYVISION =========================
\multirow{3}{*}{\rotatebox{90}{\scriptsize BabyVision}}
  & RSA & Kimi-2.5-Instant & --- & 36.61 & \$0.29 & --- \\
  & RSA & --- & Kimi-2.5-Thinking & 43.23 & \$2.05 & 1.0$\times$ \\
  & \cellcolor{teal!4}\OURS ($p{=}0$) & \cellcolor{teal!4}Kimi-2.5-Instant & \cellcolor{teal!4}Kimi-2.5-Thinking & \cellcolor{teal!4}41.56 & \cellcolor{teal!4}\$0.81 & \cellcolor{teal!4}2.5$\times$ \\

\end{xltabular}
}

\section{Empirical Cost Results: Heterogeneous Model Pairs for Vision Tasks}
\label{app:empirical-vision-heterogeneous}
% -- Empirical cost results: Vision tasks (heterogeneous) ---------------------
{\small
\setlength{\tabcolsep}{3pt}
\renewcommand{\arraystretch}{1.3}
\begin{xltabular}{\textwidth}{@{}c l p{0.22\textwidth} p{0.20\textwidth} c c c@{}}
\caption{Empirical (dollar) cost results for vision tasks with heterogeneous model pairs.
\$/Prob is the average API cost per problem.
$^\dagger$Image tokens are only processed by the Model 2 in loop~0; subsequent loops use Model 1 as a text-only causal LM (no image input).}
\label{tab:empirical-hetero-tasks} \\
\toprule
\textbf{Data} & \textbf{Strategy} & \textbf{Model 1} & \textbf{Model 2}
& \textbf{Acc.} & \textbf{\$/Prob} & \textbf{\$ Savings} \\
\midrule
\endfirsthead
\toprule
\textbf{Data} & \textbf{Strategy} & \textbf{Model 1} & \textbf{Model 2}
& \textbf{Acc.} & \textbf{\$/Prob} & \textbf{\$ Savings} \\
\midrule
\endhead
\midrule
\multicolumn{7}{r}{\textit{Continued on next page}} \\
\endfoot
\bottomrule
\endlastfoot

% ========================= MMMU-PRO =========================
\multirow{3}{*}{\rotatebox{90}{\scriptsize MMMU-Pro}}
  & RSA & Qwen3.5-35B-A3B$^\dagger$ & --- & --- & --- & --- \\
  & RSA & --- & Kimi-2.5-Thinking & 78.58 & \$1.04 & 1.0$\times$ \\
  & \cellcolor{teal!4}\OURS ($p{=}0$) & \cellcolor{teal!4}Qwen3.5-35B-A3B$^\dagger$ & \cellcolor{teal!4}Kimi-2.5-Thinking & \cellcolor{teal!4}79.06 & \cellcolor{teal!4}\$0.46 & \cellcolor{teal!4}2.3$\times$ \\
\midrule

% ========================= BABYVISION =========================
\multirow{3}{*}{\rotatebox{90}{\scriptsize BabyVision}}
  & RSA & Qwen3.5-35B-A3B$^\dagger$ & --- & --- & --- & --- \\
  & RSA & --- & Kimi-2.5-Thinking & 43.23 & \$2.05 & 1.0$\times$ \\
  & \cellcolor{teal!4}\OURS ($p{=}0$) & \cellcolor{teal!4}Qwen3.5-35B-A3B$^\dagger$ & \cellcolor{teal!4}Kimi-2.5-Thinking & \cellcolor{teal!4}41.27 & \cellcolor{teal!4}\$0.83 & \cellcolor{teal!4}2.5$\times$ \\

\end{xltabular}
}

\clearpage

% D. System Implementation Details

\section{Prefill Engine Microbenchmark}
\label{app:prefill-engine}
\begin{figure}[H]
  \centering
  \includegraphics[width=0.7\columnwidth]{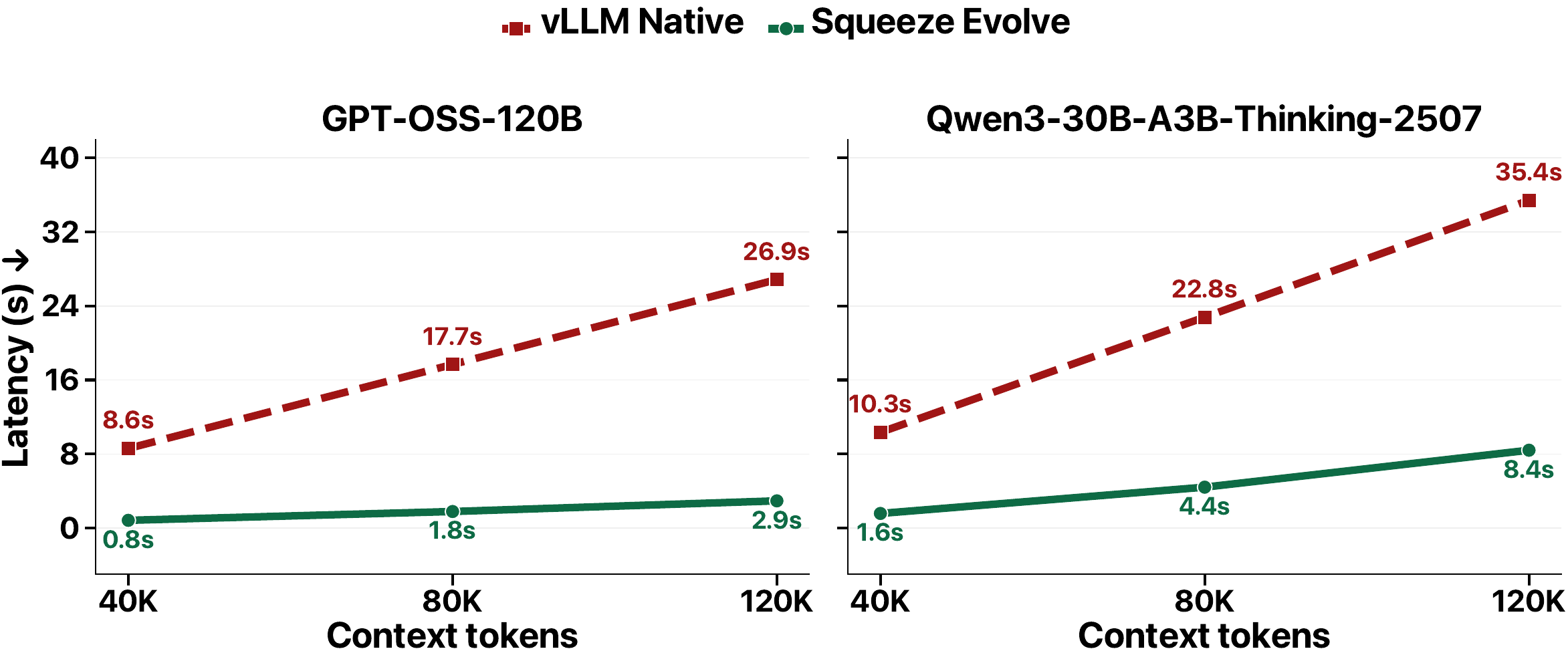}
  \caption{\textbf{Prefill engine speedup over native vLLM.}
  Confidence-scoring latency as a function of context length for GPT-OSS-120B (left) and Qwen3-30B-A3B-Thinking-2507 (right).
  The \OURS{} custom prefill engine (green) computes the confidence scalar on GPU and returns only the final score, achieving $4$--$10{\times}$ lower latency than the native vLLM prompt-logprob path (red), which materializes full token-level tensors and transfers them over HTTP.
  Speedup is larger for the 120B model where serialization and transfer costs dominate.
  Both paths scale approximately linearly with context length, but the custom engine has a substantially gentler slope.}
  \label{fig:prefill-engine-bench}
\end{figure}

As described in Section~\ref{sec:system-impl}, the \OURS{} custom prefill engine computes the confidence scalar directly on GPU and returns only a single float per request, bypassing the native vLLM path that materializes full token-level logprob tensors and serializes them over HTTP. Table~\ref{tab:prefill-engine-bench} reports per-request confidence-scoring latency for both paths across two models (GPT-OSS-120B and Qwen3-30B-A3B-Thinking-2507) and three context lengths (40K, 80K, and 120K tokens), measured at batch size~1 with 3 trials per configuration. The custom engine achieves $9.1$--$10.3{\times}$ speedup on GPT-OSS-120B and $4.2$--$6.6{\times}$ on Qwen3-30B-A3B-Thinking. The larger speedup on the 120B model reflects the heavier serialization and transfer burden at larger model scales: the native path transfers ${\sim}$13\,MB of token-level logprob data per request regardless of model size, while the custom engine returns only ${\sim}$100\,bytes. At the largest scale (Qwen3-235B-A22B), the native path OOMs entirely when materializing full prompt-logprob tensors, making the custom engine a prerequisite for confidence-based routing at this model size. Figure~\ref{fig:prefill-engine-bench} shows that latency scales approximately linearly with context length under both paths, but the custom engine's slope is substantially gentler. This is because the native path's cost is dominated by tensor materialization and HTTP transfer, both of which grow with sequence length, whereas the custom engine performs an in-place reduction on GPU and transmits only the scalar result.
\begin{table}[h]
\begin{center}
\small
\setlength{\tabcolsep}{4pt}
\renewcommand{\arraystretch}{0.98}
\caption{\textbf{Prefill engine microbenchmark.}
Confidence-scoring latency (seconds, mean over 3 trials) for native vLLM and vLLM \OURS{}, measured at batch size~1 across three context lengths.
vLLM \OURS{} computes the confidence scalar on GPU and returns only a single float per request, reducing transfer volume from ${\sim}$13\,MB to ${\sim}$100\,bytes.
Native vLLM OOMs on Qwen3-235B-A22B because materializing full prompt-logprob tensors at this model scale exceeds memory; vLLM \OURS{} does not exhibit this issue.}
\label{tab:prefill-engine-bench}
\begin{tabular}{p{0.2\linewidth} c c c c}
\toprule
Model & Context & vLLM native (s) & vLLM \OURS{} (s) & Speedup \\
\midrule
GPT-OSS-120B & 40K  & 8.60  & 0.83 & 10.3$\times$ \\
             & 80K  & 17.68 & 1.79 & 9.9$\times$  \\
             & 120K & 26.86 & 2.94 & 9.1$\times$  \\
\midrule
Qwen3-30B-A3B & 40K  & 10.34 & 1.58 & 6.6$\times$ \\
                            & 80K  & 22.79 & 4.42 & 5.2$\times$ \\
                            & 120K & 35.42 & 8.41 & 4.2$\times$ \\
\bottomrule
\end{tabular}
\end{center}
\end{table}

%------------------------------------------------------------------------------------
\clearpage
\section{Routing Overhead Results}
\label{app:routing-overhead}

\paragraph{Experimental setup.}
For each benchmark and model $M_2$, we fix the full inference configuration: evolution parameters $(N=16, K=4, T=10)$, prompts, decoding limits, hardware, serving engine, and batching policy.
We compare two conditions.
\textbf{RSA-$M_2$} is standard RSA with all calls executed by $M_2$ and no routing logic.
\textbf{\OURS{}-$M_2$} enables confidence scoring and threshold computation, but forces every aggregation call to $M_2$.
This second condition preserves the routing machinery while removing any latency change due to sending work to $M_1$.
The difference between RSA-$M_2$ and \OURS{}-$M_2$ therefore isolates the routing overhead itself. This setup is also a conservative worst-case measurement: although \OURS{} normally reduces latency by routing a subset of aggregations to $M_1$, here all aggregation work is still pinned to $M_2$.

\paragraph{Measurement protocol.}
We measure latency in a single-request setting, processing one problem at a time so that routing overhead is not confounded by cross-request queueing effects.
Within each problem, however, we preserve the production execution strategy: the $N$ confidence-scoring calls and aggregation calls for a loop are batched exactly as in normal serving.
For each problem we log end-to-end latency, per-loop routing time, and per-loop aggregation time.
We repeat the measurement across the evaluation set and report mean latency.

\paragraph{Overhead definition.}
Let $T_{\mathrm{RSA}}$ denote the end-to-end latency of RSA-$M_2$ and let $T_{\mathrm{SQE}}$ denote the latency of \OURS{}-$M_2$.
We define the absolute routing overhead as
\[
\Delta_{\mathrm{route}} = T_{\mathrm{SQE}} - T_{\mathrm{RSA}}.
\]
We define the relative routing overhead as
\[
\mathrm{Overhead}(\%) = 100 \times \frac{T_{\mathrm{SQE}} - T_{\mathrm{RSA}}}{T_{\mathrm{RSA}}}.
\]
At the loop level, we decompose latency as
\[
T_{\mathrm{loop}} = T_{\mathrm{routing}} + \sum_{i=1}^{N} T_{\mathrm{agg}}(m_i),
\]
where $T_{\mathrm{routing}}$ includes both prefill-only confidence scoring and percentile thresholding / dispatch, and $T_{\mathrm{agg}}(m_i)$ is the aggregation time for the selected model $m_i \in \{M_1, M_2\}$.
In this section, all aggregations are forced to $M_2$, so the observed gap isolates the latency overhead of routing logic alone.

\begin{figure}[h]
\centering
\includegraphics[width=\columnwidth]{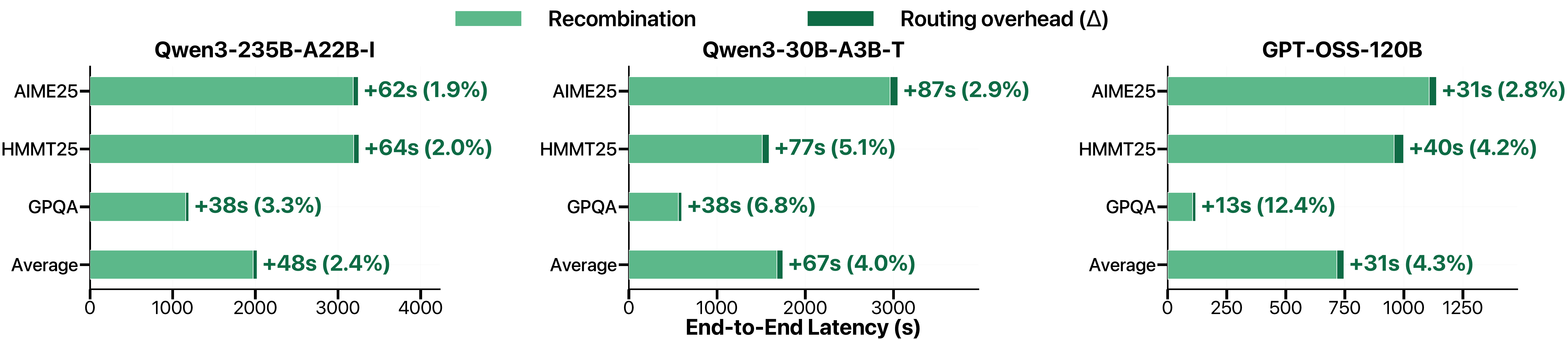}
\caption{\textbf{Routing overhead is minimal.}
Routing adds 1.9--6.8\% to end-to-end latency for the Qwen models and 2.8--12.4\% for GPT-OSS-120B, with the higher relative overhead on GPQA reflecting its short absolute generation time (106s).
Full results in Table~\ref{tab:routing-overhead} (Appendix).}
\label{fig:routing-overhead}
\end{figure}

\paragraph{Results.}
Table~\ref{tab:routing-overhead} reports per-benchmark routing overhead for each model $M_2$.
Across all configurations, routing adds 1.9\text{--}6.8\% to end-to-end latency for the Qwen models and 2.8\text{--}12.4\% for GPT-OSS-120B, with cross-model averages of 2.4\text{--}4.3\%.
The higher relative overhead on GPQA-Diamond for GPT-OSS-120B (12.4\%) reflects its short absolute generation time (106s), which makes the fixed routing cost proportionally larger.
\begin{table*}[h]
\begin{center}
\small
\setlength{\tabcolsep}{3pt}
\renewcommand{\arraystretch}{1.0}
\caption{Routing-overhead measurements in the single-request setting.
Comparing RSA-$M_2$ and \OURS{}-$M_2$ isolates the latency overhead of confidence scoring and dispatch.
Highlighted average rows summarize per-target-model averages across benchmarks.}
\label{tab:routing-overhead}
\begin{tabular}{l l c c c c}
\toprule
$M_L$ & Benchmark & RSA (s) & \OURS{} (s) & $\Delta_{\mathrm{route}}$ (s) & Overhead (\%) \\
\midrule
\multirow{5}{*}{Qwen3-30B-A3B-T}
  & AIME25       & 2961.72 & 3048.44 & 86.72 & 2.93\% \\
  & HMMT25       & 1512.39 & 1589.61 & 77.22 & 5.11\% \\
  & GPQA-Diamond & 561.45  & 599.43  & 37.98 & 6.76\% \\
  & LCB-V6       & 1798.91      & 1864.46      & 65.55    & 3.64\% \\
  \cmidrule(l){2-6}
  & \cellcolor{teal!4}Average
  & \cellcolor{teal!4}1678.52
  & \cellcolor{teal!4}1745.83
  & \cellcolor{teal!4}67.31
  & \cellcolor{teal!4}4.01\% \\
\midrule
\multirow{5}{*}{Qwen3-235B-A22B-I}
  & AIME25       & 3184.11 & 3246.08 & 61.97 & 1.95\% \\
  & HMMT25       & 3190.07 & 3253.75 & 63.68 & 2.00\% \\
  & GPQA-Diamond & 1157.17 & 1195.57 & 38.40 & 3.32\% \\
  & LCB-V6       & 359.22  & 385.93  & 26.71 & 7.44\% \\
  \cmidrule(l){2-6}
  & \cellcolor{teal!4}Average
  & \cellcolor{teal!4}1972.64
  & \cellcolor{teal!4}2020.33
  & \cellcolor{teal!4}47.69
  & \cellcolor{teal!4}2.42\% \\
\midrule
\multirow{5}{*}{GPT-OSS-120B}
  & AIME25       & 1107.44 & 1138.35 & 30.91 & 2.79\% \\
  & HMMT25       & 958.92  & 999.24  & 40.32 & 4.20\% \\
  & GPQA-Diamond & 105.74  & 118.87  & 13.13 & 12.42\% \\
  & LCB-V6       & 691.80  & 729.70  & 37.90 & 5.47\% \\
  \cmidrule(l){2-6}
  & \cellcolor{teal!4}Average
  & \cellcolor{teal!4}715.98
  & \cellcolor{teal!4}746.54
  & \cellcolor{teal!4}30.57
  & \cellcolor{teal!4}4.27\% \\
\bottomrule
\end{tabular}
\end{center}
\end{table*}

%------------------------------------------------------------------------------------

\section{System Throughput Results}
\label{app:throughput}

\paragraph{Fairness rule.}
We compare RSA and \OURS{} under the same total GPU budget $G$.
RSA allocates all $G$ GPUs to Model~2.
\OURS{} partitions the same budget into a Model~2 pool $G_2$ and a Model~1 pool $G_1$, subject to
\[
G_2 + G_1 = G.
\]
This fixed-budget constraint makes the throughput comparison deployment-fair.

\paragraph{Operating points.}
We sweep the routing percentile $p$ from Section~\ref{sec:method} across several values. Because the realized Model~1 routing share does not exactly equal $p/100$; we report both.

\paragraph{Pool sizing.}
Given a configured percentile $p$ and its observed routing mix, we size the two pools so that their loop service times are approximately matched.
Let $T_2(G_2)$ denote the time for the Model~2 pool to process the groups routed to Model~2, and let $T_1(G_1)$ denote the corresponding time for the Model~1 pool.
We choose integer $G_2$ and $G_1$ satisfying $G_2 + G_1 = G$ and minimizing
\[
\left| T_2(G_2) - T_1(G_1) \right|.
\]
This latency-matching rule avoids provisioning a fast idle pool while the slower pool remains the throughput bottleneck.

\paragraph{Measurement protocol.}
After fixing the GPU split, we drive the system with enough concurrent requests to keep serving saturated and measure steady-state throughput.
We report requests per second rather than tokens per second because Model~1 and Model~2 produce different numbers of output tokens for the same query, making a token-based metric an unfair comparison across routing configurations.
Let $N_{\mathrm{req}}$ denote the total number of requests completed by the system over wall-clock interval $\Delta t$.
We define throughput as
\[
\mathrm{Req/s} = \frac{N_{\mathrm{req}}}{\Delta t}.
\]
We use the same query stream, prompts, and serving engine for both RSA and \OURS{}, and report completed requests per second after warmup.
Because confidence-based routing causes Model~1 and Model~2 to observe different input and output lengths, we fix the input context length and output context length for each model to the values observed under the corresponding routing share.

\begin{figure*}[h]
\centering
\includegraphics[width=\linewidth]{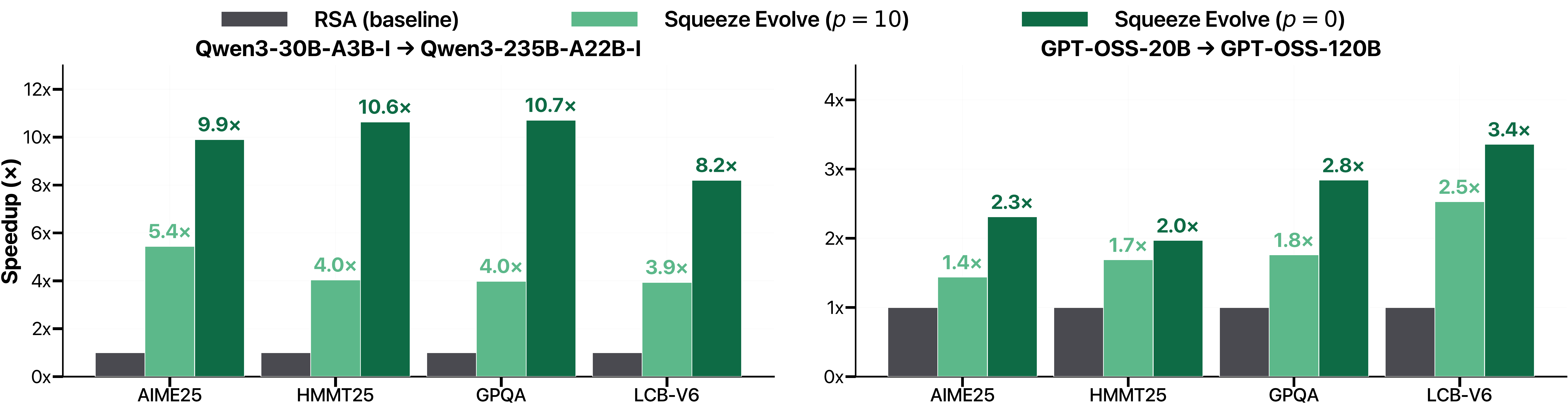}
\caption{\textbf{Fixed-budget throughput speedup over RSA.}
Under the same total GPU budget, the Qwen pair achieves 4--10$\times$ speedup and the GPT-OSS pair 1.4--3.4$\times$.
Full results in Table~\ref{tab:throughput} (Appendix).}
\label{fig:throughput}
\end{figure*}

\paragraph{Results.}
Table~\ref{tab:throughput} reports steady-state throughput under a fixed total GPU budget $G$ for two model pairs across four benchmarks.
For each routing percentile $p$, the GPU budget is partitioned into large-model and small-model pools with approximately matched loop service times.
The Accuracy column reports mean accuracy from Table~\ref{tab:empirical}.
\begin{table*}[h]
\begin{center}
\small
\caption{Fixed-budget system throughput under saturated serving.
RSA and \OURS{} use the same total GPU budget $G$.
For each routing percentile $p$, \OURS{} partitions the budget into Model 2 and Model 1 pools with approximately matched loop service times under the observed routing mix.
Throughput is reported as steady-state completed requests per second after warmup.
Accuracy is mean accuracy (\%) from Table~\ref{tab:empirical}.}
\label{tab:throughput}
\resizebox{\textwidth}{!}{%
\begin{tabular}{l l c l c c c c c}
\toprule
Model 1 & Model 2 & Benchmark & Strategy & Obs.\ $M_1$ share & GPU split $(G_2{:}G_1)$ & Req/s & Speedup & Acc.\ (\%) \\
\midrule
\multirow{12}{*}{\makecell[l]{Qwen3-30B-A3B-I}}
  & \multirow{12}{*}{\makecell[l]{Qwen3-235B-A22B-I}}
  & \multirow{3}{*}{AIME25}
  & RSA & 0\% & 16:0 & 1.36 & 1.00$\times$ & 82.0 \\
  &  &  & \cellcolor{teal!4}\OURS{} ($p{=}10$) & \cellcolor{teal!4}88.9\% & \cellcolor{teal!4}8:8 & \cellcolor{teal!4}7.41 & \cellcolor{teal!4}5.44$\times$ & \cellcolor{teal!4}80.1 \\
  &  &  & \cellcolor{teal!10}\OURS{} ($p{=}0$) & \cellcolor{teal!10}100\% & \cellcolor{teal!10}0:16 & \cellcolor{teal!10}13.47 & \cellcolor{teal!10}9.90$\times$ & \cellcolor{teal!10}81.0 \\
\cmidrule(l){3-9}
  &  & \multirow{3}{*}{HMMT25}
  & RSA & 0\% & 16:0 & 1.23 & 1.00$\times$ & 72.1 \\
  &  &  & \cellcolor{teal!4}\OURS{} ($p{=}10$) & \cellcolor{teal!4}87.5\% & \cellcolor{teal!4}8:8 & \cellcolor{teal!4}4.95 & \cellcolor{teal!4}4.04$\times$ & \cellcolor{teal!4}71.4 \\
  &  &  & \cellcolor{teal!10}\OURS{} ($p{=}0$) & \cellcolor{teal!10}100\% & \cellcolor{teal!10}0:16 & \cellcolor{teal!10}13.02 & \cellcolor{teal!10}10.63$\times$ & \cellcolor{teal!10}67.4 \\
\cmidrule(l){3-9}
  &  & \multirow{3}{*}{GPQA Diamond}
  & RSA & 0\% & 16:0 & 2.05 & 1.00$\times$ & 84.3 \\
  &  &  & \cellcolor{teal!4}\OURS{} ($p{=}10$) & \cellcolor{teal!4}87.5\% & \cellcolor{teal!4}8:8 & \cellcolor{teal!4}8.17 & \cellcolor{teal!4}3.98$\times$ & \cellcolor{teal!4}84.0 \\
  &  &  & \cellcolor{teal!10}\OURS{} ($p{=}0$) & \cellcolor{teal!10}100\% & \cellcolor{teal!10}0:16 & \cellcolor{teal!10}22.00 & \cellcolor{teal!10}10.71$\times$ & \cellcolor{teal!10}83.8 \\
\cmidrule(l){3-9}
  &  & \multirow{3}{*}{LCB-V6}
  & RSA & 0\% & 16:0 & 3.83 & 1.00$\times$ & 59.1 \\
  &  &  & \cellcolor{teal!4}\OURS{} ($p{=}10$) & \cellcolor{teal!4}87.5\% & \cellcolor{teal!4}8:8 & \cellcolor{teal!4}15.07 & \cellcolor{teal!4}3.93$\times$ & \cellcolor{teal!4}55.9 \\
  &  &  & \cellcolor{teal!10}\OURS{} ($p{=}0$) & \cellcolor{teal!10}100\% & \cellcolor{teal!10}0:16 & \cellcolor{teal!10}31.42 & \cellcolor{teal!10}8.20$\times$ & \cellcolor{teal!10}55.3 \\
\midrule
\multirow{12}{*}{GPT-OSS-20B}
  & \multirow{12}{*}{GPT-OSS-120B}
  & \multirow{3}{*}{AIME25}
  & RSA & 0\% & 20:0 & 17.09 & 1.00$\times$ & 90.1 \\
  &  &  & \cellcolor{teal!4}\OURS{} ($p{=}10$) & \cellcolor{teal!4}87.5\% & \cellcolor{teal!4}4:16 & \cellcolor{teal!4}24.59 & \cellcolor{teal!4}1.44$\times$ & \cellcolor{teal!4}90.5 \\
  &  &  & \cellcolor{teal!10}\OURS{} ($p{=}0$) & \cellcolor{teal!10}100\% & \cellcolor{teal!10}0:20 & \cellcolor{teal!10}39.43 & \cellcolor{teal!10}2.31$\times$ & \cellcolor{teal!10}90.8 \\
\cmidrule(l){3-9}
  &  & \multirow{3}{*}{HMMT25}
  & RSA & 0\% & 12:0 & 8.56 & 1.00$\times$ & 89.7 \\
  &  &  & \cellcolor{teal!4}\OURS{} ($p{=}10$) & \cellcolor{teal!4}87.5\% & \cellcolor{teal!4}4:8 & \cellcolor{teal!4}14.50 & \cellcolor{teal!4}1.69$\times$ & \cellcolor{teal!4}92.0 \\
  &  &  & \cellcolor{teal!10}\OURS{} ($p{=}0$) & \cellcolor{teal!10}100\% & \cellcolor{teal!10}0:12 & \cellcolor{teal!10}16.83 & \cellcolor{teal!10}1.97$\times$ & \cellcolor{teal!10}87.9 \\
\cmidrule(l){3-9}
  &  & \multirow{3}{*}{GPQA Diamond}
  & RSA & 0\% & 16:0 & 30.34 & 1.00$\times$ & 79.6 \\
  &  &  & \cellcolor{teal!4}\OURS{} ($p{=}10$) & \cellcolor{teal!4}87.5\% & \cellcolor{teal!4}4:12 & \cellcolor{teal!4}53.54 & \cellcolor{teal!4}1.76$\times$ & \cellcolor{teal!4}79.5 \\
  &  &  & \cellcolor{teal!10}\OURS{} ($p{=}0$) & \cellcolor{teal!10}100\% & \cellcolor{teal!10}0:16 & \cellcolor{teal!10}86.30 & \cellcolor{teal!10}2.84$\times$ & \cellcolor{teal!10}79.0 \\
\cmidrule(l){3-9}
  &  & \multirow{3}{*}{LCB-V6}
  & RSA & 0\% & 12:0 & 5.66 & 1.00$\times$ & 75.9 \\
  &  &  & \cellcolor{teal!4}\OURS{} ($p{=}10$) & \cellcolor{teal!4}87.1\% & \cellcolor{teal!4}4:8 & \cellcolor{teal!4}14.30 & \cellcolor{teal!4}2.53$\times$ & \cellcolor{teal!4}75.6 \\
  &  &  & \cellcolor{teal!10}\OURS{} ($p{=}0$) & \cellcolor{teal!10}100\% & \cellcolor{teal!10}0:12 & \cellcolor{teal!10}19.02 & \cellcolor{teal!10}3.36$\times$ & \cellcolor{teal!10}73.3 \\
\bottomrule
\end{tabular}%
}
\end{center}
\end{table*}

\clearpage

% E. Extended Task Results
\section{ARC-AGI-V2 Complete Results}
\begin{table}[h]
\centering
\caption{ARC-AGI-V2 full results. Ancestor model:
Gemini~3.1~Pro (High Thinking), $N{=}4$, $K{=}2$ $T{=}10$.
$^\dagger$Uses code execution and program synthesis.}
\label{tab:arc-agi-v2-full}
{\small
\setlength{\tabcolsep}{3pt}
\renewcommand{\arraystretch}{1.05}
\begin{tabular}{@{}l l l c r r@{}}
\toprule
\textbf{Strategy} & \textbf{Model 1} & \textbf{Model 2}
& \textbf{Acc.} & \textbf{\$/Task} & \textbf{Savings} \\
\midrule
\multicolumn{6}{@{}l}{\textit{Human baseline}} \\
Human panel & --- & --- & 100.0 & \$17.00 & --- \\
\midrule
\multicolumn{6}{@{}l}{\textit{Single-shot baselines}} \\
GPT-5.4 Pro (xhigh) & --- & --- & 92.2 & \$17.60 & --- \\
Gemini~3.1~Pro (High) & --- & --- & 88.1 & \$0.98 & --- \\
GPT-5.4 (xhigh) & --- & --- & 84.2 & \$1.57 & --- \\
Claude Opus~4.6 (Thinking 120K, high) & --- & --- & 79.0 & \$3.81 & --- \\
GPT-5.4 (high) & --- & --- & 75.8 & \$1.08 & --- \\
\midrule
\multicolumn{6}{@{}l}{\textit{Code-execution methods}} \\
Imbue + Gemini~3.1~Pro$^\dagger$ & --- & --- & 95.1 & \$8.71 & --- \\
Confluence Lab$^\dagger$ & --- & --- & 97.9 & \$11.77 & --- \\
\midrule
RSA & --- & Gemini~3.0~Flash & 45.0 & \$9.83 & --- \\
RSA & --- & Gemini~3.1~Pro & 93.3 & \$28.85 & 1.0$\times$ \\
\rowcolor{teal!4}
\OURS & --- & Gemini~3.1~Pro & 97.5 & \$7.74 & 3.7$\times$ \\
\bottomrule
\end{tabular}
}
\end{table}
\clearpage

\section{Circle Packing Complete Results}
\label{sec:circle_packing}
\subsection{Algorithm summary}
\label{sec:algorithm-summary}

The evolved algorithm (Section~\ref{sec:code}) combines three strategies:
(1)~a diverse initialization ensemble that generates hundreds of candidate
center layouts via hexagonal lattices, greedy farthest-point insertion,
jittered grids, and random placements, scoring each with an exact linear
programmed (LP) that maximizes $\sum r_i$ for fixed centres;
(2)~a hybrid optimization pipeline integrating LP-guided simulated annealing
with SLSQP gradient-based refinement, where each stochastic perturbation of
1--3 centers is immediately followed by an LP solve to obtain provably optimal
radii, and an adaptive temperature schedule prevents premature convergence
before a final SLSQP pass jointly optimizes all~$3N{=}78$ variables under
wall-distance and non-overlap constraints; and
(3)~a principled decomposition that separates the hard combinatorial center
placement ($\mathbb{R}^{52}$) from the easy convex radius assignment (an LP).

\subsection{Hyperparameters}
\label{sec:circle_packing_params}

We instantiate \OURS with GPT-OSS-120B and GPT-OSS-20B as M2 and M1 models, use group confidence as the fitness signal, fitness-weighted sampling
($\zeta{=}0.5$) for selection, a fixed confidence threshold at the 50th
percentile for routing, and the \emph{accumulate} update rule
(Table~\ref{tab:instantiations}), with $N{=}128$, $K{=}4$, $T{=}50$.
At termination, we draw $N$ candidates from the cumulative pool via
confidence-weighted sampling and report the highest circle packing score.

\subsection{Source code}
\label{sec:code}

\begin{lstlisting}[caption={Evolved circle packing program ($n{=}26$).},label=lst:circle-packing]
#!/usr/bin/env python3
"""
Optimised packing of 26 non-overlapping circles inside the unit square.

The algorithm combines:
  * Several diverse initialisation strategies (hexagonal lattice,
    farthest point, random grid, pure random).
  * A cheap but exact linear programme (LP) that, for any fixed set of
    centre positions, yields the maximal radii (maximising the sum of radii).
  * Stochastic hill-climbing with a temperature schedule to refine the
    centre positions.
  * A final local optimisation with SciPy's SLSQP optimiser, which moves
    centres and radii simultaneously while respecting all constraints.
  * A tiny post-processing step that enforces strict feasibility.

The program prints exactly 26 lines "x y r" (nine decimal digits each)
to stdout.
"""

import sys
import numpy as np

try:
    from scipy.optimize import linprog, minimize
    _SCIPY = True
except Exception:
    _SCIPY = False

N = 26
PAIR_I, PAIR_J = np.triu_indices(N, 1)
M_PAIRS = len(PAIR_I)

# Pre-computed constraint matrix for the LP:
# one row per pair (r_i + r_j <= d_ij)
A_UB = np.zeros((M_PAIRS, N), dtype=float)
A_UB[np.arange(M_PAIRS), PAIR_I] = 1.0
A_UB[np.arange(M_PAIRS), PAIR_J] = 1.0

def _fallback_radii(centres: np.ndarray) -> np.ndarray:
    """Simple pairwise scaling - used only if the LP fails."""
    n = centres.shape[0]
    r = np.minimum.reduce([centres[:, 0],
                           centres[:, 1],
                           1.0 - centres[:, 0],
                           1.0 - centres[:, 1]])
    for i in range(n):
        for j in range(i + 1, n):
            d = np.linalg.norm(centres[i] - centres[j])
            if r[i] + r[j] > d:
                scale = d / (r[i] + r[j])
                r[i] *= scale
                r[j] *= scale
    return np.maximum(r, 0.0)


def _lp_optimal(centres: np.ndarray) -> tuple[np.ndarray, float]:
    """Solve the LP that maximises sum(r_i) for the fixed centres."""
    wall = np.minimum.reduce([centres[:, 0],
                              centres[:, 1],
                              1.0 - centres[:, 0],
                              1.0 - centres[:, 1]])
    bounds = [(0.0, w) for w in wall]

    diffs = centres[PAIR_I] - centres[PAIR_J]
    pair_dist = np.linalg.norm(diffs, axis=1)

    if _SCIPY:
        res = linprog(c=-np.ones(N),
                      A_ub=A_UB,
                      b_ub=pair_dist,
                      bounds=bounds,
                      method='highs',
                      options={'presolve': True})
        if res.success:
            radii = np.clip(res.x, 0.0, wall)
            return radii, float(radii.sum())
    radii = _fallback_radii(centres)
    return radii, float(radii.sum())

def _hex_lattice(dx: float) -> np.ndarray:
    """Generate points on a hexagonal lattice (up to 26 points)."""
    dy = dx * np.sqrt(3.0) / 2.0
    pts = []
    row = 0
    y = 0.0
    while y <= 1.0 + 1e-12:
        offset = 0.0 if (row % 2 == 0) else dx / 2.0
        x = offset
        while x <= 1.0 + 1e-12:
            pts.append([x, y])
            x += dx
        y += dy
        row += 1
    pts = np.asarray(pts)
    pts = pts[(pts[:, 0] >= 0.0) & (pts[:, 0] <= 1.0) &
              (pts[:, 1] >= 0.0) & (pts[:, 1] <= 1.0)]
    if pts.shape[0] > N:
        margins = np.minimum.reduce([pts[:, 0],
                                    pts[:, 1],
                                    1.0 - pts[:, 0],
                                    1.0 - pts[:, 1]])
        keep = np.argsort(-margins)[:N]
        pts = pts[keep]
    elif pts.shape[0] < N:
        rng = np.random.default_rng(0)
        extra = rng.uniform(0.0, 1.0, size=(N - pts.shape[0], 2))
        pts = np.vstack([pts, extra])
    return pts


def _init_farthest(rng: np.random.Generator, n: int = N) -> np.ndarray:
    """Greedy farthest-point placement (points stay inside [0,1]^2)."""
    pts = []
    pts.append(rng.uniform(0.1, 0.9, size=2))
    while len(pts) < n:
        cand = rng.uniform(0.1, 0.9, size=(200, 2))
        existing = np.asarray(pts)
        dists = np.linalg.norm(
            cand[:, None, :] - existing[None, :, :], axis=2)
        min_d = dists.min(axis=1)
        best = np.argmax(min_d)
        pts.append(cand[best])
    return np.asarray(pts)


def _init_grid_jitter(rng: np.random.Generator,
                      jitter: float = 0.02) -> np.ndarray:
    """5x5 grid (0.1-0.9) with small random jitter + one extra point."""
    xs = np.linspace(0.1, 0.9, 5)
    ys = np.linspace(0.1, 0.9, 5)
    xv, yv = np.meshgrid(xs, ys)
    base = np.column_stack([xv.ravel(), yv.ravel()])  # 25 points
    base += rng.uniform(-jitter, jitter, base.shape)
    base = np.clip(base, 0.01, 0.99)
    extra = rng.uniform(0.01, 0.99, size=(1, 2))
    return np.vstack([base, extra])


def _init_random(rng: np.random.Generator) -> np.ndarray:
    """Pure uniform random points."""
    return rng.uniform(0.0, 1.0, size=(N, 2))

def _hill_climb(start: np.ndarray, rng: np.random.Generator,
                iterations: int = 1500
                ) -> tuple[np.ndarray, np.ndarray, float]:
    best_c = start.copy()
    best_r, best_sum = _lp_optimal(best_c)

    temperature = 0.02
    for it in range(iterations):
        cand_c = best_c.copy()
        k = rng.integers(1, 4)
        sel = rng.choice(N, size=k, replace=False)
        max_step = 0.09 * (1.0 - it / iterations) + 0.005
        cand_c[sel] += rng.normal(scale=max_step, size=(k, 2))
        cand_c = np.clip(cand_c, 0.0, 1.0)

        cand_r, cand_sum = _lp_optimal(cand_c)
        delta = cand_sum - best_sum

        if delta > 1e-9:
            best_c, best_r, best_sum = cand_c, cand_r, cand_sum
            temperature = max(temperature * 0.95, 1e-6)
        else:
            if (temperature > 0.0
                    and rng.random() < np.exp(delta / temperature)):
                best_c, best_r, best_sum = cand_c, cand_r, cand_sum
            temperature *= 0.9995
    return best_c, best_r, best_sum

def _refine_slsqp(centres: np.ndarray, radii_start: np.ndarray,
                  rng: np.random.Generator
                  ) -> tuple[np.ndarray, np.ndarray, float]:
    if not _SCIPY:
        return centres, radii_start, radii_start.sum()

    n = N
    x0 = np.empty(3 * n)
    x0[:n] = centres[:, 0]
    x0[n:2 * n] = centres[:, 1]
    x0[2 * n:] = radii_start

    bounds = ([(0.0, 1.0)] * n
              + [(0.0, 1.0)] * n
              + [(0.0, None)] * n)

    def cons_fun(x):
        xs = x[:n]
        ys = x[n:2 * n]
        rs = x[2 * n:]
        c1 = xs - rs
        c2 = ys - rs
        c3 = (1.0 - xs) - rs
        c4 = (1.0 - ys) - rs
        dx = xs[:, None] - xs[None, :]
        dy = ys[:, None] - ys[None, :]
        d = np.sqrt(dx * dx + dy * dy)
        iu = np.triu_indices(n, 1)
        c5 = d[iu] - (rs[iu[0]] + rs[iu[1]])
        return np.concatenate([c1, c2, c3, c4, c5])

    constraints = {'type': 'ineq', 'fun': cons_fun}

    def obj_fun(x):
        return -np.sum(x[2 * n:])

    res = minimize(obj_fun,
                   x0,
                   method='SLSQP',
                   bounds=bounds,
                   constraints=[constraints],
                   options={'ftol': 1e-9, 'maxiter': 2000,
                            'disp': False})

    if not res.success:
        final_c = centres
    else:
        xs_opt = res.x[:n]
        ys_opt = res.x[n:2 * n]
        final_c = np.vstack([xs_opt, ys_opt]).T

    final_r, final_sum = _lp_optimal(final_c)
    return final_c, final_r, final_sum

def _make_feasible(centres: np.ndarray, radii: np.ndarray,
                   eps: float = 1e-12) -> np.ndarray:
    wall = np.minimum.reduce([centres[:, 0],
                              centres[:, 1],
                              1.0 - centres[:, 0],
                              1.0 - centres[:, 1]])
    radii = np.minimum(radii, wall - eps)
    for _ in range(5):
        changed = False
        for i in range(N):
            for j in range(i + 1, N):
                d = np.linalg.norm(centres[i] - centres[j])
                if radii[i] + radii[j] > d - eps:
                    scale = (d - eps) / (radii[i] + radii[j])
                    radii[i] *= scale
                    radii[j] *= scale
                    changed = True
        if not changed:
            break
    return np.maximum(radii, 0.0)

def construct_packing() -> tuple[np.ndarray, np.ndarray, float]:
    rng = np.random.default_rng(123456789)
    best_sum = -np.inf
    best_centres = None
    best_radii = None

    # -- Stage 1: diverse starts --
    candidates = []
    for dx in np.linspace(0.16, 0.30, 15):
        base = _hex_lattice(dx)
        for jitter_scale in (0.0, 0.008, 0.02, 0.04):
            for _ in range(5):
                jitter = rng.normal(
                    scale=jitter_scale, size=base.shape)
                centres = np.clip(base + jitter, 0.0, 1.0)
                radii, s = _lp_optimal(centres)
                candidates.append((s, centres, radii))

    for _ in range(5):
        centres = _init_farthest(rng)
        radii, s = _lp_optimal(centres)
        candidates.append((s, centres, radii))

    for jit in (0.015, 0.03):
        for _ in range(5):
            centres = _init_grid_jitter(rng, jitter=jit)
            radii, s = _lp_optimal(centres)
            candidates.append((s, centres, radii))

    for _ in range(5):
        centres = _init_random(rng)
        radii, s = _lp_optimal(centres)
        candidates.append((s, centres, radii))

    candidates.sort(key=lambda x: x[0], reverse=True)
    top_candidates = candidates[:6]

    # -- Stage 2: hill climbing --
    for s0, cent0, rad0 in top_candidates:
        cent_opt, rad_opt, sum_opt = _hill_climb(
            cent0, rng, iterations=1800)
        if sum_opt > best_sum:
            best_sum = sum_opt
            best_centres = cent_opt
            best_radii = rad_opt

    # -- Stage 3: SLSQP refinement --
    if best_centres is not None:
        refined_c, refined_r, refined_sum = _refine_slsqp(
            best_centres, best_radii, rng)
        if refined_sum > best_sum:
            best_sum = refined_sum
            best_centres = refined_c
            best_radii = refined_r

    final_radii = _make_feasible(best_centres, best_radii)
    final_sum = float(final_radii.sum())
    return best_centres, final_radii, final_sum

def run_packing() -> None:
    centres, radii, _ = construct_packing()
    out = sys.stdout
    fmt = "{:.9f} {:.9f} {:.9f}\n"
    for i in range(N):
        out.write(fmt.format(
            centres[i, 0], centres[i, 1], radii[i]))


if __name__ == "__main__":
    run_packing()
\end{lstlisting}

\subsection{Correlation between confidence and score across loops}
\begin{figure*}[h]
    \centering
    \includegraphics[width=0.7\linewidth]{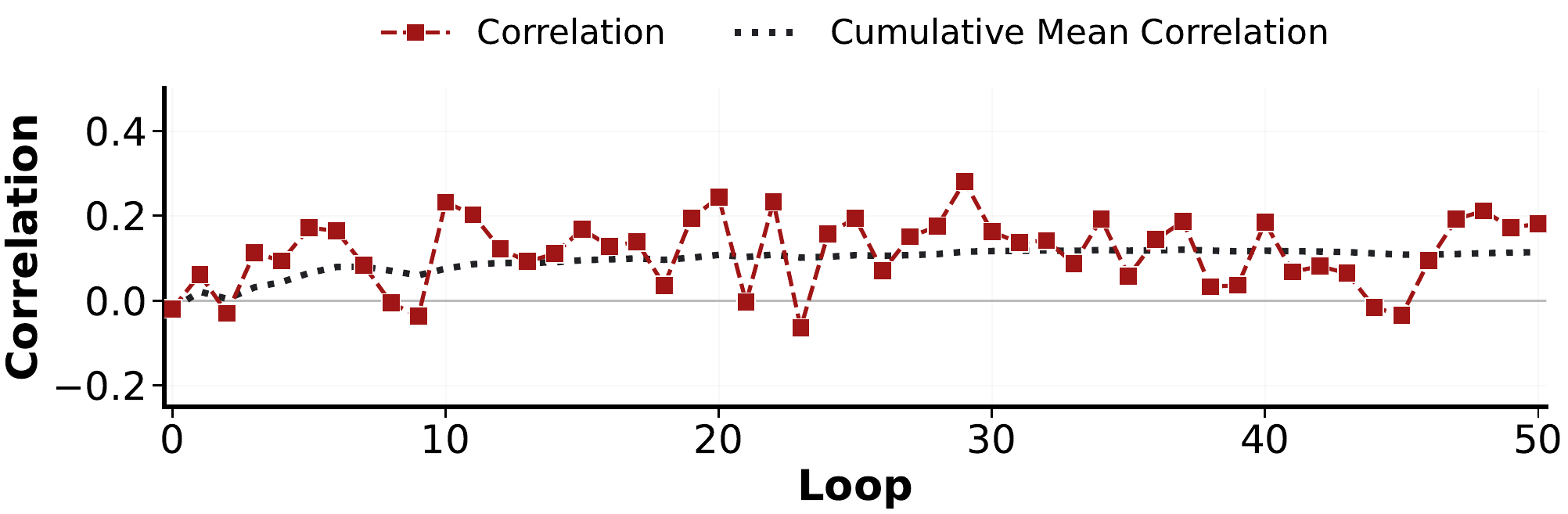}
    \caption{\textbf{Spearman rank correlation between confidence and score}}
    \label{fig:circle_packing_correlation}
\end{figure*}

\end{document}